\documentclass[lettersize,journal]{IEEEtran}
\usepackage{amsmath,amsfonts}
\usepackage{algorithmic}
\usepackage{algorithm}
\usepackage{array}
\usepackage{textcomp}
\usepackage{stfloats}
\usepackage{url}
\usepackage{verbatim}
\usepackage{graphicx}
\usepackage{natbib}
\usepackage{caption} 
\usepackage{multirow}
\usepackage[utf8]{inputenc}
\usepackage{subcaption}
\usepackage{booktabs}
\usepackage{hyperref}
\usepackage{xcolor}
\usepackage{tikz}
\usetikzlibrary{positioning, shapes.geometric, arrows.meta, backgrounds, fit}

\definecolor{virdark}{HTML}{440154}
\definecolor{viryellow}{HTML}{FDE725}

\newcommand{\mpc}[1]{\textcolor{black}{#1}}

\begin{document}

\title{Plume Segmentation from MethaneSAT with Cross-Sensor Transfer Learning and Physics-Informed Postprocessing}

\author{Manuel Pérez-Carrasco$^{1,2}$, Maya Nasr$^{3,4}$, Zhan Zhang$^{3,4}$, Apisada Chulakadabba$^{2,5}$, Javier Roger$^{6,7}$, \\ Raia Ottenheimer$^{8}$, Sébastien Roche$^{3,4}$, Maryann Sargent$^{4}$, Chris Chan Miller$^{3,4}$, Daniel Varon$^{9,10}$, Jack Warren$^{3}$, Luis Guanter$^{6}$, Kang Sun$^{11}$, Jonathan Franklin$^{4}$, Jia Chen$^{2}$, Cecilia Garraffo$^{1}$, Xiong Liu$^{1}$, Ritesh Gautam$^{3}$, and Steven Wofsy$^{4}$}


\markboth{SUBMITTED TO IEEE TRANSACTIONS ON GEOSCIENCE AND REMOTE SENSING}
{Pérez-Carrasco \MakeLowercase{\textit{et al.}}: Methane Plumes Detection}


\maketitle

\footnotetext[1]{Center for Astrophysics $|$ Harvard \& Smithsonian, Cambridge, MA}
\footnotetext[2]{Professorship of Environmental Sensing and Modeling, Technical University of Munich, Munich, Germany}
\footnotetext[3]{Environmental Defense Fund, Washington, D.C., USA}
\footnotetext[4]{Department of Earth and Planetary Sciences, Harvard University, Cambridge, MA, USA}
\footnotetext[5]{The Joint Graduate School of Energy and Environment, King Mongkut’s University of Technology Thonburi, 126 Pracha Uthit Road, Bang Mot, Thung Khru, Bangkok 10140, Thailand}
\footnotetext[6]{Research Institute of Water and Environmental Engineering (IIAMA), Universitat Politècnica de València, València, Spain}
\footnotetext[7]{Institute of Environmental Physics (IUP), University of Bremen, Bremen, Germany}
\footnotetext[8]{John A. Paulson School of Engineering and Applied Sciences, Harvard University, Cambridge, MA, USA}
\footnotetext[9]{Department of Aeronautics and Astronautics, Massachusetts Institute of Technology, 77 Massachusetts Avenue, Cambridge, MA, USA}
\footnotetext[10]{Institute for Data, Systems, and Society, Massachusetts Institute of Technology, 77 Massachusetts Avenue, Cambridge, MA, USA}
\footnotetext[11]{Department of Civil, Structural and Environmental Engineering, University at Buffalo, Buffalo, NY, USA}

\begin{abstract}
Automated detection and masking of individual methane plumes from satellite imagery is important for operational emission attribution and quantification. We present a machine learning framework for plume detection from MethaneSAT retrieved column-averaged dry-air mole fractions of methane (XCH$_4$).  We address two core challenges: the scarcity of labeled MethaneSAT data and the need for inference reliability across diverse atmospheric and surface conditions. We first demonstrate that Mask R-CNN with a ResNet-50 backbone \mpc{outperforms a U-Net semantic segmentation baseline} on both MethaneAIR (an airborne version of MethaneSAT) and MethaneSAT data, with pixel-level F1 score gains of 10.49 and 5.48 percentage points respectively. To address MethaneSAT data scarcity, we evaluate three cross-sensor transfer strategies leveraging MethaneAIR flights and synthetic plumes. Mask R-CNN with ResNet-50 fine-tuned from MethaneAIR pre-trained weights is the most effective strategy, achieving instance-level precision of 0.60 and recall of 0.98 at the baseline operating point. A physics-informed post-processing pipeline converts detections into two operationally distinct modes.  The first is a high-sensitivity mode that applies morphological filtering and proximity-based merging for comprehensive emission screening, achieving precision of 0.71 and recall of 0.94.  The second is a high-precision mode that additionally applies a distribution-based classifier for confident source attribution, achieving precision of 0.92 and recall of 0.70. Manual review of detections classified as false positives against our
wavelet-based \mpc{reference plume mask labels} shows that some correspond to real methane enhancements excluded by the conservative labeling criteria, so the reported precision likely underestimates true detection performance. Finally, we propose a confidence-weighted aggregation scheme that produces confidence-weighted detection maps for downstream screening. Together, these contributions demonstrate a framework for automated plume segmentation of the MethaneSAT data.

Our data and code used in this work are publicly available at:  \url{https://doi.org/10.7910/DVN/FR959H}
\end{abstract}

\begin{IEEEkeywords}
Remote Sensing, Methane plume detection, Machine Learning, MethaneSAT
\end{IEEEkeywords}

\thanks{This work has been submitted to the IEEE for possible publication. Copyright may be transferred without notice, after which this version may no longer be accessible.}

\section{Introduction}

\IEEEPARstart{T}{he} oil and gas sector is responsible for roughly 30\% of global anthropogenic methane emissions \citep{iea_2023}, where emissions from both smaller-emitting sources dispersed across wide areas and concentrated high-emitting facilities are important for mitigation of methane emissions  \cite{brandt_2014, zavala_2015, frankenberg_2016, Varon_2021, Omara_2022,  Williams_2025}.  Identifying and attributing individual sources is therefore essential for cost-effective mitigation efforts due to methane's high warming potential --- 80 times that of CO$_2$ over a 20-year horizon \citep{myhre_2013, Etminan_2016}. 
International efforts reflect this urgency: the Global Methane Pledge and the Oil and Gas Decarbonization Charter together bind over 150 countries and more than 50 major producers to substantial emission reductions by 2030 \citep{weforum2024methane, ogdc_2023}, but meeting these targets demands monitoring systems that can locate and quantify individual emission events at scale.

\mpc{Space-based imaging spectroscopy is a promising pathway to closing this monitoring gap \cite{jacob_2016}. High-resolution instruments such as GHGSat \citep{jervis_2021}, EMIT \citep{Green_2023}, EnMAP \citep{Storch_2023}, PRISMA \citep{COGLIATI_2021}, and Carbon Mapper's Tanager-1 \cite{duren_2020, Duren_2025} resolve facility-level plumes through absorption features in the shortwave infrared \citep{frankenberg_2016, cusworth_2021, Varon_2021, Zhang_2022, Guanter_2021, Irakulis_2022}, while wide-area mappers such as TROPOMI \citep{Veefkind_2012}, GOSAT \citep{kasuya_2009}, and GOES provide daily regional coverage but cannot isolate point sources \citep{Watine-Guiu_2023}. An observational gap therefore persists at basin scale, where quantification needs both the resolution to resolve individual sources and the coverage to assess an entire production region.}

\begin{figure*}[ht!]
  \centering
\includegraphics[width=.98\linewidth]{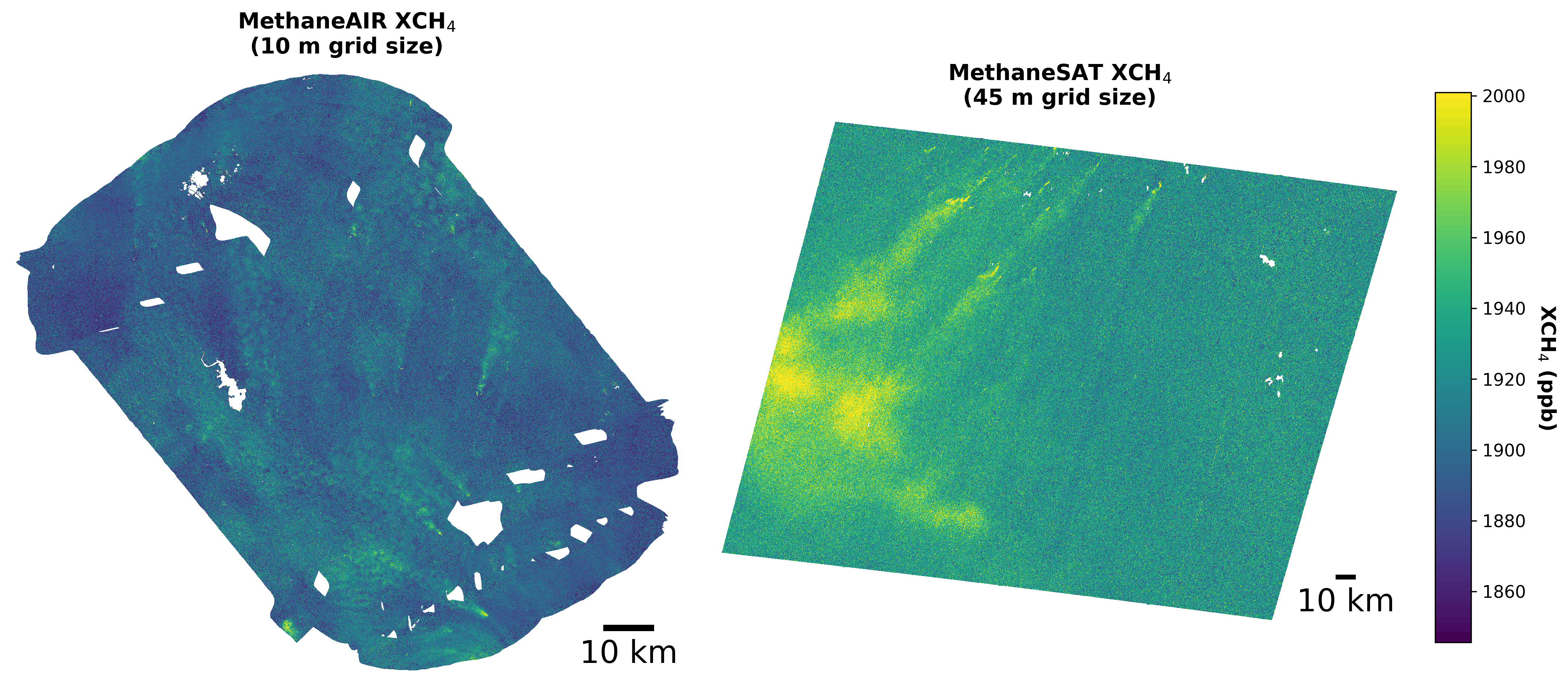}
\caption{\textbf{Gridded column-averaged dry-air mole fractions of methane (XCH$_4$)} from the Permian Basin (a) MethaneAIR scene covering 120 km $\times$ 130 km at 10 m grid size and (b) MethaneSAT scene covering 220 km $\times$ 335 km at 45 m grid size over an oil and gas production region. Both L3 products are derived using the CO$_2$ proxy method applied to vertical column densities of CH$_4$ and CO$_2$. Enhanced XCH$_4$ values (yellow colors) indicate elevated methane concentrations associated with dispersed area sources and high-emitting point sources, while lower background concentrations appear in dark green-blue-purple. White areas represent data gaps from cloud screening, quality filtering, or for MethaneAIR gaps in coverage from the flight pattern.}
  \label{fig:xch4maps}
\end{figure*}

MethaneSAT \cite{Rohrschneider, ccm2026Omethanesat} was designed specifically to bridge this gap. Launched in March 2024, it combined high sensitivity with wide-area coverage, observing oil and gas basins at approximately $110 \times 400$ m spatial resolution at nadir, covering ~220 km x 220 km target areas to capture both dispersed emissions and discrete plumes at basin scales. 
Methane column-averaged dry-air mole fractions (XCH$_4$) are derived from the CO$_2$ proxy retrieval method \cite{Frankenberg_2005, Krings_2011, Chan_2024}, revealing spatial patterns of enhanced methane concentration that indicate emission sources. All MethaneSAT analysis in this work uses the Level 3 (L3) product, in which the native Level 2 (L2) pixels are oversampled onto a regular 45 m grid using a physics-based oversampling method that models each pixel as a two-dimensional spatial response function \cite{Sun_2018} (see Figure \ref{fig:xch4maps}). Over approximately fourteen months of operation, MethaneSAT acquired data across major oil and gas production regions worldwide, providing a basin-scale view of methane emissions at the spatial resolution needed to resolve individual sources. On June 20, 2025, MethaneSAT lost contact with ground operations marking the end of the operational lifetime of the satellite \citep{EDF_2025}. The dataset collected is therefore less than the intended lifetime of the mission, making it imperative to extract maximum scientific value from existing observations through effective automated analysis methods. Its airborne precursor, MethaneAIR \cite{Conway_2024}, shares similar instrument characteristics and has accumulated over 80 annotated scenes across major U.S. oil and gas basins since 2021, providing a complementary source of plume observations using the same CO$_2$ proxy retrieval methodology at finer 10 m grid size.

Translating XCH$_4$ maps into actionable emission information from point sources requires identifying and masking individual methane plumes. The wavelet transform-based detection method \cite{zhang_2026} applies 2D discrete wavelet transforms to enhance plume signals above background noise, achieving higher sensitivity to faint plumes than conventional thresholding approaches. However, the method still requires careful parameter tuning across scenes and atmospheric conditions, and residual false positives from dispersed enhancements, overlapping plumes, and background variability necessitate manual inspection to produce clean detection catalogs. Similar quality control challenges are faced by plume detection methods across airborne and spaceborne platforms \citep{JONGARAMRUNGRUANG_2022, rouet_2024, ruzicka_2023, Bue_2025}. Machine learning offers a complementary pathway: by learning directly from labeled concentration imagery, deep learning models can capture the morphological and contextual signatures of real plumes in ways that hand-crafted signal processing cannot easily encode.

\mpc{Most existing methods for methane plume segmentation use semantic segmentation, particularly U-Net \cite{Ronneberger2015, wang_2020, ruzicka_2023, JONGARAMRUNGRUANG_2022, Schuit_2023, Mancoridis_2025, Xu_2025}, which classifies each pixel as plume or background and therefore merges spatially proximate plumes into a single connected region. This blocks source attribution, which is consequential for MethaneSAT because it targets oil and gas basins where several sources often occur in one scene. Instance segmentation avoids this by predicting a separate mask for each plume, and Mask R-CNN \cite{he2017mask} has been applied in this way to airborne and spaceborne methane data \cite{Kumar_2020, si_2024}, and Bue et al. \cite{Bue_2025} added an instance-level objective to a U-Net. Section~\ref{sec:related_work} reviews both lines of work. Two challenges specific to the MethaneSAT setting must be solved before an instance-level system is operationally reliable.} 

The first challenge is data scarcity. Supervised instance segmentation requires substantial volumes of labeled examples to learn reliable plume representations, yet the number of MethaneSAT scenes with validated plume annotations is limited. Training a high-capacity detection model from this data alone risks overfitting and poor generalization across the diverse atmospheric and surface conditions encountered. We address this through two complementary strategies. First, we leverage MethaneAIR, whose shared CO$_2$ proxy retrieval makes learned plume representations directly transferable to the MethaneSAT domain, and we treat cross-sensor fine-tuning as the primary mechanism for overcoming the MethaneSAT data scarcity. Second, we augment training with simulated plumes generated from atmospheric dispersion models \cite{Varon_2021, chulakadabba2023methane}, which provide controlled diversity across emission rates and meteorological conditions that real data alone cannot cover. Together, these two sources extend the effective training distribution well beyond what the MethaneSAT archive provides on its own.

The second challenge is inference reliability. A model trained on patches must be applied to full satellite scenes spanning thousands of pixels, and the raw predictions it produces will include false positives arising from retrieval artifacts and instrument noise. Yet, the appropriate treatment of uncertain detections is not uniform across monitoring applications. Regulatory reporting and source attribution workflows require high confidence in every flagged detection, tolerating missed sources over incorrectly attributed ones. Comprehensive emission surveys and longitudinal monitoring across repeated overpasses, by contrast, benefit from retaining marginal detections that may represent real low-flux sources near the instrument detection limit, whose presence can be confirmed or refuted by accumulating evidence across multiple passes. These two use cases have fundamentally different precision-recall requirements and cannot be simultaneously satisfied by a single detection threshold. We therefore design our inference pipeline to support two distinct modes: a high-sensitivity mode that applies conservative morphological filtering and merges overlapping predictions for longitudinal emission monitoring, and a high-precision mode that suppresses spurious detections for confident source attribution. For the latter, we use the distribution-based classifier proposed by Ottenheimer et al.~\cite{Ottenheimer2026} and compare it with size thresholding (Appendix~\ref{app:size_threshold_vs_QND}).

This paper makes three contributions toward automated methane plume detection for MethaneSAT. \textbf{(i) We present the first application of instance segmentation to MethaneSAT and MethaneAIR plume detection.} We show that Mask R-CNN with a ResNet-50 backbone achieves strong detection performance on both sensors, improving precision over a U-Net baseline while maintaining competitive recall. \textbf{(ii) We demonstrate that cross-sensor fine-tuning from MethaneAIR is the most effective adaptation strategy for MethaneSAT under limited labeled data.} At scene level, fine-tuning from MethaneAIR pre-trained weights reduces false positives; at patch level the gain has the same sign but is within the fold-to-fold spread. Synthetic data augmentation gives no measurable benefit beyond fine-tuning alone. \textbf{(iii) We introduce a physics-informed post-processing pipeline} that converts patch detections into two operationally distinct scene-level modes\mpc{, one enforcing advection-consistent morphology, the other albedo-independent methane enhancement}. The high-sensitivity mode targets comprehensive emission screening, while the high-precision mode reduces artifact-driven false positives for confident source attribution. We further propose a spatial aggregation scheme that produces smooth confidence-weighted detection maps across overlapping sliding windows \mpc{as an exploratory screening product for downstream analysis}. \mpc{The combined contribution is an alternative operational MethaneSAT plume-detection workflow, in which model selection, cross-sensor transfer, and physics-informed post-processing are evaluated end to end at the scene level.}

\section{Related Work}
\label{sec:related_work}

\subsection{Satellite Methane Monitoring and Plume Detection}

Remote sensing-based detection and quantification of methane point sources from imaging spectrometers typically involves two stages: retrieving XCH$_4$ from measured radiances, and then detecting and quantifying individual plumes from the resulting concentration maps. This work focuses on plume detection; we briefly describe both.

MethaneAIR \cite{Conway_2024} and MethaneSAT \cite{Rohrschneider} retrieve XCH$_4$ using the CO$_2$ proxy method \cite{Chan_2024,ccm2026Omethanesat}, in which the ratio of the retrieved CH$_4$ and CO$_2$ vertical columns is scaled by an a priori estimate of the dry-air column-averaged CO$_2$ mole fraction (XCO$_2$), implicitly accounting for aerosol-induced light-path modifications and other systematic retrieval errors affecting both column amounts. Most other point-source imagers, including EMIT, PRISMA, EnMAP, and the recently launched Tanager-1 \cite{Roger_2024, Duren_2025}, instead use the matched filter (MF) algorithm \cite{Thorpe_2014, Guanter_2021}: background radiance is modeled as a multivariate Gaussian and pixels deviating toward the known methane spectral signature are flagged as enhancements \cite{Cusworth_2018}. 
Foote et al. added sparsity regularization and albedo correction to improve sensitivity in complex surface environments \cite{foote_2020}, and MF was recently applied to MethaneAIR \cite{Guanter_2025} and MethaneSAT \cite{Guanter_2026} data. For multispectral sensors such as Sentinel-2, where full spectral inversion is unavailable, band ratio methods exploiting the differential methane absorption between SWIR bands B11 and B12 serve the same purpose \cite{Varon_2021, sanchezgarcia_2022}.

Given a concentration map, detecting and quantifying individual plume fluxes requires a separate step. The Integrated Mass Enhancement (IME) method \cite{Thompson_2016, Varon_2018}, extended to MethaneAIR by Chulakadabba et al. \cite{chulakadabba2023methane}, relates the excess methane mass within a masked plume region to the emission rate; the mask boundary determines both the integration area and the scale length used to convert total mass to an emission rate, so mask quality propagates directly into flux uncertainty.

The divergence integral (DI) method \cite{chulakadabba2023methane} has both a detection and a quantification component. For detection, Warren et al. \cite{Warren_2025} applied a gridded DI thresholding approach to MethaneAIR data: flux divergence is computed across the scene, and two-stage thresholding and clumping on both the flux map and the XCH$_4$ map identify candidate plumes. For quantification, Gauss's divergence theorem is applied over a series of expanding rectangles around the plume origin and the estimates are averaged.

Zhang et al. (2026) \cite{zhang_2026} introduced a wavelet-based automatic masking method that applies a 2D discrete wavelet transform to XCH$_4$ imagery to enhance plume signals while suppressing background noise, followed by thresholding, connected-component labeling, and filtering by concentration hotspot, shape, and wind direction. Applied to MethaneAIR and MethaneSAT data, the wavelet method detected 60\% more plumes than the DI thresholding method, primarily in lower flux ranges, while producing fewer false positives, without requiring labeled training data. 

Despite these advances, plume detection and masking at scale remains a bottleneck. Threshold-based methods require expert intervention and parameter tuning to distinguish real enhancements from instrument artifacts and background noise. This scalability problem directly affects missions such as MethaneSAT \cite{Rohrschneider} and MethaneAIR \cite{Conway_2024}, which require consistent and reproducible plume detection across different scenes. The present work addresses this by developing an instance segmentation model for methane plume detection that operates on XCH$_4$ maps and generalizes across the MethaneAIR and MethaneSAT platforms.

\subsection{Machine Learning for Methane Plume Segmentation}

Machine learning approaches have emerged as promising tools for automating methane plume detection from satellite imagery. The encoder-decoder architecture of U-Net \cite{Ronneberger2015}, with its skip connections preserving fine spatial detail across scales, has become the dominant framework for pixel-level plume segmentation \cite{vaughan_2024b}. For example, Joyce et al. applied a standard U-Net to PRISMA hyperspectral data, training on synthetic plumes generated via Large Eddy Simulation (LES) to supplement limited real data \cite{joyce_2023}. CH4Net \cite{Vaughan_2024} applied a U\text{-}Net architecture to Sentinel\text{-}2 imagery for monitoring known super\text{-}emitter sites. The HyperSTARCOP model introduced a lightweight U-Net with a MobileNetV2 encoder that ingests matched filter products rather than raw spectra \cite{ruzicka_2023}. For GHGSat data, U-Plume demonstrated automated segmentation and quantification in an end-to-end pipeline \cite{Bruno_2024}. Recently, AttMetNet has incorporated attention gates in U-Net's skip connections and a Normalized Difference Methane Index \cite{Xu_2025} as an additional input channel \cite{ahsan_2025}.

Vision Transformer architectures (ViT; \cite{Dosovitskiy_2020}) have extended these capabilities by capturing long-range spatial dependencies that convolutional encoders struggle to model, a property particularly relevant for elongated plume structures whose spatial extent can span hundreds of pixels. Rouet-Leduc and Hulbert combined a ViT encoder with a U-Net decoder and multi-temporal Sentinel-2 inputs \cite{rouet_2024}. HyperspectralViTs adapted SegFormer \cite{xie_2021} and EfficientViT \cite{liu_2023} for direct processing of raw hyperspectral cubes, eliminating the matched filter preprocessing bottleneck and improving F1 scores by 13–27\% \cite{ruvzivcka_2025}.

Instance segmentation extends semantic segmentation by incorporating additional detection and pixel-level masking of individual object instances \cite{he2017mask}. Popular models such as Mask R-CNN \cite{he2017mask} have received comparatively less attention in the methane detection literature despite its natural fit for plume source attribution. Kumar et al. applied Mask R-CNN to AVIRIS-NG and GHGSat data for simultaneous plume detection and boundary segmentation \cite{Kumar_2020}. Si et al. developed a multi-task learning framework combining Mask R-CNN with ResNet-50 for joint segmentation and quantification on PRISMA data \cite{si_2024}. Recently, Bue et al. jointly optimized a U-Net for both pixelwise segmentation and instance-level detection of AVIRIS-NG and EMIT methane plumes using a multi-task loss inspired by Mask R-CNN \cite{Bue_2025}. This approach was recently extended by Batchu et al. \cite{batchu_2026} to detect methane plumes directly from EMIT radiance data. \mpc{Bue et al. \cite{Bue_2025} report pixelwise and instance-level metrics for a U-Net trained with and without an instance objective, however no work has compared a proposal-based instance segmentation model against a semantic segmentation baseline on identical methane data.}

A persistent challenge across all supervised approaches is the scarcity of labeled plume data relative to the diversity of instrument configurations, atmospheric conditions, and surface types encountered in operational monitoring. Several studies have addressed this through synthetic data generated from WRF-LES atmospheric simulations \cite{radman_2023, joyce_2023, Bruno_2024}, but the domain gap between simulated and real plume morphology can degrade performance on real data. Cross-sensor transfer learning compounds this further, as models trained on one instrument routinely fail when applied to another due to differences in spatial resolution, noise characteristics, and retrieval dynamic range. 

Two strategies have been explored to bridge this gap. The first operates at the data level, either by translating source-domain images into the target domain using generative models, or by jointly training on real and synthetic data. Mancoridis et al. \cite{Mancoridis_2025} used CycleGAN-based data translation \cite{zhu_2017} for AVIRIS-NG to EMIT generalization, finding that translating spaceborne data into the airborne domain before classification achieved the best results (F1 = 0.88), substantially outperforming zero-shot transfer (F1 = 0.52) and training from scratch (F1 = 0.76). Joint training on real and synthetic data has also been explored as a lighter-weight alternative that does not require a separate generative model \cite{joyce_2023, Bruno_2024}. \mpc{Si et al. \cite{si_2024} extend the use of simulated plumes to a multi-task setting, training networks for concentration estimation, plume identification, and emission quantification on PRISMA hyperspectral imagery; their focus is quantitative retrieval from hyperspectral radiances, whereas we target instance-level detection on gridded XCH$_4$ maps with scene-level post-processing.} Curriculum learning extends this idea by structuring the exposure schedule, beginning training with a higher proportion of synthetic data and progressively shifting toward real data as training proceeds \cite{bengio_2009}.

The second strategy operates at the feature level. Fine-tuning adapts a model pre-trained on a data-rich source domain to the target domain using available labeled data \cite{yosinski_2014}, allowing target-domain training to build on representations already learned from the source rather than initializing from scratch. Fine-tuning has emerged as a reliable cross-sensor adaptation strategy across remote sensing tasks \cite{Mancoridis_2025}. 

A complementary line of work addresses inference reliability by classifying artifact-driven detections after the primary segmentation step. Automated plume finding algorithms operating on MethaneAIR data produce false positive rates exceeding 60\%, driven primarily by surface albedo variability, clouds, linear striping artifacts, and instrument noise \cite{Ottenheimer2026}. Ottenheimer et al. \cite{Ottenheimer2026} proposed a distribution-based statistical framework that characterizes the within-mask distributions of methane enhancement and surface albedo using percentile-based polynomial representations ---the Quantile Normality Deviation (QND) polynomials--- and combines them with geometric plume features in a random forest classifier to discriminate real plumes from artifact-driven detections. Applied to MethaneAIR data at 10 m and 30 m resolution, the method achieves F1 scores above 91\% and demonstrates cross-instrument generalization to Tanager without retraining, suggesting that the discriminative signatures captured by the QND descriptors reflect physical properties of methane enhancements rather than instrument-specific noise patterns.

Our work evaluates fine-tuning, joint training, and curriculum learning as primary cross-sensor adaptation strategies, within a unified experimental framework spanning MethaneAIR, synthetic data, and MethaneSAT. Post-hoc artifact classification via the QND framework is then integrated as the primary false positive suppression mechanism in our high-precision inference mode.

\section{Data}
\label{sec:data}

\subsection{Methane Concentration Imagery}
Our study utilizes XCH$_4$ maps from both MethaneAIR and MethaneSAT as input data for training and evaluating machine learning-based plume detection algorithms. 
\mpc{Throughout, \emph{XCH$_4$ map} denotes the gridded L3 column-averaged dry-air mole fraction in ppb (absolute values, as in Fig.~\ref{fig:xch4maps}); \emph{XCH$_4$ enhancement} ($\Delta$XCH$_4$) denotes the excess above the local background in ppb, used when describing plumes and synthetic plume injection; and \emph{normalized XCH$_4$} denotes the per-scene z-scored field used as model input (Sec.~\ref{sec:data_prep})}.

\subsubsection{MethaneAIR Dataset}

MethaneAIR is an airborne imaging spectrometer that served as a testbed for the MethaneSAT satellite mission. Operating at a nominal altitude of 12 km with a 4.5 km observing swath, MethaneAIR achieves a native spatial resolution of 5 $\times$ 25 m \cite{Conway_2024}, and retrievals are performed with 5 pixels aggregated across-track, yielding 25 $\times$ 25 m spatial resolution at L2 \cite{Chan_2024}. The instrument uses two Offner spectrometers covering wavelength ranges of 1236–1319 nm (O$_2$ sensor) and 1592–1678 nm (CH$_4$ sensor), with a spectral resolution of approximately 0.28 nm full-width-at-half-maximum (FWHM) and 0.10 nm sampling for the CH$_4$ sensor (See Table S1 of \cite{chulakadabba2023methane} for details). We utilize XCH$_4$ maps at 10 m grid size, generated using the CO$_2$ proxy retrieval method \cite{Chan_2024,Sun_2018}.

Our MethaneAIR dataset comprises 84 scenes from campaigns conducted between 2021 and 2025, covering major oil and gas production regions across the United States, including the Permian, Appalachian, and Uinta basins. All the MethaneAIR target basins can be seen in Appendix \ref{app:target-basins}. Cloud-contaminated pixels and low-quality retrievals were filtered during upstream preprocessing using a cloud screening algorithm based on O$_2$ surface pressure retrievals \cite{Chan_2024, perezcarrasco_2025}. The dataset includes scenes with varying atmospheric conditions, surface characteristics, and emission intensities, providing diverse training examples for machine learning models. A sample from the MethaneAIR dataset is shown in Figure \ref{fig:methane_comparison}(a).

\subsubsection{MethaneSAT Dataset}

MethaneSAT, launched in March 2024 and operational until June 2025, was a dedicated methane monitoring satellite designed to quantify emissions from oil and gas basins globally \cite{Rohrschneider, Wofsy_2019, ccm2026Omethanesat}. The satellite featured a pair of spectrometers with spectral resolution similar to MethaneAIR, covering the 1.27 $\mu$m O$_2$ singlet delta and 1.65 $\mu$m CH$_4$ bands. Unlike global monitoring satellites, MethaneSAT was maneuverable and targeted specific scenes at approximately the scale of oil and gas production basins (220$\times$220 km$^2$ at nadir) at a native pixel size of 110$\times$400 m$^2$ at nadir. Unlike MethaneAIR, whose XCH$_4$ maps are assembled from flight swaths acquired over multiple hours to cover oil and gas basins, MethaneSAT acquires each scene in a near-instantaneous snapshot  ($\sim$32 s). We utilize L3 gridded XCH$_4$ maps at 45 m grid size, also derived using the CO$_2$ proxy retrieval method \cite{Chan_2024,Sun_2018}. 

\begin{figure}[ht!]
    \centering
    \includegraphics[width=.98\linewidth]{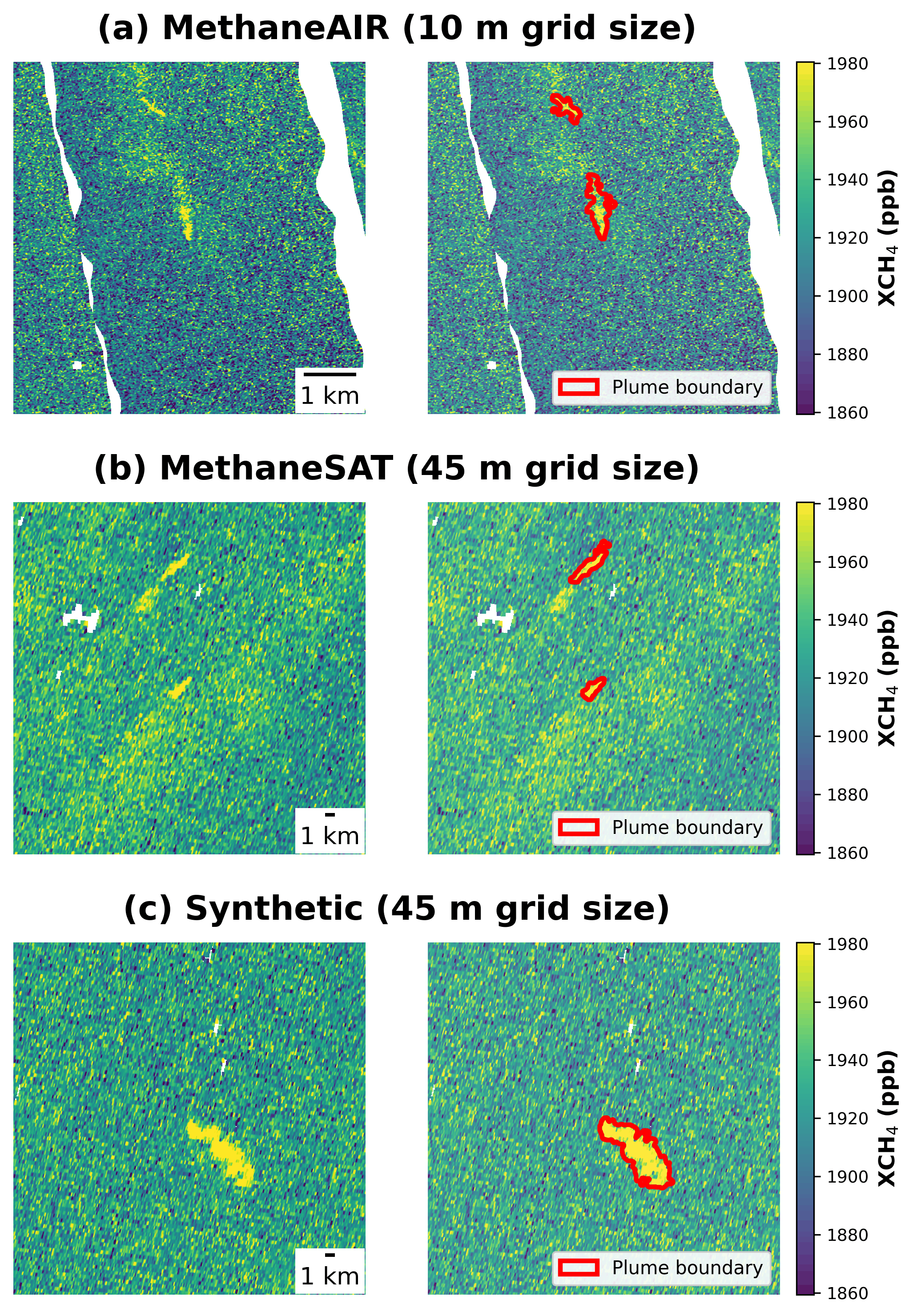}
\caption{\textbf{Methane plume detection} sample patches in the Permian Basin, each shown at its native grid size with its own scale bar; reference plume mask labels in red. (a) MethaneAIR at 10 m, covering approximately 6.7 km $\times$ 6.7 km. (b) MethaneSAT at 45 m, covering approximately 36 km $\times$ 36 km. (c) Synthetic plume injected into a MethaneSAT scene, at the same grid size and extent as (b). The panels differ in both pixel count and physical extent; each scale bar gives the physical scale of its panel.}
    \label{fig:methane_comparison}
\end{figure}

Our MethaneSAT dataset includes 27 scenes acquired between 2024 and 2025, covering  oil and gas production regions worldwide. \mpc{The 27 scenes span 13 basin labels, and the 100 reference plume mask labels are unevenly distributed across them. The Permian basin contributes both the most scenes (7) and the most reference plumes (31), followed by Appalachian-Marcellus (15 plumes in a single scene), South Caspian (14 in two scenes), Zagros Fold Belt (11 in two scenes) and Maturin (8 in two scenes). These five basins account for 79 of the 100 reference plumes, while six basins contribute three plumes or fewer. Scene-level density ranges from 1 to 15 reference plumes per scene, with a median of 3. The full inventory, with the basin of each scene and its role in every cross-validation fold, is given in Appendix~\ref{app:msat_scenes}.} These scenes were selected based on the availability of manually validated plume labels. Similar to MethaneAIR data, cloud-contaminated and low-quality soundings were masked during preprocessing. Figure \ref{fig:methane_comparison} b) shows a sample image from the MethaneSAT dataset. The coarser grid size compared to MethaneAIR (45 m vs. 10 m) presents different challenges for plume detection, particularly for distinguishing spatially proximate point sources and capturing fine-scale plume structures near emission sources.

\subsubsection{Synthetic Dataset}

To increase training diversity beyond the limited number of real scenes, we inject physically realistic synthetic plumes into MethaneSAT Level 2 retrievals using tracer fields from two distinct Weather Research and Forecasting Large Eddy Simulation (WRF-LES) frameworks. The first dataset is based on the idealized WRF-LES ensemble from Varon et al. \cite{Varon_2021}. These are 25 m by 25 m resolution large-eddy simulations over a 9$\times$9$\times$2.4 km$^3$ domain with a 3-hour duration, spanning nine scenarios with initial conditions covering a range of wind speeds and mixed-layer depths. From these simulations, we extracted 136 snapshots at 120 s temporal spacing to ensure morphological independence \cite{Ouerghi_2025}.

The second dataset consists of WRF-LES simulations derived from MethaneAIR scenes. In contrast to the idealized configurations of \cite{Varon_2018}, these simulations are driven by real meteorological fields from the HRRR model. The simulations cover a horizontal domain of 11.11 km $\times$ 11.11 km with a vertical extent of approximately 20 km, corresponding to the altitude of the 50 hPa level. A total of four 6-hour simulations were conducted over four days, based on MethaneAIR controlled-release flights in 2021 (Permian Basin) and 2022 (Arizona). This framework builds on the work of Chulakadabba et al. \cite{chulakadabba2023methane}, which coupled WRF-LES with real meteorological inputs, following the approach of \cite{lauvaux_constraining_2012}. The framework enables additional plume simulations under realistic atmospheric conditions representative of actual MethaneAIR scenes. Snapshot extraction follows the same procedure as for the first dataset.

For each selected LES snapshot, the tracer field is vertically integrated across all model levels and converted to an XCH$_4$ enhancement following the mass-conserving formulation of \cite{Varon_2021}, then linearly scaled to target emission rates sampled from a lognormal distribution fitted to observed MethaneSAT flux estimates, with the range restricted to [0.9, 4.0] t/h to avoid plume truncation at the LES domain boundaries. Each plume is rotated to align with local wind direction (combining geometric grid correction and meteorological wind components), resampled from the native LES resolution (25 m $\times$ 25 m for the first dataset and 111.11 m $\times$ 111.11 m for the second) to MethaneSAT's operational nadir resolution (400 m along-track $\times$ 110 m cross-track), and injected into L2 fields at randomly selected locations satisfying spatial non-overlap with previously injected plumes and $<$20\% bad-pixel fraction within the plume footprint. The modified L2 products are processed through standard L3 gridding \cite{Chan_2024} to produce labeled training data at 45 m resolution. 

This procedure generated 27 synthetic plumes from Varon et al. \cite{Varon_2021}, and 158 synthetic plumes from the HRRR-driven WRF-LES simulations \cite{chulakadabba2023methane} across 21 MethaneSAT scenes, providing controlled diversity while preserving realistic plume morphology, sensor-specific spatial sampling patterns, atmospheric variability, and instrumental noise characteristics. A sample image from the HRRR-driven dataset is shown in Figure \ref{fig:methane_comparison} c).

\subsection{Plume Mask Labels}
\label{sec:labels}

\mpc{Reference plume mask labels} for training and validating our models were generated using the wavelet transform-based detection method of \cite{zhang_2026}, which achieves 60\% higher plume detection than the thresholding-based approach of \cite{Warren_2025} on MethaneSAT scenes. Figure \ref{fig:methane_comparison} shows sample patches from the XCH$_4$ maps with corresponding plume masks.

The method applies 2D discrete wavelet transforms to XCH$_4$ imagery through three steps: denoising via multilevel wavelet decomposition to suppress background and enhance plume signals, adaptive thresholding with connected component analysis to generate binary masks, and filtering based on concentration hotspot presence, plume shape consistency, and wind direction alignment against HRRR/GFS meteorological data within a $\pm 55^\circ$ wind direction tolerance (see \cite{zhang_2026} for full details).

Applied to our dataset, the wavelet method produced labels for 1050 plumes across 84 MethaneAIR scenes and 100 plumes across 27 MethaneSAT scenes. \mpc{All automatically generated masks then passed three sequential rounds of visual quality control: a first review at Harvard University, a second review by the MethaneSAT team at the Environmental Defense Fund before the plumes were reported, and a final review by the first author, who also selected the plumes used for model training. Each round verified plume boundaries, removed residual false detections, and flagged ambiguous cases such as overlapping plumes or dispersed enhancements lacking a clear source hotspot.  Ambiguous detections were excluded from training to avoid introducing label noise.}

\mpc{The resulting masks are \emph{reference plume mask labels} derived from the wavelet method with manual quality control. They are not fully independent ground truth, so our evaluation partly measures agreement with the wavelet reference. The bias of these labels has a known direction: they are conservative. Both the wavelet criteria (hotspot presence, wind alignment within $\pm 55^\circ$, shape consistency) and the spatial extent of the masks are chosen to keep false detections and flux bias low; for MethaneSAT, the masks cover only the plume segment within roughly 5--10~km of the source and are not extended farther downwind \cite{zhang_2026}. As a result, parts of the downwind tail lie outside the reference masks, and, as our false-positive review shows (Section~\ref{sec:fp_analysis}), some real enhancements can be missed from the reference labels entirely.}

\section{Methodology}

\begin{figure*}[ht!]
  \centering
\includegraphics[width=.98\linewidth]{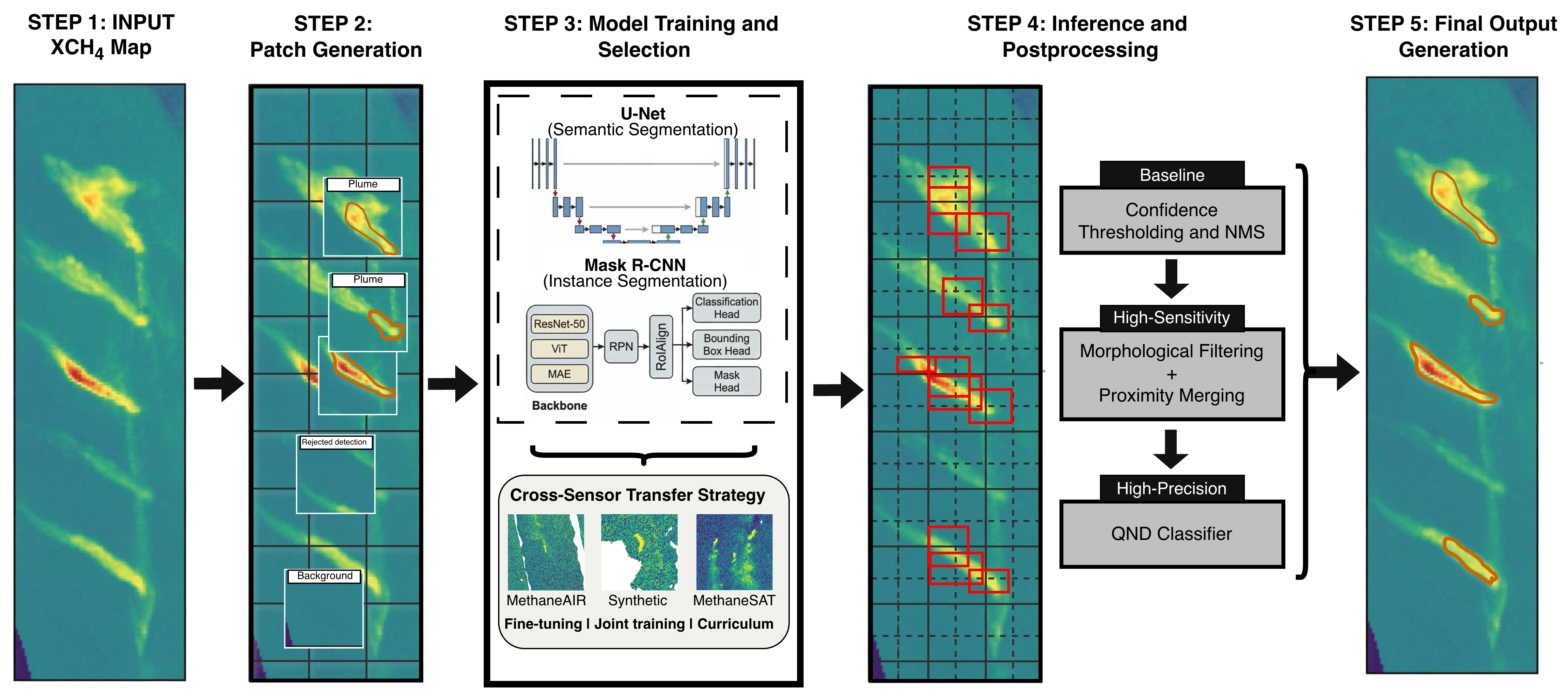}
\caption{Overview of the proposed methane plume detection pipeline. (Step 1) Input XCH$_4$ maps are ingested from MethaneAIR or MethaneSAT retrievals. (Step 2) Patches are extracted, balancing annotated plume instances with background, rejected detections, and hard negative examples. (Step 3) Two model families are trained and compared: a U-Net semantic segmentation baseline and Mask R-CNN instance segmentation with interchangeable ResNet-50, ViT, and MAE backbones; cross-sensor transfer from MethaneAIR and synthetic data to MethaneSAT is achieved via fine-tuning, joint training, or curriculum learning. (Step 4) Full-scene inference applies a sliding window strategy followed by confidence thresholding and NMS, morphological filtering, proximity-based merging, and QND-based classification, yielding the high-precision/sensitivity modes (Step 5).} 
  \label{fig:method}
\end{figure*}

Our approach for automated methane plume detection is depicted in Figure~\ref{fig:method}. From the input XCH$_4$ maps, we preprocess the data to generate training patches, balancing plume instances with background and hard negative examples (Section~\ref{sec:data_prep}). We then conduct a backbone trade study comparing Mask R-CNN with ResNet-50, ViT, and MAE encoders (Section~\ref{sec:instance_seg}) against a U-Net semantic segmentation baseline (Section~\ref{sec:semantic_seg}), with predictions converted to a common representation for fair comparison (Section~\ref{sec:instance_to_semantic}). We pair these models with a cross-sensor transfer strategy that leverages MethaneAIR scenes and simulated plumes, evaluating fine-tuning, joint training, and curriculum learning (Section~\ref{sec:transfer}). Finally, we apply the trained model to full scenes through a sliding window inference pipeline with post-processing (Section~\ref{sec:inference}), producing a high-precision mode optimized for confident source attribution and a high-sensitivity mode that retains marginal detections. As an additional output, we aggregate overlapping patch predictions into a continuous confidence-weighted detection map $\hat{P}$ that encodes spatial detection confidence and provides rich spatial information for downstream emission quantification (Section~\ref{sec:prob_maps}). Evaluation metrics for both pixel-level and instance-level performance are defined in Section~\ref{sec:metrics}.

\subsection{Data Preparation}
\label{sec:data_prep}

Prior to training, XCH$_4$ scenes underwent z-score normalization to account for varying concentration ranges across different acquisition conditions. Invalid pixels were set to $-10$ after applying normalization, a value chosen to lie well outside the physically plausible range. From each scene, we extract fixed-size patches of 800 $\times$ 800 pixels centered on each annotated plume origin, ensuring plume morphology is captured within the patch boundaries.

\mpc{Patches centered on plume origins provide only positive examples. To teach the detector to reject structures that resemble plumes but are not, we add for each instrument an approximately equal number of negative patches that contain no reference plume (Table~\ref{tab:dataset}).
Negative examples fall into three groups. \emph{Background} patches are randomly sampled regions containing no methane plumes. \emph{Rejected detections} are wavelet candidates discarded as false detections during manual review. \emph{Hard negatives}, used for MethaneSAT only, are patches containing instrument-specific false-positive patterns identified from preliminary validation experiments, in three categories: \emph{striping artifacts} (linear stripes from instrumental noise); \emph{dispersed enhancements} (spatially extended elevated XCH$_4$ with no concentration hotspot and no coherent wind-aligned shape); and \emph{small enhancements} (compact anomalies near the detection limit without a downwind tail).}

\mpc{Dispersed enhancements are distinguished from real plumes of low enhancement by the same criteria used for labeling (Sec.~\ref{sec:labels}): a plume, however faint, retains a local hotspot and a shape aligned with the wind direction within $\pm 55^\circ$, whereas a dispersed enhancement lacks both, regardless of magnitude. Because the review rounds were sequential, with each reviewer seeing the previous round's result, they are not independent annotations and inter-annotator agreement cannot be computed. The negative-sample categorization was performed in the final round by a single annotator following these criteria; we acknowledge this as a limitation and note that the categories only shape the composition of negative training samples and do not enter any evaluation metric. For MethaneSAT, background patches are replaced by hard negatives, since background regions are repetitive whereas hard negatives carry richer, more informative contrast.} The counts for each category are summarized in Table~\ref{tab:dataset}.

Data augmentation is applied on-the-fly during training, consisting of random rotations (90°, 180°, or 270°) and random horizontal and vertical flips (each with probability 0.5). Patches are randomly cropped to 768 $\times$ 768 pixels for MethaneSAT and 448 $\times$ 448 pixels for MethaneAIR. These crop sizes were selected based on validation performance, which we analyze in Appendix~\ref{app:imagesize}. \mpc{Rotations are restricted to multiples of $90^\circ$ because these are exact on the pixel grid: they rearrange pixels without interpolating the XCH$_4$ field or the invalid-pixel mask, while rotation by an arbitrary angle resamples both and fills the patch corners with artificial values. As a check, we retrained the MethaneSAT-only configuration with rotation angles drawn uniformly from [0°, 360°) on the same splits. The patch-level F1 was $70.23 \pm 4.94$, against $70.51 \pm 5.50$ with 90° rotations, a difference well within one cross-validation standard deviation.} 

Each of our datasets was partitioned into training, validation, and test sets at the scene level, such that all patches extracted from a given scene are assigned exclusively to one split, preventing information leakage between sets. Specifically, 20\% of scenes were held out as the final test set. On the remaining 80\%, we employ a 3-fold cross-validation strategy to ensure models are evaluated across diverse atmospheric conditions and geographic regions. All results reported in this paper are computed on the test set and expressed as the mean over the three cross-validation folds. 


\begin{table}[h!]
\centering
\begin{tabular}{llc}
\toprule
\textbf{Instrument} & \textbf{Sample Type} & \textbf{Samples} \\
\midrule
MethaneAIR   & Plume                & 1,050 \\
(84 scenes)  & Rejected detection   & 142   \\
             & Background           & 908   \\
             & \textit{Subtotal}    & 2,100 \\
\midrule
MethaneSAT   & Plume              &  100   \\
(27 scenes)  & Rejected detection &  38    \\
             & Small enhancements &  37    \\
             & Striping artifacts &  35    \\
             & Dispersed enhancements   &  15    \\
             & \textit{Subtotal}  &  225   \\
\midrule
Synthetic    & Idealized WRF-LES \cite{Varon_2021}   & 27   \\
(21 scenes)  & HRRR-driven WRF-LES \cite{chulakadabba2023methane} & 158   \\

             & \textit{Subtotal}  &   185   \\
             
\midrule
\textbf{Total} &                 & 2,510 \\
\bottomrule
\end{tabular}
\caption{Labeled patch dataset composition by instrument and sample type, before the scene-level train/validation/test split. Plumes are wavelet-detected instances that passed manual quality control. Rejected detections are wavelet candidates discarded during manual review as false positives. MethaneSAT hard negatives comprise three instrument-specific artifact categories. Synthetic plumes are WRF-LES tracer fields injected into real MethaneSAT L2 retrievals: idealized simulations~\cite{Varon_2021} and meteorologically realistic simulations driven by HRRR~\cite{chulakadabba2023methane}.}

\label{tab:dataset}
\end{table}

\subsection{Semantic Segmentation Model: U-Net}
\label{sec:semantic_seg}

For semantic segmentation, we employ a U-Net architecture \cite{Ronneberger2015} adapted for single-channel XCH$_4$ input. The network follows the standard encoder-decoder design with skip connections. Specifically, we use three contracting blocks of (32, 64, and 128) channel sizes across the encoder, and three expanding blocks producing output feature maps of (128, 64, 32) channels at each stage, with skip connection concatenation doubling the input channel count at each expanding block. Each contracting block applies two convolution-BatchNorm-ReLU \cite{agarap_2018} operations followed by max pooling with stride 2, and each expanding block upsamples with a transposed convolution before concatenating the corresponding skip connection.


The network is trained with a weighted cross-entropy loss to address the severe class imbalance between plume and background pixels:

\begin{equation}
\mathcal{L}_{\text{seg}} = -\sum_{c \in \{0,1\}} w_c \sum_{i} y_{ic} \log \hat{y}_{ic}
\end{equation}

\noindent where $w_c$ are class weights inversely proportional to pixel frequency, $y_{ic}$ is the \mpc{reference plume mask label} of pixel $i$ for class $c$, and $\hat{y}_{ic}$ is the predicted probability. The final layer produces a two-channel output and class probabilities are obtained via softmax. We use a threshold of 0.5 on the class probabilities to generate discrete plume categories for evaluation.

Optimization uses mini-batches of size 32, and AdamW \cite{loshchilov_2017} with a learning rate of $10^{-4}$, selected through cross-validation on the training folds. Further hyperparameter details are provided in Appendix~\ref{app:hyperparams}.

\subsection{Instance Segmentation Model: Mask-RCNN}
\label{sec:instance_seg}

We employ Mask R-CNN \cite{he2017mask} for instance segmentation, extending Faster R-CNN \cite{Ren_2015} with a mask prediction branch parallel to the bounding box regression and classification branches. This architecture enables simultaneous detection and pixel-wise masking of individual plume instances within a single forward pass.

\begin{figure}[ht!]
    \centering
    \includegraphics[width=0.8\columnwidth]{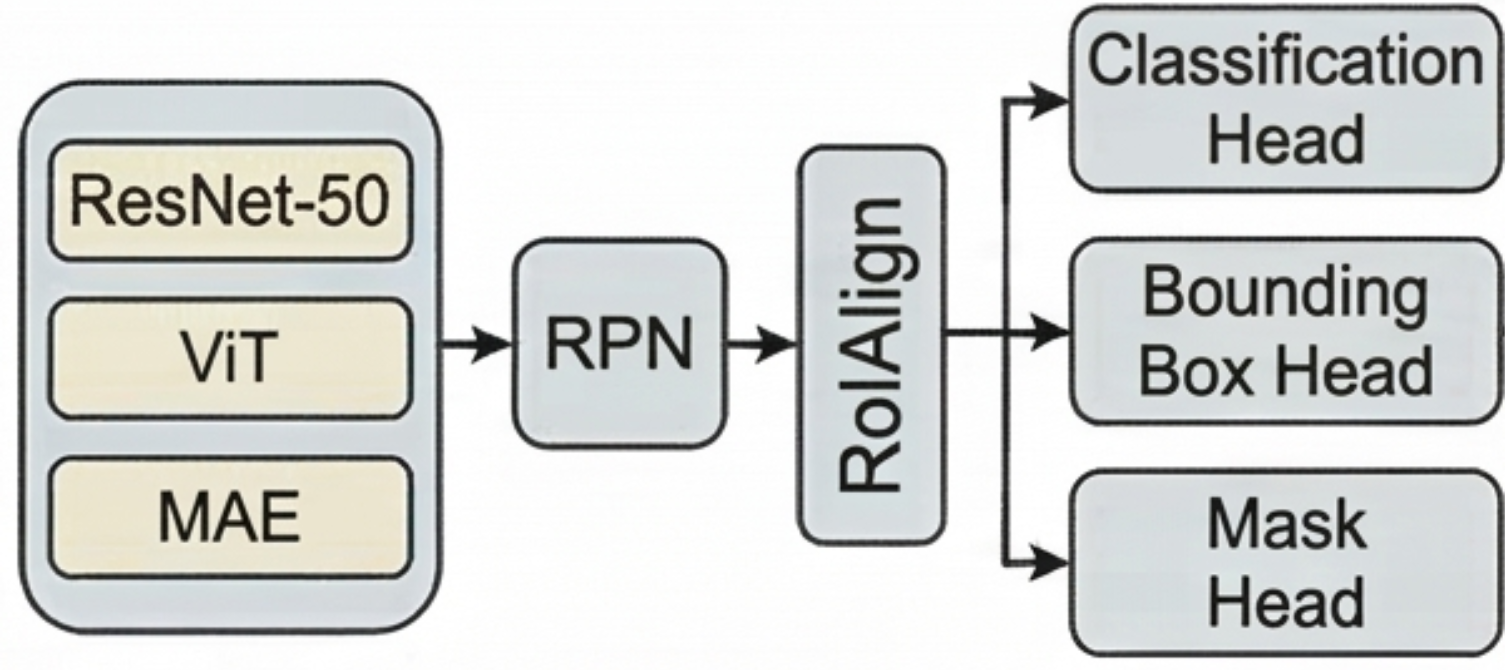}
    \caption{\textbf{Mask R-CNN architecture used in this work.} The backbone (ResNet-50, ViT, or MAE) extracts feature maps from the input XCH$_4$ patch. A Region Proposal Network (RPN) generates candidate plume regions, which are resampled to a fixed spatial size via RoIAlign and passed to three parallel heads: a classification head that predicts plume vs.\ background, a bounding box head that refines region coordinates, and a mask head that predicts a binary segmentation mask per region. The three heads are trained jointly with the multi-task loss in Eq.~\ref{eq:loss_instance}.}
    \label{fig:maskrcnn_arch}
\end{figure}

Mask R-CNN operates in two stages. The first stage uses a Region Proposal Network (RPN) to generate candidate bounding boxes. The second stage extracts features from each proposed region using a RoIAlign module, which preserves spatial alignment between the input and the extracted features, while resizing images to a common size. These features are processed by three parallel branches: a classification head that determines whether the region contains a plume, a bounding box regression head that refines the region coordinates, and a mask head that predicts a binary segmentation mask through a small fully convolutional network \cite{Shelhamer_2014}. Figure~\ref{fig:maskrcnn_arch} summarizes the full architecture.

The model is trained with a multi-task loss that jointly optimizes all four objectives:

\begin{equation}
\label{eq:loss_instance}
\mathcal{L}_{\text{instance}} = 
\mathcal{L}_{\text{cls}} +
\mathcal{L}_{\text{bbox}} +
\mathcal{L}_{\text{mask}} +
\mathcal{L}_{\text{RPN}} 
\end{equation}

\noindent where $\mathcal{L}_{\text{cls}}$ is the cross-entropy loss for plume classification, $\mathcal{L}_{\text{bbox}}$ is the smooth L1 loss for bounding box regression, $\mathcal{L}_{\text{mask}}$ is the binary cross-entropy loss computed per-pixel on the predicted mask, and $\mathcal{L}_{\text{RPN}}$ combines classification and regression losses for the region proposal stage. Optimization uses mini-batches of size 32 and AdamW with a learning rate of $10^{-4}$, consistent with the U-Net baseline.

We investigate three backbone architectures to assess the role of feature representation and initialization strategy in detection performance. All backbones except MAE, which uses domain-specific self-supervised pre-training as described below, are trained from scratch, as ImageNet weights are optimized for 3-channel RGB inputs and do not transfer directly to single-channel XCH$_4$ maps with a fundamentally different value distribution.

\textbf{ResNet-50} \cite{He_2015} is a 50-layer residual convolutional network. Its hierarchical feature extraction and residual connections provide a strong convolutional baseline for plume detection, and its relatively compact parameter count makes it well-suited to training from scratch on a limited labeled dataset.

\textbf{Vision Transformer (ViT)} \cite{Dosovitskiy_2020} divides the input into non-overlapping patches and processes them through transformer encoder layers. Its self-attention mechanism \cite{vaswani_2017}  can model long-range spatial dependencies that convolutional encoders may miss, which is relevant for elongated plume structures whose spatial extent can span hundreds of pixels across a scene.

\textbf{Masked Autoencoder (MAE)} \cite{He_2021} is a ViT backbone pre-trained on XCH$_4$ data using self-supervised reconstruction. During pre-training, a random fraction of input patches is masked (masking ratio 0.5), and the model learns to reconstruct the original XCH$_4$ signal from the visible patches. This strategy is motivated by the data scarcity challenge: by leveraging the full unlabeled MethaneSAT archive rather than only the annotated subset, MAE pre-training may learn spatial representations of XCH$_4$ fields more relevant to plume detection than random initialization, without requiring additional labeled data.

\subsection{Instance-to-Semantic Conversion}
\label{sec:instance_to_semantic}

To enable direct comparison between instance and semantic segmentation approaches at evaluation, we convert Mask R-CNN instance predictions to binary semantic segmentation maps. For each scene, individual instance masks are aggregated through a union operation:

\begin{equation}
\mathbf{S} = \bigcup_{i=1}^{N} \mathbf{M}_i
\end{equation}

where $\mathbf{S} \in \{0,1\}^{H \times W}$ represents the resulting semantic segmentation map, and $\mathbf{M}_i$ are the individual instance masks. This conversion allows us to evaluate both approaches using identical semantic segmentation metrics while maintaining the instance-level detection capabilities of Mask R-CNN for downstream emission quantification tasks that require associating plumes with specific sources.

\subsection{Cross-Sensor Transfer Strategy}
\label{sec:transfer}

MethaneAIR and MethaneSAT XCH$_4$ maps are generated by the same retrieval algorithms, but differ in grid size, atmospheric path length, and signal-to-noise characteristics. These domain differences motivate a structured transfer strategy rather than direct training on the limited MethaneSAT labeled set.

\textbf{Stage 1 — MethaneAIR pre-training.} Models are first trained on MethaneAIR, whose 84 annotated scenes provide substantially more labeled plume examples than the MethaneSAT archive. This stage allows the model to learn fundamental plume characteristics: typical morphologies and orientations governed by wind direction, concentration gradients from source to background, and contextual patterns that distinguish real plumes from retrieval artifacts.

\textbf{Stage 2 — Adaptation to MethaneSAT.} Starting from the MethaneAIR-pretrained weights, we investigate three strategies and one control for adapting the model to the MethaneSAT domain, each making different use of the synthetic plume dataset described in Section~\ref{sec:data}:

\textit{Fine-tuning:} continues training exclusively on real MethaneSAT labeled scenes, adapting all weights to satellite-specific observational characteristics.

\textit{Joint training:} trains simultaneously on real MethaneSAT scenes and synthetic plumes, exposing the model to broader emission rate and meteorological diversity.

\textit{Curriculum learning:} the real MethaneSAT sampling fraction increases linearly from 25\% at epoch 0 to 100\% at epoch 50, while the synthetic fraction decreases correspondingly from 75\% to 0\%. This schedule prioritizes synthetic data early in training, and progressively emphasizes real MethaneSAT data as training proceeds.

\mpc{\textit{Synthetic-only:} trains on the synthetic plumes only, from the same MethaneAIR initialization, with no real MethaneSAT plume labels. This configuration is not an operational candidate. It shows what the MethaneAIR initialization and synthetic supervision reach together without real plume labels.}

Together, these strategies span the space from straightforward supervised adaptation to alignment on synthetic scenes, allowing us to assess whether synthetic data diversity provides greater benefit beyond simple fine-tuning from MethaneAIR. Results for domain-adversarial adaptation (DANN) \cite{ganin_2016} as an additional feature-level transfer strategy are provided in Appendix~\ref{app:domain_adaptation}.

\subsection{Scene-Level Inference and Post-Processing}
\label{sec:inference}

At inference time, the trained model is applied to full XCH$_4$ scenes that far exceed the training patch dimensions. We address this through a sliding window strategy followed by a post-processing pipeline that reconstructs coherent scene-level predictions from overlapping patch detections and produces two operationally distinct modes.

\textbf{Sliding window inference.} Given a full scene $I$ of dimensions $H \times W$, we partition it into overlapping patches of size $s$ (as defined in Section~\ref{sec:data_prep}) with overlap ratio $\alpha = 0.75$. This substantial overlap ensures that plumes near patch boundaries are fully captured by at least one window, reducing edge artifacts in the aggregated predictions. All predicted masks and bounding boxes are mapped back to their corresponding coordinates in the original scene after each forward pass. \mpc{The overlapping tiles serve only to produce predictions; scene-level metrics (Section~\ref{sec:results_scene}) are computed once, on the single reconstructed full-scene prediction obtained after the post-processing stages of the respective operating mode. Overlapping tiles are never evaluated as separate samples.}

\textbf{Baseline: confidence filtering and NMS.} Per-patch detections with confidence scores below a threshold $\tau$ are discarded, and non-maximum suppression (NMS) is applied with an IoU threshold $\delta$ to remove redundant detections of the same plume instance, retaining only the highest-confidence prediction among overlapping candidates. Both $\tau$ and $\delta$ were selected by cross-validation on the validation set (see Appendix~\ref{app:thresholds} for details). The resulting detections constitute our baseline scene-level predictions, against which the two operational modes are evaluated in Section~\ref{sec:results_scene}.

\textbf{High-sensitivity mode.} Two additional post-processing stages are applied on top of the baseline to remove physically implausible predictions while retaining marginal detections that may represent real low-flux sources.

\textit{Morphological filtering.} We apply a physics-informed shape filter to reject predicted masks whose geometry is inconsistent with real methane plume morphology, following the criterion introduced in the wavelet-based labeling method of Zhang et al.~\cite{zhang_2026}. The filter targets masks with a branching or ``spider'' topology, where multiple curling arms extend from a central region, arising from retrieval artifacts that exceed the model's confidence threshold but lack coherent plume structure. \mpc{The physical basis of this filter is advection: a plume from a single source in a single overpass is transported by one local wind field and therefore forms an elongated shape along the downwind axis. A mask with several curling arms would require several simultaneous transport directions, which is inconsistent with a single wind field.} To quantify this topology, we compute the fiber length of each predicted mask via skeletonization, defined as the length of the longest path along the mask's one-pixel-wide skeleton. For a plume-like mask, the skeleton traces a roughly linear downwind axis, so the fiber length and the major axis length of the bounding ellipse are close in magnitude. For a spider-shaped mask, curling branches inflate the fiber length well beyond the major axis, producing a high fiber-to-major-axis ratio. Following Zhang et al.~\cite{zhang_2026}, we discard predictions whose fiber-to-major-axis ratio exceeds $1.25$, a threshold they selected by sensitivity analysis across MethaneSAT scenes with varying surface features and wind speeds.

\textit{Proximity-based merging.} Predictions whose masks intersect across patch boundaries are grouped and merged into a single instance: we compute the union of their bounding boxes and masks and assign an area-weighted average confidence score to the merged detection.

Together, these two stages define the \textbf{high-sensitivity mode}, which preserves marginal detections near the instrument detection limit for longitudinal monitoring across the MethaneSAT archive. A source that appears marginal in a single scene may be confirmed by repeated detections across multiple overpasses.

\textbf{High-precision mode.} Some operational applications, including regulatory reporting and confident source attribution, require minimizing false positives even at the cost of recall. To support these use cases, we add a third stage on top of the high-sensitivity mode: the distribution-based QND classifier~\cite{Ottenheimer2026}. For each remaining detection, the classifier extracts percentile-based polynomial descriptors from the within-mask distributions of XCH$_4$ and surface albedo, and combines them with geometric plume features in a random forest to discriminate real methane enhancements from spurious artifact-driven detections (see Appendix~\ref{app:rfc} for details). \mpc{The QND descriptors are physically motivated rather than learned from raw imagery. A real methane enhancement is a column-abundance signal independent of surface reflectance, whereas many false detections (albedo variability, topographic shadows, cloud brightness) are retrieval biases that co-vary with albedo; within a candidate mask, a real plume therefore shows XCH$_4$ elevated across the full range of albedo values, while such artifacts show XCH$_4$ tracking albedo. The QND polynomials encode this through the shape of the within-mask XCH$_4$ and albedo distributions, which also separates diffuse enhancements and noise-driven candidates from plumes~\cite{Ottenheimer2026}. Since these signatures derive from the CO$_2$-proxy retrieval physics shared by MethaneAIR and MethaneSAT rather than from instrument-specific noise, the classifier trained on MethaneAIR is applied to MethaneSAT without retraining.} Predictions classified as artifacts are discarded, producing a set of detections optimized for confident source attribution at the cost of potentially removing marginal low-flux sources. We compare this approach against morphological size thresholding as a simpler alternative baseline in Appendix~\ref{app:size_threshold_vs_QND}.

The trade-off between precision and sensitivity across the baseline, high-sensitivity, and high-precision modes is characterized quantitatively in Section~\ref{sec:results_scene}.

\subsection{Confidence-Weighted Detection Maps}
\label{sec:prob_maps}

In addition to the discrete plume detections produced by the post-processing pipeline, the sliding window inference stage yields a continuous scene-level confidence-weighted detection map as a byproduct of aggregating overlapping patch predictions. For each pixel $p$ in the full scene, let $\mathcal{D}_p = \{(s_k, \mathbf{M}_k)\}$ denote the set of detections with confidence above $\tau$ in all sliding windows that contain $p$, where $s_k \in [0,1]$ is the classification confidence score of detection $k$ and $\mathbf{M}_k(p) \in [0,1]$ is the soft mask probability assigned to pixel $p$ by detection $k$ (zero outside its mask). The map is defined as:

\begin{equation}
\hat{P}(p) = \dfrac{\displaystyle\sum_{k \in \mathcal{D}_p} s_k\, \mathbf{M}_k(p)}{\displaystyle\sum_{k \in \mathcal{D}_p} s_k} 
\end{equation}

\noindent where the denominator accumulates the confidence of all detections in the windows containing $p$; $\hat{P}(p)=0$ when $D_p$ is empty. A pixel receives a nonzero value only if at least one detection claims it. A high value marks a pixel that the detections in those windows consistently assign to a plume; the value is lower when the pixel is claimed in only some windows, or when the same windows also contain other detections. \mpc{$\hat{P}$ is a confidence-weighted aggregate of detector outputs, not a calibrated probability of methane presence; we present it as an exploratory screening product and spatial prior, and defer calibration against independent data to future work (Appendix~\ref{app:prob_maps}).}

\subsection{Evaluation Metrics}
\label{sec:metrics}

We evaluate model performance at two levels of granularity corresponding to the two operational objectives of the system: pixel-level segmentation quality and instance-level detection performance.

\textbf{Pixel-level metrics.} For patch-level comparisons between semantic and instance segmentation approaches, we report macro-averaged precision, recall, and F1 score computed against binary reference plume masks. Macro averaging gives equal weight to the plume and background classes. Pixel accuracy is omitted because the severe class imbalance between plume and background pixels can yield high accuracy even when plume segmentation is poor. For each class $c \in \{\text{plume}, \text{background}\}$, precision, recall, and F1 are defined as:

\begin{equation}
P_c = \frac{TP_{c}}{TP_{c} + FP_{c}},
\qquad
R_c = \frac{TP_{c}}{TP_{c} + FN_{c}},
\end{equation}

\begin{equation}
F1_c =
\frac{2TP_{c}}
{2TP_{c} + FP_{c} + FN_{c}},
\end{equation}

\noindent where $TP_{c}$, $FP_{c}$, and $FN_{c}$ denote the true positive, false positive, and false negative pixel counts for class $c$, respectively, treating that class as positive and the other class as negative. The reported macro-averaged metrics are:

\begin{equation}
\text{Precision} = \frac{1}{2}\sum_c P_c,
\qquad
\text{Recall} = \frac{1}{2}\sum_c R_c,
\end{equation}
\begin{equation}
\text{F1} = \frac{1}{2}\sum_c F1_c.
\end{equation}

Macro F1 is therefore the average of the class-specific F1 scores, rather than the harmonic mean of macro precision and macro recall. For Mask R-CNN, instance predictions are converted to a binary semantic map via the union operation described in Section~\ref{sec:instance_to_semantic} prior to computing pixel-level metrics, allowing comparison with U-Net outputs using the same evaluation procedure.

\textbf{Instance-level metrics.} For scene-level evaluation, we shift to instance-level detection metrics that directly reflect the operational objective of identifying and masking individual emission sources. A predicted instance is counted as a true positive if its mask overlaps a \mpc{reference plume instance} with an IoU exceeding $\theta$; predictions with no sufficiently overlapping \mpc{reference instance} are counted as false positives, and \mpc{reference instances} with no sufficiently overlapping prediction are counted as false negatives. \mpc{Matching is greedy and strictly one-to-one: predictions are processed in descending confidence order, each prediction may match at most one reference instance, and each reference instance may be claimed by at most one prediction; a single prediction can therefore never be counted as more than one true positive.} Precision, recall, and F1 score are computed from these instance-level TP, FP, and FN counts as defined above.

We additionally report mean Average Precision (mAP) at IoU threshold $\theta$, defined as the area under the precision--recall curve obtained by varying the model confidence threshold:

\begin{equation}
\text{mAP}_{\theta} = \int_0^1 p(r)\, dr
\end{equation}

\noindent where $p(r)$ is the precision at recall level $r$, computed at IoU $\theta$. 

\mpc{Three conventions apply to all reported results. First, all metrics are computed against the reference labels of Section~\ref{sec:labels}. Because these labels are conservative, real plumes outside them count as false positives, so precision likely underestimates detection performance (see Sec.~\ref{sec:fp_analysis}). Second, reported $\pm$ values denote the standard deviation across the three scene-exclusive cross-validation folds (i.e., across independently trained models), not across patches or pixels; since scenes are assigned exclusively to one split, spatial autocorrelation between neighboring patches cannot leak across the train/validation/test boundary. Because the test scenes are the same in all folds, these values measure sensitivity to the training/validation split, not the sampling uncertainty of the test set. Third, the learning rate, patch size, $\tau$, and $\delta$ were selected on the validation folds only (Appendix~\ref{app:hyperparams}--\ref{app:thresholds}); the morphological threshold is taken from Zhang et al. \cite{zhang_2026} and was not tuned in this work; the five test scenes were never used for any selection in this work.}

\section{Results} \label{sec:results}

\subsection{Instance vs. Semantic Segmentation}
\label{sec:results_baseline}

We compare U-Net against three Mask R-CNN backbone variants on both MethaneAIR and MethaneSAT. All models are evaluated on patches of $448\times448$ pixels for MethaneAIR and $768
\times768$ pixels for MethaneSAT, cropped from $800\times800$ pixel extractions centered on annotated plume origins; full results of this trade-off analysis are provided in Appendix~\ref{app:imagesize}. The goal is to determine \mpc{whether Mask R-CNN offers a performance advantage over the U-Net baseline} when both are evaluated under identical pixel-level metrics, and to select the strongest backbone for the experiments that follow. 

\subsubsection{MethaneAIR}

Results on the MethaneAIR test set are presented in Table~\ref{tab:mair_results}. U-Net achieves the highest recall (92.99\%) among all models but the lowest precision (59.19\%), indicating that the model successfully detects the spatial extent of most plumes while generating a substantial number of false positive pixels at plume boundaries and in regions of dispersed enhancement. Mask R-CNN with a ResNet-50 backbone improves precision by 7.87 percentage points over U-Net while maintaining competitive recall (88.97\%), yielding the highest overall F1 score of 74.90\%. The instance-level formulation enforces a detection confidence threshold before committing pixels to a plume mask, which reduces spurious activations in low-enhancement regions and produces tighter boundary masking.

\begin{figure*}[ht!]
  \centering
\includegraphics[width=.98\linewidth]{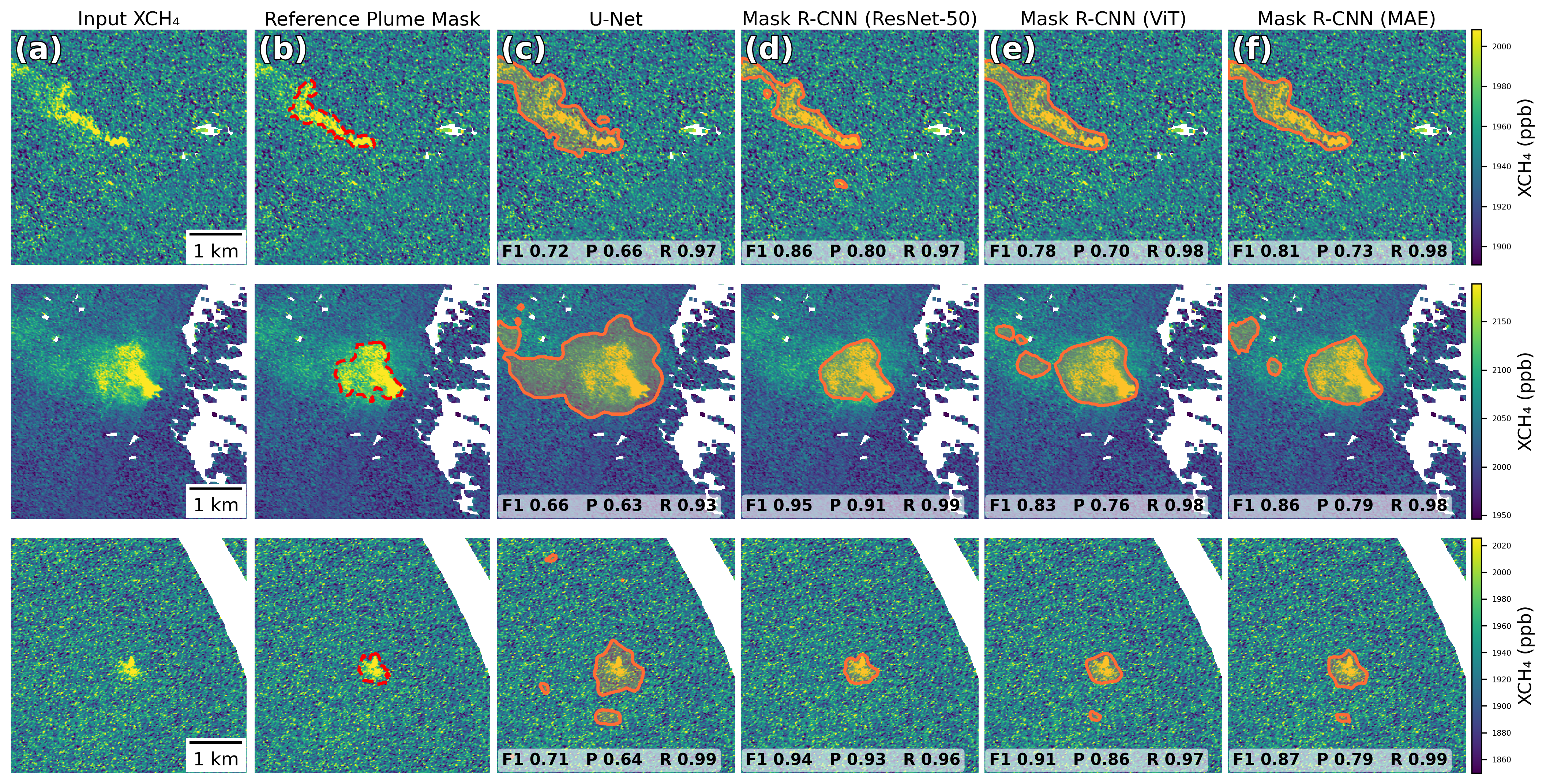}
\caption{Qualitative comparison of plume segmentation predictions   on three MethaneAIR test scenes at patch size 448$\times$448. Column (a) shows the input XCH$_4$ map; column (b) shows the \mpc{reference plume mask labels}; columns (c)--(f) show predictions from U-Net, Mask R-CNN (ResNet-50), Mask R-CNN (ViT), and Mask R-CNN (MAE) respectively. Predicted masks are shown as filled semi-transparent overlays with solid contour boundaries; per-panel metrics (F1, precision P, recall R) are computed at pixel level against the \mpc{reference plume mask labels}. Metrics are computed for the illustrated examples only and are not directly comparable to aggregate test set results in Table~\ref{tab:mair_results}.} 
  \label{fig:mair_comparison_semantic}
\end{figure*}

Mask R-CNN with ViT and MAE backbones lands between U-Net and ResNet-50, with nearly identical F1 (67.33\% and 67.32\%). ViT precision (64.83\%) is marginally above U-Net but its recall (84.57\%) is the lowest of the four; MAE recovers recall (88.43\%) at lower precision (61.52\%). Neither improves on ResNet-50, and MAE does not improve on the randomly initialized ViT despite its self-supervised pre-training on XCH$_4$ data, suggesting that the long-range context modeling of ViT encoders needs more labeled examples than are available here. 

Figure~\ref{fig:mair_comparison_semantic} shows qualitative predictions for three representative plume examples. All models recover the main plume body with comparable overall performance, though U-Net tends to overpredict boundaries. Differences become more pronounced for the high-flux dispersed enhancement (row 2), where Mask R-CNN (ResNet-50) achieves a pixel-level F1 of 0.95 with precision of 0.91 by making a tight mask closely following the plume axis, while U-Net produces a broader prediction (F1 0.66, P 0.63) with additional spurious activations visible upwind of the source.

\begin{table}[h]
\centering
\begin{tabular}{lccc}
\toprule
\textbf{Model} & \textbf{F1} & \textbf{Precision} & \textbf{Recall} \\
\midrule
U-Net & 
64.41\scriptsize{$\pm$1.09} & 59.19\scriptsize{$\pm$0.78} & \textbf{92.99\scriptsize{$\pm$1.06}} \\
Mask R-CNN (ResNet) & 
\textbf{74.90\scriptsize{$\pm$1.57}} & \textbf{67.06\scriptsize{$\pm$1.57}} & 88.97\scriptsize{$\pm$0.62} \\
Mask R-CNN (ViT)    & 
67.33\scriptsize{$\pm$2.63} & 64.83\scriptsize{$\pm$1.62} & 84.57\scriptsize{$\pm$1.14} \\
Mask R-CNN (MAE)  &  
67.32\scriptsize{$\pm$0.83} & 61.52\scriptsize{$\pm$0.65} & 88.43\scriptsize{$\pm$0.86} \\
\bottomrule
\end{tabular}
\caption{Patch-level performance on the MethaneAIR test set at patch size 448$\times$448. Instance segmentation predictions are converted to semantic maps 
for fair comparison. Results are mean $\pm$ standard deviation 
over three cross-validation folds.}
\label{tab:mair_results}
\end{table}

\subsubsection{MethaneSAT}

\begin{figure*}[ht!]
  \centering
\includegraphics[width=.98\linewidth]{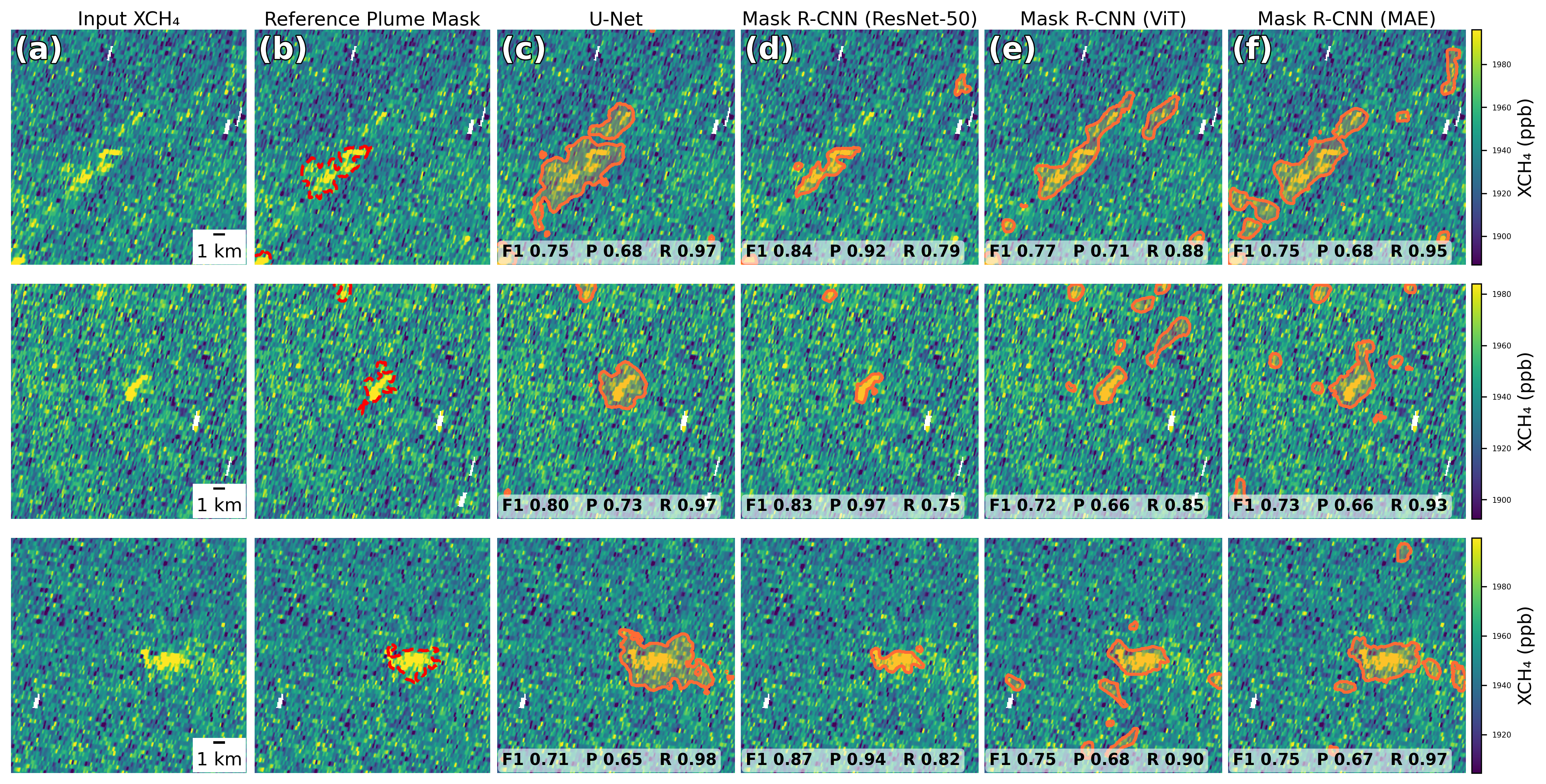}
\caption{Qualitative comparison of plume segmentation predictions on three MethaneSAT test scenes at patch size 768$\times$768. Column (a) shows the input XCH$_4$ map; column (b) shows the \mpc{reference plume mask labels}; columns (c)--(f) show predictions from U-Net, Mask R-CNN (ResNet-50), Mask R-CNN (ViT), and Mask R-CNN (MAE) respectively. Predicted masks are shown as filled semi-transparent overlays with solid contour boundaries; per-panel metrics (F1, precision P, recall R) are computed at pixel level against the \mpc{reference plume mask labels}.} 
  \label{fig:msat_comparison_semantic_msat}
\end{figure*}

Results on the MethaneSAT test set are presented in Table~\ref{tab:msat_results} and qualitative predictions for three representative scenes are shown in Figure~\ref{fig:msat_comparison_semantic_msat}.  The overall ranking of architectures partially replicates the MethaneAIR findings, but the MethaneSAT setting reveals a more fragmented prediction distribution and a distinct pattern of precision-recall trade-offs that reflects the greater difficulty of plume masking.

Mask R-CNN (ResNet-50) achieves the highest F1 score (70.51\%), reflecting a 5.48 percentage point improvement over U-Net (65.03\%), driven primarily by a precision gain of 6.30 points (67.98\% vs. 61.68\%) at a moderate recall cost (80.19\% vs. 84.72\%). This difference is within the fold-to-fold standard deviation; the model choice rests on MethaneAIR, where the F1 gap (10.49 points) is several times the fold spread. The precision advantage over the semantic baseline is qualitatively confirmed in Figure~\ref{fig:msat_comparison_semantic_msat}: U-Net consistently produces masks that extend beyond the plume boundaries visible in the \mpc{reference plume mask labels}, particularly for the compact source in row~1 and the dispersed enhancement in row~3, producing precisions of P~0.68 and P~0.65 respectively, whereas Mask R-CNN (ResNet-50) masks tighter boundaries with markedly higher precision (P~0.92 and P~0.94 respectively) at the cost of lower recall. 

Mask R-CNN with ViT and MAE backbones again produce closely matched results (F1 66.98\% and 66.86\%), below ResNet-50 but above U-Net in precision. Both tend to fragment plumes into several partially overlapping instances, visible in rows~1 and~2 of Figure~\ref{fig:msat_comparison_semantic_msat}, which raises recall by covering more plume pixels while depressing precision. MAE again does not improve on random-initialized ViT despite pre-training on MethaneSAT L3 data.

\begin{table}[h]
\centering
\begin{tabular}{lccc}
\toprule
\textbf{Model} & \textbf{F1} & \textbf{Precision} & \textbf{Recall} \\
\midrule
U-Net            & 
65.03\scriptsize{$\pm$6.43} & 
61.68\scriptsize{$\pm$6.27} & 
84.72\scriptsize{$\pm$4.82} \\
Mask R-CNN (ResNet) & 
\textbf{70.51\scriptsize{$\pm$5.50}} & 
\textbf{67.98\scriptsize{$\pm$8.46}} & 
80.19\scriptsize{$\pm$4.15} \\

Mask R-CNN (ViT)    & 
66.98\scriptsize{$\pm$7.02} & 
62.37\scriptsize{$\pm$6.62} & 
\textbf{85.46\scriptsize{$\pm$1.17}} \\
Mask R-CNN (MAE)  & 
66.86\scriptsize{$\pm$7.20} & 
64.07\scriptsize{$\pm$10.88} & 
84.31\scriptsize{$\pm$7.43} \\
\bottomrule
\end{tabular}
\caption{Patch-level performance on the MethaneSAT test set at patch size 768$\times$768. Results are mean $\pm$ standard deviation over three cross-validation folds.}
\label{tab:msat_results}
\end{table}

\subsection{Cross-Sensor Transfer Learning}
\label{sec:results_transfer}

\begin{table*}[hb!]
\centering
\begin{tabular}{llllcc}
\toprule
\textbf{Method} & \textbf{Backbone} & \textbf{Strategy} & \textbf{F1} & \textbf{Precision} & \textbf{Recall} \\
\midrule
\multirow{4}{*}{U-Net}
 & \multirow{4}{*}{Encoder-Decoder}
 & MethaneSAT-only   & 65.03\scriptsize{$\pm$6.43} & 61.68\scriptsize{$\pm$6.27} & 84.72\scriptsize{$\pm$4.82} \\
 & & Fine-Tuning   &    68.04\scriptsize{$\pm$5.34} & 63.68\scriptsize{$\pm$4.87} & 88.87\scriptsize{$\pm$4.27} \\
 & & Joint &
 57.08\scriptsize{$\pm$5.12} & 
 54.49\scriptsize{$\pm$3.20} & 
 74.80\scriptsize{$\pm$17.63} \\
 & & Curriculum &
  66.66\scriptsize{$\pm$7.26} & 
 61.77\scriptsize{$\pm$6.45} & 
 88.89\scriptsize{$\pm$2.93} \\
\midrule
\multirow{12}{*}{Mask R-CNN}
 & \multirow{6}{*}{ResNet-50}
 & MethaneSAT-only   & 70.51\scriptsize{$\pm$5.50} & 67.98\scriptsize{$\pm$8.46} & 80.19\scriptsize{$\pm$4.15} \\
  & & Synthetic-only & 38.70\scriptsize{$\pm$4.70} & 56.10\scriptsize{$\pm$3.30} & 29.60\scriptsize{$\pm$4.60} \\
 & & Fine-Tuning       & \textbf{72.73\scriptsize{$\pm$7.47}} & \textbf{70.43\scriptsize{$\pm$8.69}} & 87.13\scriptsize{$\pm$7.79} \\
 & & Joint             & 70.57\scriptsize{$\pm$8.10} & 67.42\scriptsize{$\pm$11.36} & \textbf{89.06\scriptsize{$\pm$6.09}} \\
 & & Curriculum        & 70.69\scriptsize{$\pm$7.53} & 67.38\scriptsize{$\pm$10.56} & 87.69\scriptsize{$\pm$5.85} \\
\cmidrule{2-6}
 & \multirow{5}{*}{ViT}
 & MethaneSAT-only   & 66.98\scriptsize{$\pm$7.02} & 62.37\scriptsize{$\pm$6.62} & 85.46\scriptsize{$\pm$1.17} \\
 & & Fine-Tuning       & 66.77\scriptsize{$\pm$7.56} & 62.90\scriptsize{$\pm$8.09} & 84.84\scriptsize{$\pm$2.69} \\
 & & Joint             & 66.78\scriptsize{$\pm$7.44} & 62.73\scriptsize{$\pm$7.79} & 86.31\scriptsize{$\pm$3.80} \\
 & & Curriculum        & 66.11\scriptsize{$\pm$7.03} & 61.65\scriptsize{$\pm$6.60} & 87.02\scriptsize{$\pm$2.67} \\
\cmidrule{2-6}
 & MAE
 & MethaneSAT-only$^\dagger$ & 66.86\scriptsize{$\pm$7.20} & 
 64.07\scriptsize{$\pm$10.88} & 
 84.31\scriptsize{$\pm$7.43}\\

\bottomrule
\end{tabular}
\caption{Patch-level performance on the MethaneSAT test set at patch size 768$\times$768 across methods, backbones, and cross-sensor transfer strategies. Results are mean $\pm$ standard deviation over three cross-validation folds. All strategies initialize from MethaneAIR pre-trained weights, except MethaneSAT-only (trained from scratch) and MAE (self-supervised pre-training). 
\mpc{$^\dagger$MAE shares the ViT encoder and differs only in its initialization (self-supervised pre-training on unlabeled MethaneSAT data). The transfer strategies re-initialize the backbone from the MethaneAIR checkpoint, superseding this pre-training, so those runs coincide with the ViT rows and are not repeated; only the MethaneSAT-only row retains the MAE initialization.}
}
\label{tab:transfer_results}
\end{table*}

We evaluate whether cross-sensor transfer from MethaneAIR and synthetic data augmentation improve MethaneSAT detection performance under limited labeled data. Results for all transfer strategies and architectures are presented in Table~\ref{tab:transfer_results}, with the MethaneSAT-only (with no pre-training) serving as the reference point for each architecture.

\textbf{Mask R-CNN with ResNet-50} achieves the strongest overall patch-level performance. \textit{Fine-tuning} from MethaneAIR pre-trained weights improves F1 over the baseline, reaching 72.73\% with precision of 70.43\% and recall of 87.13\%. The simultaneous improvement in both precision and recall suggests that MethaneAIR pre-training provides genuinely better feature representations for the MethaneSAT domain, rather than simply shifting the precision-recall trade-off. At patch level this gain (2.2 points) is within the fold-to-fold spread; the evidence for the benefit of pre-training is at scene level (Section \ref{sec:results_scene}). \textit{Joint training} achieves the highest recall among all ResNet-50 strategies (89.06\%), with F1 of 70.57\% nearly identical to the baseline, although its margin over fine-tuning (87.13\%) is within the fold-to-fold standard deviation. \textit{Curriculum learning} closely tracks joint training (F1 70.69\%, recall 87.69\%), with no meaningful advantage over the simpler simultaneous exposure strategy. \mpc{Synthetic-only achieves a patch-level F1 of 38.70\% (precision 56.10\%, recall 29.60\%), far below every strategy trained on real MethaneSAT data. Together with its scene-level behavior (Section~\ref{sec:results_scene}), this shows what the model reaches without real MethaneSAT plume labels: it localizes 14 of the 22 test plumes at scene level, but its pixel-level masks agree poorly with the reference labels.}

\textbf{Mask R-CNN with ViT and MAE} backbones underperform ResNet-50 across all strategies, with F1 confined to 66.11--66.98\% for ViT and 66.86\% for MAE in the MethaneSAT-only setting. No transfer strategy improves the F1 of the ViT encoder over its MethaneSAT-only value. Its recall (84.84–87.02\%) exceeds that of ResNet-50 only in the MethaneSAT-only setting (85.46\% vs. 80.19\%) and is slightly lower under the three transfer strategies, while its precision is lower in every setting, by 4.7 to 7.5 percentage points. \mpc{Because the MethaneAIR-initialized transfer strategies replace the backbone weights, and MAE and ViT share the same encoder, MAE under those strategies reduces to the ViT rows of Table~\ref{tab:transfer_results}; only the MethaneSAT-only setting, where the self-supervised initialization is retained, is specific to MAE.}

\textbf{U-Net} with fine-tuning reaches F1 68.04\% and recall 88.87\%, a modest gain over its MethaneSAT-only baseline (F1 65.03\%) and still below ResNet-50 fine-tuning. Fine-tuning from MethaneAIR with a ResNet-50 backbone therefore remains the most effective strategy for patch-level adaptation, and the ViT-family backbones offer no advantage at this dataset scale, consistent with the data efficiency of convolutional inductive biases under limited training data. The large fold-to-fold standard deviations (up to $\pm$11 percentage points in precision) reflect the small number of MethaneSAT training scenes and motivate the scene-level evaluation of the next section.

\subsection{Scene-Level Evaluation and Post-Processing Analysis}
\label{sec:results_scene}

\subsubsection{Evaluation Setup} 

Patch-level pixel metrics assess segmentation quality within individual windows but do not reflect the operational performance of the full system: a model that correctly segments plume pixels within a patch may still produce fragmented or redundant detections at scene level, and conversely may generate false positives from retrieval artifacts that are spatially coherent across patch boundaries. We therefore shift to instance-level evaluation for scene-level analysis, where each predicted mask is treated as a discrete detection and evaluated against individual \mpc{reference plume mask labels}. Predictions are matched one-to-one to reference instances as defined in Section~\ref{sec:metrics}, with matching threshold $\theta = 0.1$. \mpc{This threshold is deliberately low because it acts as a source-localization criterion rather than a mask-fidelity criterion: reference masks are restricted to the plume segment within roughly 5--10~km of the source and are not extended farther downwind \cite{zhang_2026}, whereas predicted masks often follow the visible downwind tail, so a prediction that correctly identifies a source can have a small IoU with its reference mask. The dependence of all scene-level metrics on $\theta$ is reported in Appendix~\ref{app:theta_sensitivity}.} 


\mpc{Our test set is composed of five scenes from four basins; the scene inventory, the split assignment, and per-scene results are given in Appendix~\ref{app:msat_scenes}}. We compute scene-level results applying a confidence threshold $\tau = 0.8$ and NMS IoU threshold $\delta = 0.2$; these values were selected by cross-validation (see Appendix~\ref{app:thresholds} for details). Results are reported across three operating modes defined in Section~\ref{sec:inference}: the \textbf{baseline} mode applies only confidence filtering and NMS; the \textbf{high-sensitivity} mode additionally applies morphological filtering and proximity-based merging; the \textbf{high-precision} mode additionally applies the QND classifier, chosen for its physical basis before any test-set evaluation (a comparison with size thresholding is given for reference in Appendix~\ref{app:size_threshold_vs_QND}). Figure~\ref{fig:detection_panel} illustrates a representative scene-level detection result under the full post-processing pipeline. 

\begin{figure*}[hb!]
    \centering
    \includegraphics[width=\textwidth]{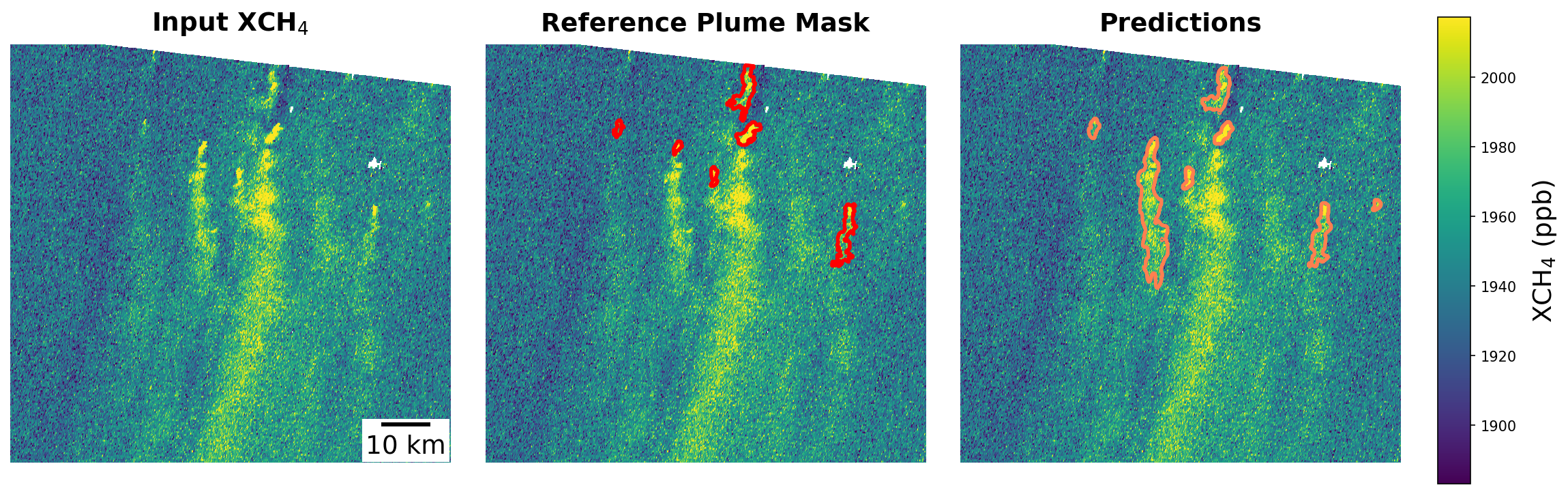}
    \caption{\textbf{Scene-level detection results for a representative MethaneSAT test scene.} Left: input XCH$_4$ map (ppb). Center: \mpc{reference plume mask labels} generated by the wavelet-based labeling method (red contours). Right: model predictions after the full post-processing pipeline (orange contours). Reference masks are conservative (Section~\ref{sec:labels}), so visible downwind tails extend beyond annotated boundaries. Predicted mask extents vary across plumes because the sliding window does not always observe the full plume in a single context window, and proximity-based merging across patch boundaries can produce spatially broader boundaries for some instances with large dispersed tails.}
    \label{fig:detection_panel}
\end{figure*}

\subsubsection{Baseline Results and Transfer Learning Comparison} 

Table~\ref{tab:scene_results} presents baseline instance-level results across all cross-sensor transfer strategies. The MethaneSAT-only baseline produces 18 true positives, 63 false positives, and 4 false negatives, with precision of 0.23 and F1 of 0.35. The extremely low precision confirms the finding from Section~\ref{sec:results_transfer} that training on MethaneSAT data alone is insufficient: without feature representations learned from MethaneAIR, the model cannot reliably distinguish real plumes from the retrieval artifacts and dispersed enhancements characteristic of MethaneSAT scenes.

The three strategies that add real MethaneSAT data to the MethaneAIR initialization improve substantially over this baseline. Fine-tuning achieves the strongest overall balance, recovering 22 true positives with at most one missed plume per fold, 15 false positives, and an F1 of 0.74. Joint training and curriculum learning both recover the same 22 true positives but accumulate 23 and 24 false positives respectively, compared to only 15 for fine-tuning. \mpc{To help interpret this difference, Table~\ref{tab:scene_results} includes a synthetic-only control: the same MethaneAIR initialization, trained on synthetic plumes only. On the real test scenes this control recovers 14 of the 22 reference plumes (recall 0.65), and it produces 24 false positives, the same level as joint training and curriculum learning and well above fine-tuning. The false-positive count of the strategies trained with synthetic data is thus close to that of the model trained on synthetic data alone, which is consistent with these detections being inherited from the synthetic training signal.} The three strategies trained with real MethaneSAT data miss almost no reference plume (FN rounds to 0), with or without synthetic plumes. Domain adaptation results are discussed separately in Appendix~\ref{app:domain_adaptation}, where they are evaluated as a supplementary comparison.

\begin{table*}[ht!]
\centering
\begin{tabular}{lcccccccc}
\toprule
\textbf{Architecture} & \textbf{Configuration} & \textbf{TP} & \textbf{FP} & \textbf{FN} & \textbf{mAP$_{\theta=0.1}$} & \textbf{F1} & \textbf{Precision} & \textbf{Recall} \\
\midrule
Mask R-CNN (ResNet) & MethaneSAT-only & 18 & 63 & 4 & 0.73\scriptsize{$\pm$0.14} & 0.35\scriptsize{$\pm$0.03} & 0.23\scriptsize{$\pm$0.05} & 0.80\scriptsize{$\pm$0.19} \\
& Synthetic-only & 14 & 24 & 8 & 0.55\scriptsize{$\pm$0.04} & 0.48\scriptsize{$\pm$0.08} & 0.39\scriptsize{$\pm$0.10} & 0.65\scriptsize{$\pm$0.05} \\
          & Fine-tuning      & \textbf{22} & \textbf{15} & \textbf{0} & 0.92\scriptsize{$\pm$0.04} & \textbf{0.74\scriptsize{$\pm$0.05}} & \textbf{0.60\scriptsize{$\pm$0.07}} & \textbf{0.98\scriptsize{$\pm$0.03}} \\
          & Joint       & \textbf{22} & 23 & \textbf{0} & \textbf{0.95\scriptsize{$\pm$0.03}} & 0.65\scriptsize{$\pm$0.06} & 0.49\scriptsize{$\pm$0.06} & \textbf{0.98\scriptsize{$\pm$0.03}} \\
          & Curriculum  & \textbf{22} & 24 & \textbf{0} & 0.93\scriptsize{$\pm$0.03} & 0.65\scriptsize{$\pm$0.08} & 0.49\scriptsize{$\pm$0.08} & \textbf{0.98\scriptsize{$\pm$0.03}} \\
\bottomrule
\end{tabular}
\caption{Baseline instance-level scene detection results with confidence threshold $\tau = 0.8$, NMS IoU $\delta = 0.2$, evaluation IoU $\theta = 0.1$, and patch size $768\times768$. TP, FP, and FN are counted at the plume instance level and averaged across three cross-validation folds. All MethaneAIR-initialized strategies use the same pre-trained weights; MethaneSAT-only is trained from scratch.} 
\label{tab:scene_results}
\end{table*}

\subsubsection{False positive analysis} 
\label{sec:fp_analysis}

To characterize the origin of residual false positives and assess whether any correspond to real methane enhancements excluded by conservative \mpc{reference plume mask labels} criteria, we conducted a manual review of 100 different false positive detections produced by the cross-sensor transfer configurations on the validation set. Each detection was visually inspected and assigned to one of four categories: confirmed false positive (FP), real plume missed by the wavelet labeling method (TP), uncertain enhancement (UNCERTAIN), or cloud artifact (CLOUD). Review criteria included whether the detection exhibited clear hotspots relative to the background, whether its shape is physically plausible and aligns with wind direction, and whether it could be attributed to nearby infrastructure. This taxonomy is intended as a qualitative characterization of systematic failure modes rather than a statistically powered frequency estimate: the validation set is too small for category counts to be interpreted as precise occurrence rates, and the findings should be read as illustrative patterns expected to recur across the full MethaneSAT archive.

\begin{figure*}[hb!]
\centering
\begin{tabular}{cccc}
\includegraphics[width=0.23\linewidth]{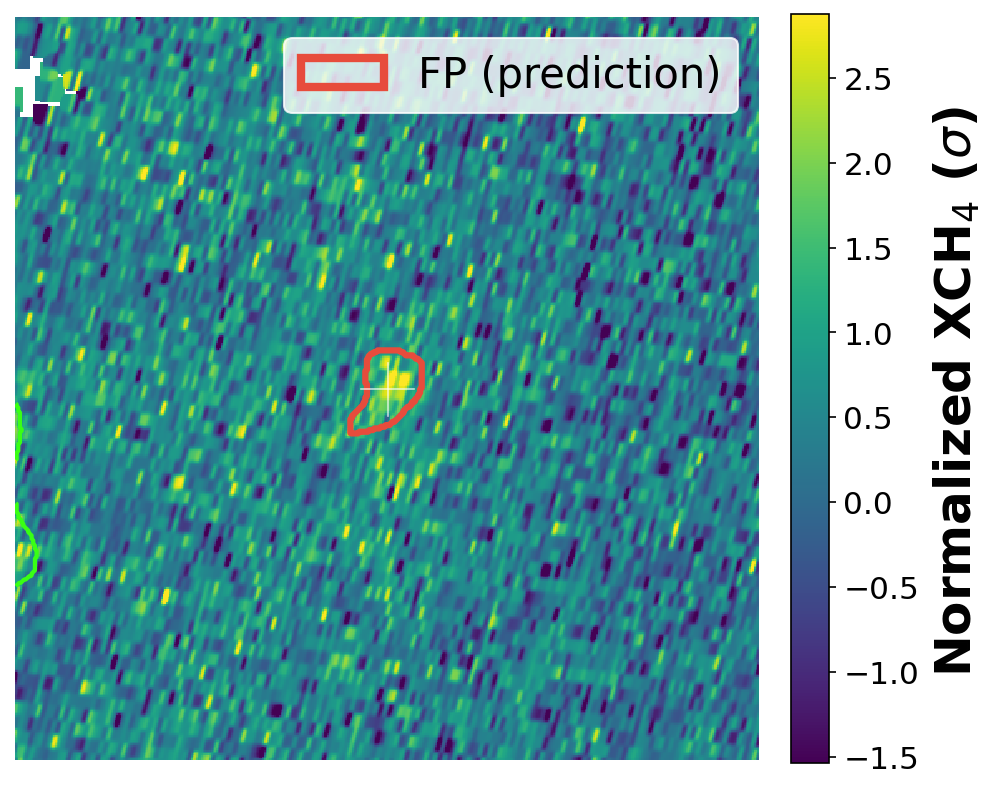} &
\includegraphics[width=0.23\linewidth]{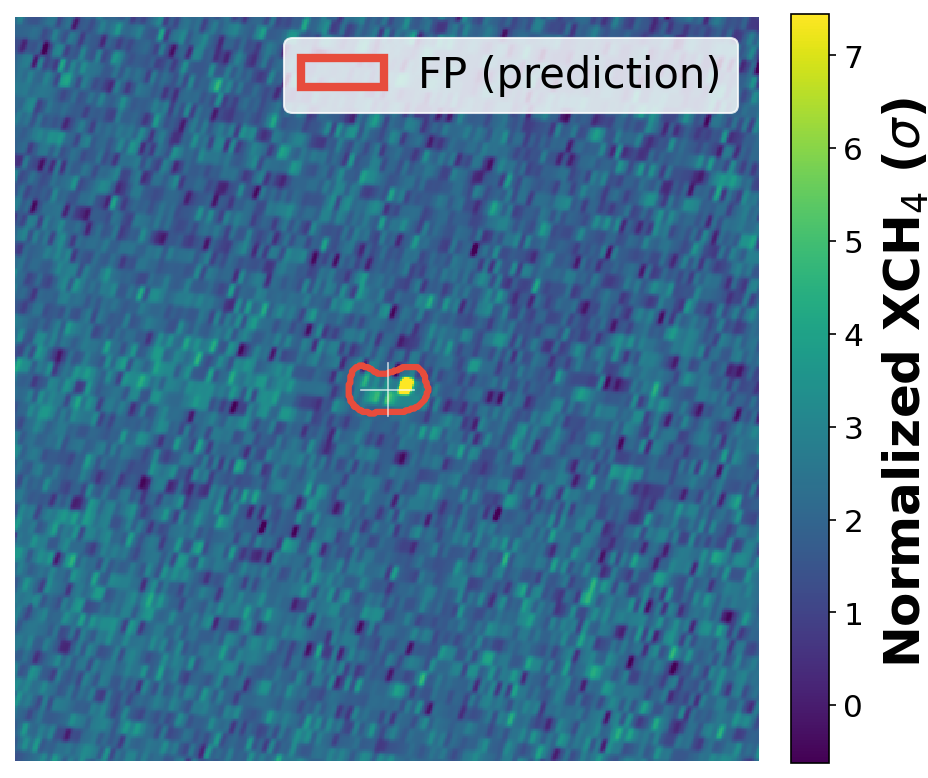} &
\includegraphics[width=0.23\linewidth]{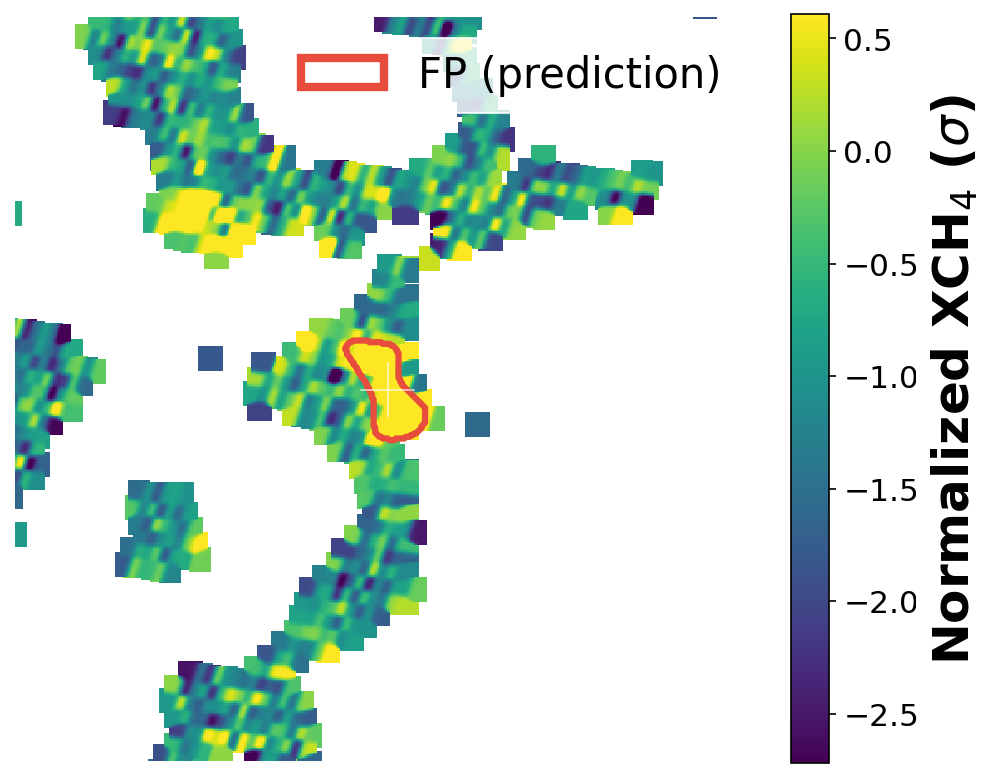} &
\includegraphics[width=0.23\linewidth]{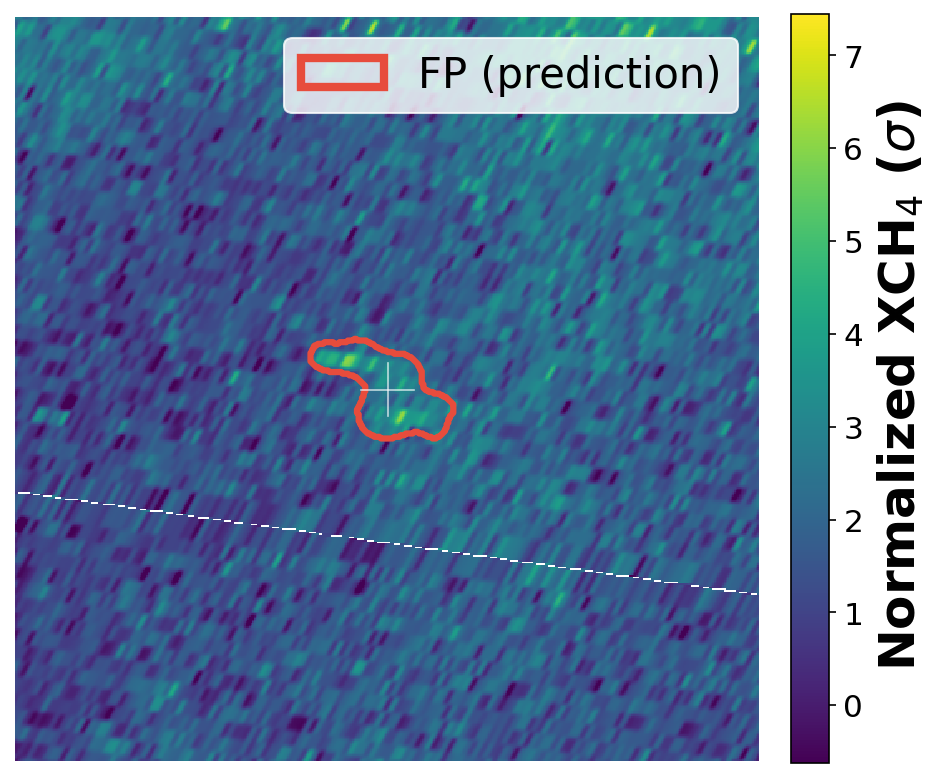} \\
\small (a) Real plume & \small (b) Small enhancement & \small (c) Cloud artifact & \small (d) Uncertain \\
\end{tabular}
\caption{Representative false positive categories identified during manual review. Normalized XCH$_4$ (per-scene z-score, $\sigma$ units). Red contours indicate the false positive predictions. (a) A detection reclassified as a real true positive: the model correctly identifies a real elongated plume that the wavelet method failed to retain. (b) A small compact enhancement representing the dominant false positive category ($\sim$60\% of detections), located near the detection limit in an otherwise clean scene. (c) A false positive associated with cloud-contaminated retrieval, where spatially coherent background anomalies mimic plume signals ($\sim$20\% of detections). (d) An uncertain detection near a retrieval quality boundary whose spatial pattern is consistent with a weak methane enhancement but cannot be confirmed from available evidence.}
\label{fig:fp_examples}
\end{figure*}

Small enhancements account for approximately 60\% of validation false positive detections, and cloud-contaminated retrievals account for a further 20\%, together explaining the large majority of spurious predictions. Representative examples of each category are shown in Figure~\ref{fig:fp_examples}. The small enhancement category (Figure~\ref{fig:fp_examples}b) consists of spatially compact concentration anomalies that lack the elongated morphology of a wind-dispersed plume but nonetheless exceed the model's detection threshold. The cloudy category (Figure~\ref{fig:fp_examples}c) arises from scattering artifacts in the XCH$_4$ retrieval that produce spatially coherent concentration patterns resembling plume signals, particularly at cloud edges where the CO$_2$ proxy retrieval is most sensitive to light path modifications.

Separately, on the test set, a manual inspection of the 15 false positives produced by fine-tuning reveals that not all are genuine errors. One detection was reclassified by an expert as a real methane enhancement that the wavelet method failed to retain, most likely because the enhancement lacked a sufficiently strong hotspot to pass the concentration filter (Figure~\ref{fig:fp_examples}a). One detection was classified as UNCERTAIN: the spatial pattern is consistent with a weak methane enhancement but insufficient evidence exists to confirm or exclude a real emission source (Figure~\ref{fig:fp_examples}d). These findings reflect a systematic property of the wavelet-based \mpc{reference plume mask labels}: conservative filtering criteria structurally exclude a fraction of real enhancements, and model predictions that disagree with these labels are not necessarily errors. This distinction directly motivates the dual-mode design discussed in Section~\ref{sec:results_postprocessing}. 


\subsubsection{Post-Processing Modes}
\label{sec:results_postprocessing}

Table~\ref{tab:scene_results_postprocessing} presents the full three-mode results across all transfer strategies. We focus the discussion on fine-tuning as the strongest baseline configuration, noting that patterns are broadly consistent across MethaneAIR-initialized strategies.

The high-sensitivity mode applies morphological filtering and proximity-based merging on top of the baseline. For fine-tuning, this raises F1 from 0.74 to 0.81, driven by a reduction in false positives from 15 to 8 with minimal recall cost (TP drops from 22 to 21, FN rises from 0 to 1), improving precision from 0.60 to 0.71. The pattern is consistent across joint training and curriculum learning, where false positives decrease by 8--11 while true positives decline by at most one. 

The high-precision mode applies an additional false positive suppression stage on top of the high-sensitivity output, the distribution-based QND classifier. Unlike a mask-area threshold, it uses the within-mask distributions of XCH$_4$ and surface albedo, so it can separate a real weak source from a surface artifact of similar extent; Appendix~\ref{app:size_threshold_vs_QND} compares the two (precision 0.92 for QND against 0.78 for size thresholding).

Applied to the fine-tuning configuration, the QND classifier achieves the highest precision across all strategies and modes (0.92), at the cost of dropping F1 slightly to 0.79 due to a 24 percentage point recall reduction (recall drops from 0.94 to 0.70). The six additional false negatives relative to high-sensitivity represent detections whose within-mask distributional signatures were classified as artifact-like, likely reflecting the challenge of separating real low-flux enhancements from surface-driven artifacts at MethaneSAT resolution. Joint training and curriculum learning show a similar pattern, with precision rising to 0.78 and 0.81 respectively, but at a false negative cost (FN = 6 for both).

\mpc{The per-scene breakdown (Table~\ref{tab:per_scene}, Appendix~\ref{app:msat_scenes}) shows that the aggregate numbers depend on individual scenes: the South Caspian scene c01A30410 alone contributes nine of the fifteen baseline false positives, and the misses of the high-precision mode fall in three of the five test scenes.}

\begin{table*}[ht!]
\centering
\begin{tabular}{lllccccccc}
\toprule
\textbf{Strategy} & \textbf{Mode} & \textbf{TP} & \textbf{FP} & \textbf{FN} & \textbf{mAP} & \textbf{F1} & \textbf{Precision} & \textbf{Recall} \\
\midrule
\multirow{3}{*}{MethaneSAT-only}
 & Baseline & 18 & 63 & 4 & 0.73\scriptsize{$\pm$0.14} & 0.35\scriptsize{$\pm$0.03} & 0.23\scriptsize{$\pm$0.05} & 0.80\scriptsize{$\pm$0.19} \\
 & High Sensitivity & 16 & 47 & 6 & 0.67\scriptsize{$\pm$0.12} & 0.38\scriptsize{$\pm$0.05} & 0.28\scriptsize{$\pm$0.08} & 0.71\scriptsize{$\pm$0.18} \\
 & High Precision (QND) & 12 & 9 & 10 & 0.58\scriptsize{$\pm$0.13} & 0.56\scriptsize{$\pm$0.09} & 0.61\scriptsize{$\pm$0.06} & 0.56\scriptsize{$\pm$0.20} \\
\cmidrule{1-9}
\multirow{3}{*}{Fine-Tuning}
 & Baseline & 22 & 15 & 0 & 0.92\scriptsize{$\pm$0.04} & 0.74\scriptsize{$\pm$0.05} & 0.60\scriptsize{$\pm$0.07} & \textbf{0.98\scriptsize{$\pm$0.03}} \\
 & High Sensitivity & 21 & 8 & 1 & 0.91\scriptsize{$\pm$0.05} & \textbf{0.81\scriptsize{$\pm$0.07}} & 0.71\scriptsize{$\pm$0.07} & 0.94\scriptsize{$\pm$0.05} \\
 & High Precision (QND) & 15 & 1 & 7 & 0.68\scriptsize{$\pm$0.04} & 0.79\scriptsize{$\pm$0.05} & \textbf{0.92\scriptsize{$\pm$0.04}} & 0.70\scriptsize{$\pm$0.05} \\
\cmidrule{1-9}
\multirow{3}{*}{Joint}
 & Baseline & 22 & 23 & 0 & \textbf{0.95\scriptsize{$\pm$0.03}} & 0.65\scriptsize{$\pm$0.06} & 0.49\scriptsize{$\pm$0.06} & \textbf{0.98\scriptsize{$\pm$0.03}} \\
 & High Sensitivity & 21 & 15 & 1 & 0.93\scriptsize{$\pm$0.01} & 0.72\scriptsize{$\pm$0.03} & 0.58\scriptsize{$\pm$0.04} & 0.94\scriptsize{$\pm$0.03} \\
 & High Precision (QND) & 16 & 5 & 6 & 0.71\scriptsize{$\pm$0.05} & 0.76\scriptsize{$\pm$0.03} & 0.78\scriptsize{$\pm$0.03} & 0.74\scriptsize{$\pm$0.03} \\
\cmidrule{1-9}
\multirow{3}{*}{Curriculum}
 & Baseline & 22 & 24 & 0 & 0.93\scriptsize{$\pm$0.03} & 0.65\scriptsize{$\pm$0.08} & 0.49\scriptsize{$\pm$0.08} & \textbf{0.98\scriptsize{$\pm$0.03}} \\
 & High Sensitivity & 21 & 13 & 1 & 0.92\scriptsize{$\pm$0.01} & 0.74\scriptsize{$\pm$0.10} & 0.62\scriptsize{$\pm$0.12} & 0.94\scriptsize{$\pm$0.05} \\
 & High Precision (QND) & 16 & 4 & 6 & 0.70\scriptsize{$\pm$0.07} & 0.76\scriptsize{$\pm$0.09} & 0.81\scriptsize{$\pm$0.11} & 0.73\scriptsize{$\pm$0.08} \\
\bottomrule
\end{tabular}
\caption{Instance-level scene detection results for \textbf{Mask R-CNN ResNet-50} across cross-sensor transfer strategies and output modes. The Baseline mode applies only confidence threshold ($\tau = 0.8$) and non-maximum suppression ($\delta = 0.2$). High Sensitivity and High Precision modes additionally apply physics-informed post-processing and spatial fragment merging, optimizing for recall and precision respectively. All results are at evaluation threshold IoU $\theta = 0.1$ and patch size 768$\times$768. All strategies initialize from MethaneAIR pre-trained weights, except MethaneSAT-only (trained from scratch). mAP is reported at $\theta = 0.1$. TP, FP, and FN are counted at the plume instance level and averaged across three cross-validation folds.}
\label{tab:scene_results_postprocessing}
\end{table*}

\subsection{Confidence-Weighted Scene Reconstruction}
\label{sec:results_prob}

\begin{figure*}[ht!]
    \centering
    \begin{subfigure}{\textwidth}
        \centering
        \includegraphics[width=\textwidth]{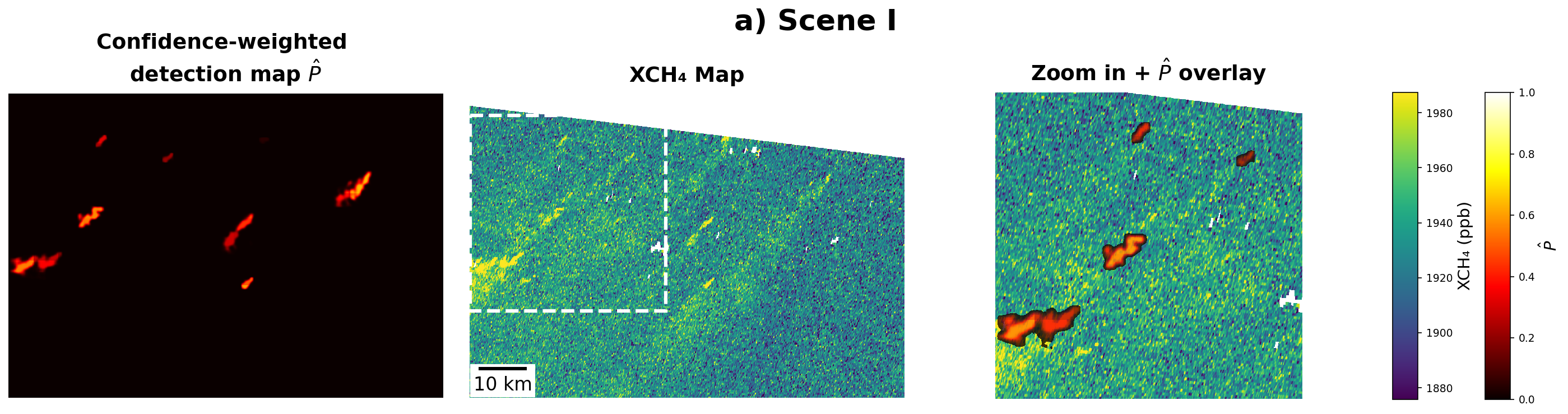}
        \label{fig:prob_scene_c01A30410}
    \end{subfigure}
    \vspace{0.5em}
    \begin{subfigure}{\textwidth}
        \centering
        \includegraphics[width=\textwidth]{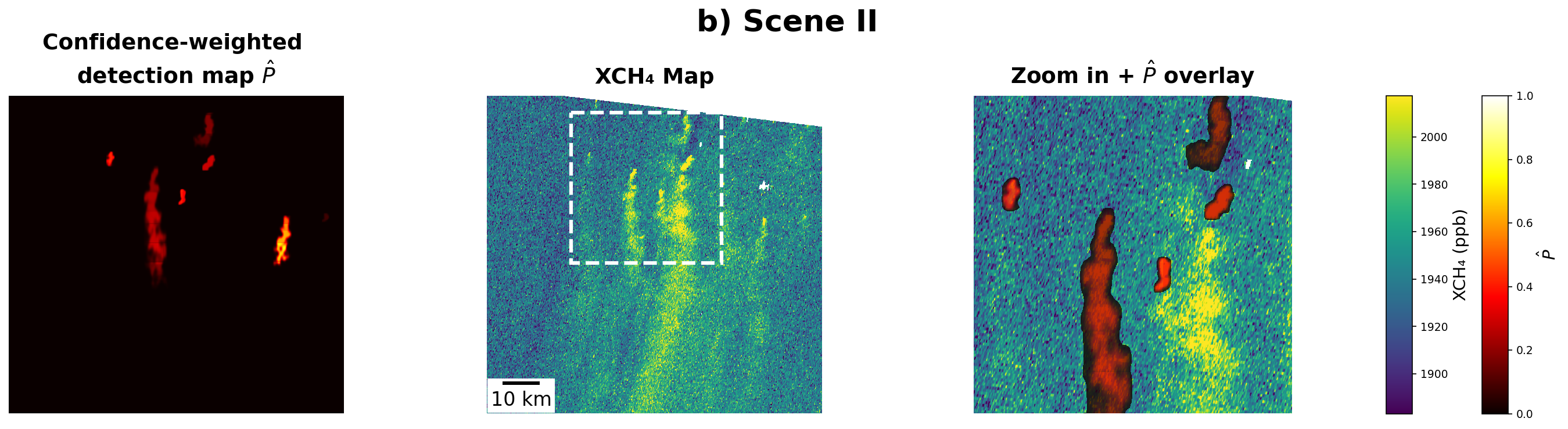}
        \label{fig:prob_scene_c03320380}
    \end{subfigure}
    \caption{\textbf{Confidence-weighted scene reconstruction for two MethaneSAT scenes.} Each row shows the confidence-weighted detection map $\hat{P}$ (left), the XCH$_4$ in ppb (center), and a zoom into the primary plume region with $\hat{P}$ overlaid on XCH$_4$ (right). The dashed white box in the center panel marks the zoomed region. High-$\hat{P}$ regions correspond to detected plume instances; background pixels are assigned near-zero $\hat{P}$ regardless of absolute XCH$_4$ concentration.}
    \label{fig:prob_maps}
\end{figure*}

The post-processing pipeline described in Section~\ref{sec:inference} operates on discrete instance detections, but the confidence-weighted aggregation scheme produces a continuous plume detection map $\hat{P}$ as an additional output that carries richer spatial information than the binary masks. Figure~\ref{fig:prob_maps} illustrates this output for two representative MethaneSAT scenes, showing the detection map, the XCH$_4$ field, and the $\hat{P}$ overlay side by side. In both scenes, high-$\hat{P}$ regions align tightly with the areas of elevated XCH$_4$. The first scene shows two intermediate-scale plumes in its center, and the compact high-intensity enhancement in its upper right portion are each assigned high $\hat{P}$ values that decay sharply at their physical boundaries and return to near-zero in the surrounding background. The second scene confirms this behavior across a different source configuration and emission intensity. This spatial sharpness is a direct consequence of the weighting scheme: pixels that the detections in the overlapping windows consistently assign to a plume receive high $\hat{P}$, and detections with confidence below $\tau$ are excluded.

Importantly, the detection map does not simply track absolute XCH$_4$ values. Background pixels with moderate methane values accumulate near $\hat{P} \approx 0$, confirming that detection confidence is driven by plume structure and multi-patch consistency rather than by concentration magnitude alone. A broader quantitative characterization of the relationship between $\hat{P}$ and XCH$_4$ maps, pooled across all test scenes, is provided in Appendix~\ref{app:prob_maps}.

\section{Discussion}
\label{sec:discussion}

The results support three interconnected claims about automated methane plume detection from MethaneSAT: that \mpc{the instance formulation we tested outperforms our U-Net baseline and supports per-source attribution}, that cross-sensor transfer from MethaneAIR is \mpc{the most effective of the adaptation strategies we evaluated} for overcoming data scarcity, and that scene-level reliability depends critically on post-processing design. We discuss each in turn, then address limitations and future directions.

\subsection{Mask R-CNN versus the U-Net Baseline}

The precision advantage of Mask R-CNN over U-Net has the same sign on both instruments and for all backbone variants. This is consistent with a structural difference between the two paradigms: U-Net classifies pixels independently and must make boundary decisions without an explicit mechanism for distinguishing plume instances from background anomalies, while Mask R-CNN first proposes and evaluates candidate regions before committing to a pixel mask. This two-stage commitment suppresses spurious activations in low-enhancement regions and produces tighter boundary masking.

The operational consequence extends beyond the F1 improvement. Semantic segmentation conflates spatially proximate plumes into a single connected region, making per-source attribution impossible without additional post-processing assumptions. Instance segmentation produces separate masks by design, enabling direct source association, independent per-instance filtering, and confidence scores that propagate naturally into emission quantification workflows.

Among backbone variants, ResNet-50 consistently outperforms ViT and MAE at the dataset scales available. The absence of any benefit from ViT fine-tuning or MAE self-supervised pre-training is consistent with the known data-efficiency advantages of convolutional inductive biases under limited labeled data: global self-attention requires more examples to realize its capacity for modeling long-range plume structure than the current MethaneSAT archive provides.

\mpc{We note that the comparison uses a single semantic baseline. U-Net is the de facto standard in the methane plume literature~\cite{Ronneberger2015, ruzicka_2023, Bruno_2024, Vaughan_2024, vaughan_2024b}, but other semantic architectures (e.g., DeepLabv3+~\cite{chen2018}, SegFormer~\cite{xie_2021}) were not evaluated, so the advantage reported here should be read as specific to the tested configurations rather than as a general superiority of instance over semantic segmentation.}

\subsection{Cross-Sensor Transfer Learning and the Role of Synthetic Data}

Fine-tuning from MethaneAIR pre-trained weights is the most effective adaptation strategy at scene level, with the same direction at patch level. At scene level it raises precision from 0.23 to 0.60 without loss of recall (0.98 vs.\ 0.80, Table~\ref{tab:scene_results}), which indicates that MethaneAIR pre-training gives better plume representations rather than a shift along the precision--recall trade-off. This is consistent with the shared CO$_2$ proxy retrieval used for both instruments, which produces qualitatively similar concentration morphologies despite the resolution gap.

Adding synthetic plumes through joint or curriculum training gave no measurable benefit over fine-tuning alone. At patch level, recall is 89.06\% and 87.69\% against 87.13\%, well within the fold-to-fold standard deviation, and precision is lower (67.4\% vs. 70.4\%). At scene level, recall is 0.98 for all three, and joint and curriculum training produce more false positives (23–24 vs. 15). The higher recall of joint training relative to the MethaneSAT-only model (89.06\% vs. 80.19\%) is shared by fine-tuning, so it comes mainly from the MethaneAIR initialization. \mpc{The synthetic-only control produces a false-positive count close to that of joint and curriculum training. This is compatible with the extra false positives coming from the synthetic supervision. It does not establish it: the control starts from MethaneAIR weights and these experiments cannot decide between a distribution gap between simulated and real plumes and conflicting updates from the two data sources during training. The F1 gap narrows after post-processing but does not close: in high-precision mode, joint and curriculum strategies reach F1 of 0.76 (precision 0.78--0.81, recall 0.73--0.74), compared to F1 of 0.79 for fine-tuning (precision 0.92, recall 0.70). The fine-tuning advantage is therefore driven by tighter false positive suppression at comparable recall. We do not conclude that synthetic data is harmful in general, only that its benefit was not measurable with this dataset and design.}

\subsection{Post-Processing Design and the Dual-Mode Framework}

The false positive analysis reveals that the model's spurious detections are not random: approximately 60\% are small compact enhancements lacking wind-dispersed plume morphology, and a further 20\% are cloud-contaminated retrievals. Crucially, a meaningful fraction of apparent false positives correspond to real methane enhancements that the wavelet-based \mpc{reference plume mask labels} conservatively exclude. Precision metrics computed against wavelet labels may therefore systematically underestimate true detection performance, a caveat that applies to all results reported in this paper. The same caveat shapes how the two operational modes should be compared: the question is not which mode is more accurate in absolute terms, but which precision-recall trade-off matches a given monitoring application.

This finding directly motivates the dual-mode design. Moving from the high-sensitivity to the high-precision mode trades recall for precision (Table~\ref{tab:scene_results_postprocessing}), and whether this trade-off is favorable depends on the downstream use case. For comprehensive emission screening, where the goal is to recover as many real plumes as possible and downstream manual review can filter false positives, the recall cost of the high-precision mode is not justified, and the high-sensitivity mode is the appropriate operating point: it preserves 21 of 22 \mpc{reference plume mask labels} with only 8 false positives, a workload that is tractable for human inspection. For regulatory reporting and automated source attribution, where every flagged detection must be defensible and false positives carry a larger operational cost than missed sources, the high-precision mode is appropriate. The dual-mode framework is intended to make this choice explicit at deployment rather than baking a single operating point into the system.



\subsection{Post-Processing vs End-to-End Learning}
\label{sec:postproc_vs_e2e}

A natural question is whether the post-processing pipeline is necessary at all: in principle, Mask R-CNN should learn to reject these failure modes (small enhancements, striping artifacts, cloudy scenes) directly from spatial context, given sufficient diverse training data. We explored this by progressively expanding the hard-negative pool to cover additional edge cases. However, we observed that as the proportion of hard negatives grew, false negative rates on real plumes increased, which we attribute to the resulting class imbalance shifting the model's prior toward the negative class. We therefore maintained an approximately balanced ratio of plume to hard-negative samples, and offloaded the residual artifact discrimination to the post-processing pipeline. 
\mpc{This is the foreground-background imbalance that affects object detectors in general, reviewed by Oksuz et al. \cite{oksuz_2021}. The standard solutions control how much of the training signal background is allowed to take: Fast R-CNN caps the sampled positive-to-negative RoI ratio at 1:3 \cite{girshick_2015}, online hard example mining replaces that fixed ratio by selecting the highest-loss regions in each mini-batch \cite{shrivastava_2016}, and focal loss down-weights easy examples in the loss \cite{lin_2020}. All three operate on the regions of the images already in the batch, and our detector applies the first of them. None of them changes the composition of the archive: a patch containing no plume yields only negative regions, and the number of plume patches relative to hard-negative patches is what sets the overall balance. The last two mechanisms work against us here, because the negatives we add are hard by construction. They are the patches the model currently gets wrong, so loss-based selection and focal weighting both amplify them rather than suppress them. Adding hard negatives therefore makes the training set more negative rather than more informative. Class-balanced losses \cite{cui_2019} and resampling schemes \cite{gupta_2019} redistribute an archive that already exists. In our pipeline the detector is trained on an approximately balanced patch set, and the remaining artifact discrimination is handled by post-processing and the QND classifier. The STARCOP hyperspectral plume dataset was built on the same principle, pairing its plume tiles with an equal number of negatives and drawing half of those from locations where the matched filter responds strongly but no plume is present \cite{ruzicka_2023}. A substantially larger labeled archive would allow richer hard-negative coverage while keeping the class ratio fixed, which we expect would reduce the importance of explicit post-processing. We leave this exploration to future work.}

\subsection{Limitations and Future Directions}
\label{sec:limitations}

The MethaneSAT dataset comprises 27 scenes from a limited range of geographic regions and atmospheric conditions, and the held-out test set only five of them, with 22 reference plumes across four basins (Appendix~\ref{app:msat_scenes}). \mpc{The per-scene breakdown (Table~\ref{tab:per_scene}) shows that one scene contributes most of the false-positive count and that the misses of the high-precision mode fall in three of the five scenes, so the aggregate scene-level metrics should be read as indicative rather than as estimates of performance across basins, surface types, and cloud conditions.} The high cross-fold variance in all reported metrics reflects the small training set, and performance on unrepresented basins or meteorological regimes may differ from reported values. \mpc{MethaneSAT was lost in June 2025, and the labeled set is limited by the manual quality control of Section~\ref{sec:labels}; only the Permian basin (7 scenes, 31 reference plumes) could be held out as a whole. Labeling more of the archive is left to future work.}


The boundary between real low-flux plumes and dispersed enhancements is inherently ambiguous at MethaneSAT resolution. Small compact enhancements near the instrument detection limit may represent real weak sources, retrieval artifacts, or surface heterogeneity, and neither the model nor the wavelet-based \mpc{reference plume mask labels} can resolve this ambiguity from a single overpass. Longitudinal aggregation of detections across repeated overpasses offers a practical path forward: sources consistently detected across multiple passes are unlikely to be retrieval artifacts, providing a data-driven mechanism for resolving ambiguous single-scene predictions without requiring additional labeled data.

Finally, the moderate spatial correlation between the confidence-weighted detection map $\hat{P}$ and XCH$_4$ within detected plume regions suggests a direct downstream application: using $\hat{P}$ as a spatial prior for emission rate estimation, replacing hard binary mask boundaries with $\hat{P}$-weighted concentration fields    to produce smoother, boundary-condition-independent estimates of enclosed methane column. This represents a natural next step toward a fully automated quantification pipeline operating on the MethaneSAT archive.

\section{Conclusions}

This paper presented an operational workflow for automated methane plume detection from MethaneSAT XCH$_4$ retrievals, addressing two core challenges: data scarcity under limited labeled satellite observations, and inference reliability across diverse atmospheric and surface conditions at basin scale.

\mpc{We showed that Mask R-CNN outperforms the tested U-Net baseline in pixel-level F1 on MethaneAIR; on MethaneSAT the difference has the same sign but is within the fold-to-fold spread. The gain comes from precision rather than recall, which is consistent with the two-stage detection commitment of the instance formulation, and instance segmentation produces a separate mask per candidate plume, so per-source attribution and per-detection filtering are possible without additional assumptions. ResNet-50 outperformed the ViT and MAE backbones, and self-supervised pre-training on unlabeled MethaneSAT data gave no improvement over random initialization, which suggests that labeled data volume rather than feature initialization is the limiting factor at this scale.}

\mpc{Cross-sensor fine-tuning from MethaneAIR was the most effective adaptation strategy under limited labeled data: at scene level it cut false positives by a factor of four and raised F1 from 0.35 to 0.74, while the patch-level gain was within the fold-to-fold spread. This is consistent with the CO$_2$ proxy retrieval shared by the two instruments: despite the difference in spatial resolution, plume representations learned from MethaneAIR transfer to the MethaneSAT scenes we evaluated. Synthetic augmentation gave no measurable gain over fine-tuning: recall was equal within fold variance and scene-level false positives were higher. A synthetic-only control produced a similar false-positive count, but our design cannot attribute it to a specific cause. Finally, physics-informed post-processing converts patch-level detections into two operating modes: a high-sensitivity mode that removes physically implausible predictions while retaining marginal detections for longitudinal screening, and a high-precision mode in which the QND classifier, applied zero-shot from MethaneAIR, reaches the highest precision of any configuration. Manual review indicates that a fraction of the remaining false positives are real enhancements excluded by the conservative labeling criteria, so the precision values reported here are likely underestimates.}

The 14 months of basin-scale XCH$_4$ data that MethaneSAT collected before its loss in June 2025 are a unique record, and the framework presented here provides a plume detection framework for extracting emission information from it. More broadly, the cross-sensor transfer strategy, dual-mode post-processing design, and confidence-weighted output representation are not instrument-specific: they can be transferred naturally to sensors sharing similar retrieval physics and future dedicated methane monitoring missions, providing a reusable foundation for scalable plume detection in support of global methane emission monitoring.

\section{Acknowledgments}
Funding for MethaneSAT and MethaneAIR activities was provided in part by Anonymous, Arnold Ventures, The Audacious Project, Ballmer Group, Bezos Earth Fund, The Children’s Investment Fund Foundation, Heising-Simons Family Fund, King Philanthropies, Robertson Foundation, Skyline Foundation and Valhalla Foundation. For a more complete list of funders, please visit \url{www.methanesat.org}. We express our gratitude to the MethaneAIR field team for their contributions to MethaneAIR data collection. We thank the AstroAI and EarthAI institutes at the Center for Astrophysics $|$ Harvard \& Smithsonian for useful discussions and guidance. CG and MPC were supported by AstroAI at the Center for Astrophysics $|$ Harvard and Smithsonian and Smithsonian Institution's combined science and research funding. TUM authors acknowledge the financial support by the EU projects “PAUL” (Grant 101037319) and “CoSense4Climate” (Grant 101089203)

\bibliographystyle{unsrt}
\bibliography{references}

@article{ccm2026Omethanesat,
  author  = {Chan Miller, Christopher and Roche, S{\'e}bastien and Wilzweski, Jonas and Liu, Xiong and Chance, Kelly V. and Luo, Bingkun and Sun, Kang and Franklin, Jonathan E. and Benmergui, Joshua S. and Nasr, Maya and Miller, David J. and Crowell, Sean and Leisso, Nathan and Walker, Eleanor and Gautam, Ritesh and LoFaso, Nicholas and Ayvazov, Sasha and Wells, David and Cervantes, Carlos and Hairfield, Christopher and Wofsy, Steven},
  title   = {Operational XCH4 Retrievals from MethaneSAT: Demonstrating Sensor Performance for Constraining Regional Methane Emissions},
  journal = {EGUsphere},
  year    = {2026},
  doi     = {10.5194/egusphere-2026-3365},
  url     = {https://doi.org/10.5194/egusphere-2026-3365},
  note    = {Preprint}
}

@Article{chulakadabba2023methane,
AUTHOR = {Chulakadabba, A. and Sargent, M. and Lauvaux, T. and Benmergui, J. S. and Franklin, J. E. and Chan Miller, C. and Wilzewski, J. S. and Roche, S. and Conway, E. and Souri, A. H. and Sun, K. and Luo, B. and Hawthrone, J. and Samra, J. and Daube, B. C. and Liu, X. and Chance, K. and Li, Y. and Gautam, R. and Omara, M. and Rutherford, J. S. and Sherwin, E. D. and Brandt, A. and Wofsy, S. C.},
TITLE = {Methane point source quantification using MethaneAIR: a new airborne imaging spectrometer},
JOURNAL = {Atmospheric Measurement Techniques},
VOLUME = {16},
YEAR = {2023},
NUMBER = {23},
PAGES = {5771--5785},
URL = {https://amt.copernicus.org/articles/16/5771/2023/},
DOI = {10.5194/amt-16-5771-2023}
}

@Article{Conway_2024,
AUTHOR = {Conway, E. K. and Souri, A. H. and Benmergui, J. and Sun, K. and Liu, X. and Staebell, C. and Chan Miller, C. and Franklin, J. and Samra, J. and Wilzewski, J. and Roche, S. and Luo, B. and Chulakadabba, A. and Sargent, M. and Hohl, J. and Daube, B. and Gordon, I. and Chance, K. and Wofsy, S.},
TITLE = {Level0 to Level1B processor for MethaneAIR},
JOURNAL = {Atmospheric Measurement Techniques},
VOLUME = {17},
YEAR = {2024},
NUMBER = {4},
PAGES = {1347--1362},
URL = {https://amt.copernicus.org/articles/17/1347/2024/},
DOI = {10.5194/amt-17-1347-2024}
}

@InProceedings{Ronneberger2015,
author="Ronneberger, Olaf
and Fischer, Philipp
and Brox, Thomas",
editor="Navab, Nassir
and Hornegger, Joachim
and Wells, William M.
and Frangi, Alejandro F.",
title="U-Net: Convolutional Networks for Biomedical Image Segmentation",
booktitle="Medical Image Computing and Computer-Assisted Intervention -- MICCAI 2015",
year="2015",
publisher="Springer International Publishing",
address="Cham",
pages="234--241",
isbn="978-3-319-24574-4"
}

@article{Etminan_2016,
  title={Radiative forcing of carbon dioxide, methane, and nitrous oxide: A significant revision of the methane radiative forcing},
  author={Maryam Etminan and Gunnar Myhre and Eleanor J. Highwood and Keith P. Shine},
  journal={Geophysical Research Letters},
  year={2016},
  volume={43},
  pages={12,614 - 12,623},
  url={https://api.semanticscholar.org/CorpusID:6813997}
}

@inbook{myhre_2013,
  author={Myhre, G. and Shindell, D. and Bréon, F.-M. and Collins, W. and Fuglestvedt, J. and Huang, J. and Koch, D. and Lamarque, J.-F. and Lee, D. and Mendoza, B. and Nakajima, T. and Robock, A. and Stephens, G. and Takemura, T. and Zhang, H.},
  editor={Stocker, T. F. and Qin, D. and Plattner, G.-K. and Tignor, M. and Allen, S. K. and Doschung, J. and Nauels, A. and Xia, Y. and Bex, V. and Midgley, P. M.},
  title={Anthropogenic and natural radiative forcing},
  booktitle={Climate Change 2013: The Physical Science Basis. Contribution of Working Group I to the Fifth Assessment Report of the Intergovernmental Panel on Climate Change},
  year={2013},
  pages={659--740},
  publisher={Cambridge University Press},
  address={Cambridge, UK},
  doi={10.1017/CBO9781107415324.018},
}

@Article{Rohrschneider,
AUTHOR = {Rohrschneider, R. R and
Wofsy, S. and Franklin, J. E. and
Benmergui, J. and Soto, J. C and
Davis, Spencer B},
TITLE = {The MethaneSAT Mission},
JOURNAL = {Small Satellite Conference},
DOI = {}
}

@article{frankenberg_2016,
author = {Christian Frankenberg  and Andrew K. Thorpe  and David R. Thompson  and Glynn Hulley  and Eric Adam Kort  and Nick Vance  and Jakob Borchardt  and Thomas Krings  and Konstantin Gerilowski  and Colm Sweeney  and Stephen Conley  and Brian D. Bue  and Andrew D. Aubrey  and Simon Hook  and Robert O. Green },
title = {Airborne methane remote measurements reveal heavy-tail flux distribution in Four Corners region},
journal = {Proceedings of the National Academy of Sciences},
volume = {113},
number = {35},
pages = {9734-9739},
year = {2016},
doi = {10.1073/pnas.1605617113},
URL = {https://www.pnas.org/doi/abs/10.1073/pnas.1605617113},
eprint = {https://www.pnas.org/doi/pdf/10.1073/pnas.1605617113}}

@Article{Varon_2021,
AUTHOR = {Varon, D. J. and Jervis, D. and McKeever, J. and Spence, I. and Gains, D. and Jacob, D. J.},
TITLE = {High-frequency monitoring of anomalous methane point sources with
multispectral Sentinel-2 satellite observations},
JOURNAL = {Atmospheric Measurement Techniques},
VOLUME = {14},
YEAR = {2021},
NUMBER = {4},
PAGES = {2771--2785},
URL = {https://amt.copernicus.org/articles/14/2771/2021/},
DOI = {10.5194/amt-14-2771-2021}
}

@article{Veefkind_2012,
title = {TROPOMI on the ESA Sentinel-5 Precursor: A GMES mission for global observations of the atmospheric composition for climate, air quality and ozone layer applications},
journal = {Remote Sensing of Environment},
volume = {120},
pages = {70-83},
year = {2012},
note = {The Sentinel Missions - New Opportunities for Science},
issn = {0034-4257},
doi = {https://doi.org/10.1016/j.rse.2011.09.027},
url = {https://www.sciencedirect.com/science/article/pii/S0034425712000661},
author = {J.P. Veefkind and I. Aben and K. McMullan and H. Förster and J. {de Vries} and G. Otter and J. Claas and H.J. Eskes and J.F. {de Haan} and Q. Kleipool and M. {van Weele} and O. Hasekamp and R. Hoogeveen and J. Landgraf and R. Snel and P. Tol and P. Ingmann and R. Voors and B. Kruizinga and R. Vink and H. Visser and P.F. Levelt}
}

@Article{jervis_2021,
AUTHOR = {Jervis, D. and McKeever, J. and Durak, B. O. A. and Sloan, J. J. and Gains, D. and Varon, D. J. and Ramier, A. and Strupler, M. and Tarrant, E.},
TITLE = {The GHGSat-D imaging spectrometer},
JOURNAL = {Atmospheric Measurement Techniques},
VOLUME = {14},
YEAR = {2021},
NUMBER = {3},
PAGES = {2127--2140},
URL = {https://amt.copernicus.org/articles/14/2127/2021/},
DOI = {10.5194/amt-14-2127-2021}
}

@ARTICLE{Guanter_2021,
       author = {{Guanter}, Luis and {Irakulis-Loitxate}, Itziar and {Gorro{\~n}o}, Javier and {S{\'a}nchez-Garc{\'\i}a}, Elena and {Cusworth}, Daniel H. and {Varon}, Daniel J. and {Cogliati}, Sergio and {Colombo}, Roberto},
        title = "{Mapping methane point emissions with the PRISMA spaceborne imaging spectrometer}",
      journal = {Remote Sensing of Environment},
         year = 2021,
        month = nov,
       volume = {265},
        pages = {112671},
          doi = {10.1016/j.rse.2021.11267110.31223/x5vc9c},
       adsurl = {https://ui.adsabs.harvard.edu/abs/2021RSEnv.26512671G}
}

@article{ruzicka_2023,
  edition = {},
  number = {1},
  journal = {Scientific Reports},
  pages = {},
  publisher = {Springer Nature},
  school = {},
  title = {Semantic segmentation of methane plumes with hyperspectral machine learning models},
  volume = {13},
  author = {Růžička, V and Mateo-Garcia, G and Gómez-Chova, L and Vaughan, A and Guanter, L and Markham, A},
  editor = {},
  year = {2023},
  series = {}
}

@inproceedings{he2017mask,
  title={Mask r-cnn},
  author={He, Kaiming and Gkioxari, Georgia and Doll{\'a}r, Piotr and Girshick, Ross},
  booktitle={Proceedings of the IEEE international conference on computer vision},
  pages={2961--2969},
  year={2017}
}

@Article{Chan_2024,
AUTHOR = {Chan Miller, C. and Roche, S. and Wilzewski, J. S. and Liu, X. and Chance, K. and Souri, A. H. and Conway, E. and Luo, B. and Samra, J. and Hawthorne, J. and Sun, K. and Staebell, C. and Chulakadabba, A. and Sargent, M. and Benmergui, J. S. and Franklin, J. E. and Daube, B. C. and Li, Y. and Laughner, J. L. and Baier, B. C. and Gautam, R. and Omara, M. and Wofsy, S. C.},
TITLE = {Methane retrieval from MethaneAIR using the CO$_2$ proxy approach: a demonstration for the upcoming MethaneSAT mission},
JOURNAL = {Atmospheric Measurement Techniques},
VOLUME = {17},
YEAR = {2024},
NUMBER = {18},
PAGES = {5429--5454},
URL = {https://amt.copernicus.org/articles/17/5429/2024/},
DOI = {10.5194/amt-17-5429-2024}
}

@article{Shelhamer_2014,
  title={Fully convolutional networks for semantic segmentation},
  author={Evan Shelhamer and Jonathan Long and Trevor Darrell},
  journal={2015 IEEE Conference on Computer Vision and Pattern Recognition (CVPR)},
  year={2014},
  pages={3431-3440},
  url={https://api.semanticscholar.org/CorpusID:1629541}
}

@article{He_2015,
  title={Deep Residual Learning for Image Recognition},
  author={Kaiming He and X. Zhang and Shaoqing Ren and Jian Sun},
  journal={2016 IEEE Conference on Computer Vision and Pattern Recognition (CVPR)},
  year={2015},
  pages={770-778},
  url={https://api.semanticscholar.org/CorpusID:206594692}
}

@article{Dosovitskiy_2020,
  title={An Image is Worth 16x16 Words: Transformers for Image Recognition at Scale},
  author={Alexey Dosovitskiy and Lucas Beyer and Alexander Kolesnikov and Dirk Weissenborn and Xiaohua Zhai and Thomas Unterthiner and Mostafa Dehghani and Matthias Minderer and Georg Heigold and Sylvain Gelly and Jakob Uszkoreit and Neil Houlsby},
  journal={ArXiv},
  year={2020},
  volume={abs/2010.11929},
  url={https://api.semanticscholar.org/CorpusID:225039882}
}

@article{He_2021,
  title={Masked Autoencoders Are Scalable Vision Learners},
  author={Kaiming He and Xinlei Chen and Saining Xie and Yanghao Li and Piotr Doll'ar and Ross B. Girshick},
  journal={2022 IEEE/CVF Conference on Computer Vision and Pattern Recognition (CVPR)},
  year={2021},
  pages={15979-15988},
  url={https://api.semanticscholar.org/CorpusID:243985980}
}

@article{weforum2024methane,
  title = {Global Methane Pledge: which countries are cutting emissions?},
  author = {{World Economic Forum}},
  journal = {World Economic Forum},
  year = {2024},
  month = {August},
  url = {https://www.weforum.org/stories/2024/08/global-methane-pledge-which-countries-are-cutting-emissions/},
  note = {Accessed: 2025-07-17}
}

@misc{ogdc_2023,
  title = {Oil \& Gas Decarbonization Charter},
  author = {{COP28 UAE}},
  year = {2023},
  howpublished = {\url{https://www.ogdc.org/}},
  note = {Launched at COP28, Dubai},
  organization = {Oil \& Gas Decarbonization Charter}
}

@inproceedings{Xu_2025,
  title={MEQNet: Deep Learning for Methane Point Source Emission Quantification from Sentinel-2 Observations},
  author={Xu, Di and Mason, Philippa and Liu, Jianguo and Wang, Yanghua},
  booktitle={NeurIPS 2025 Workshop on Tackling Climate Change with Machine Learning},
  url={https://www.climatechange.ai/papers/neurips2025/17},
  year={2025}
}

@article{Bue_2025,
  title={Towards Operational Automated Greenhouse Gas Plume Detection},
  author={Brian D. Bue and Jake H. Lee and Andrew K. Thorpe and Philip G. Brodrick and Daniel H. Cusworth and Alana K. Ayasse and Vassiliki Mancoridis and Anagha Satish and Shujun Xiong and Riley M. Duren},
  journal={ArXiv},
  year={2025},
  volume={abs/2505.21806},
  url={https://api.semanticscholar.org/CorpusID:278959261}
}

@ARTICLE{Mancoridis_2025,
  author={Mancoridis, Vassiliki and Bue, Brian and Lee, Jake H. and Thorpe, Andrew K. and Cusworth, Daniel and Ayasse, Alana and Brodrick, Philip G. and Duren, Riley},
  journal={IEEE Transactions on Geoscience and Remote Sensing}, 
  title={Multiplatform Methane Plume Detection via Model and Domain Adaptation}, 
  year={2025},
  volume={63},
  number={},
  pages={1-12},
  doi={10.1109/TGRS.2025.3608601}}

@article{agarap_2018,
  title={Deep Learning using Rectified Linear Units (ReLU)},
  author={Abien Fred Agarap},
  journal={ArXiv},
  year={2018},
  volume={abs/1803.08375},
  url={https://api.semanticscholar.org/CorpusID:4090379}
}

@article{Watine-Guiu_2023,
author = {Marc Watine-Guiu  and Daniel J. Varon  and Itziar Irakulis-Loitxate  and Nicholas Balasus  and Daniel J. Jacob },
title = {Geostationary satellite observations of extreme and transient methane emissions from oil and gas infrastructure},
journal = {Proceedings of the National Academy of Sciences},
volume = {120},
number = {52},
pages = {e2310797120},
year = {2023},
doi = {10.1073/pnas.2310797120},
URL = {https://www.pnas.org/doi/abs/10.1073/pnas.2310797120},
eprint = {https://www.pnas.org/doi/pdf/10.1073/pnas.2310797120}}

@article{zavala_2015,
  title={Reconciling divergent estimates of oil and gas methane emissions},
  author={Zavala-Araiza, Daniel and Lyon, David R and Alvarez, Ram{\'o}n A and Davis, Kenneth J and Harriss, Robert and Herndon, Scott C and Karion, Anna and Kort, Eric Adam and Lamb, Brian K and Lan, Xin and others},
  journal={Proceedings of the National Academy of Sciences},
  volume={112},
  number={51},
  pages={15597--15602},
  year={2015},
  publisher={National Academy of Sciences}
}

@article{brandt_2014,
  title={Methane leaks from North American natural gas systems},
  author={Brandt, Adam R and Heath, GA and Kort, EA and O'Sullivan, Francis and P{\'e}tron, Gabrielle and Jordaan, Sarah M and Tans, P and Wilcox, Jennifer and Gopstein, AM and Arent, Doug and others},
  journal={Science},
  volume={343},
  number={6172},
  pages={733--735},
  year={2014},
  publisher={American Association for the Advancement of Science}
}

@Article{jacob_2016,
AUTHOR = {Jacob, D. J. and Turner, A. J. and Maasakkers, J. D. and Sheng, J. and Sun, K. and Liu, X. and Chance, K. and Aben, I. and McKeever, J. and Frankenberg, C.},
TITLE = {Satellite observations of atmospheric methane and their value for
quantifying methane emissions},
JOURNAL = {Atmospheric Chemistry and Physics},
VOLUME = {16},
YEAR = {2016},
NUMBER = {22},
PAGES = {14371--14396},
URL = {https://acp.copernicus.org/articles/16/14371/2016/},
DOI = {10.5194/acp-16-14371-2016}
}

@article{cusworth_2021,
  title={Multisatellite imaging of a gas well blowout enables quantification of total methane emissions},
  author={Cusworth, Daniel H and Duren, Riley M and Thorpe, Andrew K and Pandey, Sudhanshu and Maasakkers, Joannes D and Aben, Ilse and Jervis, Dylan and Varon, Daniel J and Jacob, Daniel J and Randles, Cynthia A and others},
  journal={Geophysical Research Letters},
  volume={48},
  number={2},
  pages={e2020GL090864},
  year={2021},
  publisher={Wiley Online Library}
}

@article{COGLIATI_2021,
title = {The PRISMA imaging spectroscopy mission: overview and first performance analysis},
journal = {Remote Sensing of Environment},
volume = {262},
pages = {112499},
year = {2021},
issn = {0034-4257},
doi = {https://doi.org/10.1016/j.rse.2021.112499},
url = {https://www.sciencedirect.com/science/article/pii/S0034425721002170},
author = {S. Cogliati and F. Sarti and L. Chiarantini and M. Cosi and R. Lorusso and E. Lopinto and F. Miglietta and L. Genesio and L. Guanter and A. Damm and S. Pérez-López and D. Scheffler and G. Tagliabue and C. Panigada and U. Rascher and T.P.F. Dowling and C. Giardino and R. Colombo}
}

@INPROCEEDINGS{duren_2020,
       author = {{Duren}, R.~M. and {Guido}, J. and {Herner}, J. and {Rao}, S. and {Green}, R.~O. and {de Belloy}, M. and {Schingler}, R. and {Ardila}, D.~R. and {Thorpe}, A.~K. and {Cusworth}, D. and {Falk}, M. and {Scheehle}, E. and {Vahlsing}, C. and {Ide}, T. and {Asner}, G. and {Lawrence}, R.},
        title = "{Carbon Mapper: global tracking of methane and CO$_{2}$ point-sources}",
    booktitle = {AGU Fall Meeting Abstracts},
         year = 2020,
       series = {AGU Fall Meeting Abstracts},
       volume = {2020},
        month = dec,
          eid = {A247-01},
        pages = {A247-01},
       adsurl = {https://ui.adsabs.harvard.edu/abs/2020AGUFMA247...01D}
}

@article{Storch_2023,
title = {The EnMAP imaging spectroscopy mission towards operations},
journal = {Remote Sensing of Environment},
volume = {294},
pages = {113632},
year = {2023},
issn = {0034-4257},
doi = {https://doi.org/10.1016/j.rse.2023.113632},
url = {https://www.sciencedirect.com/science/article/pii/S0034425723001839},
author = {Tobias Storch and Hans-Peter Honold and Sabine Chabrillat and Martin Habermeyer and Paul Tucker and Maximilian Brell and Andreas Ohndorf and Katrin Wirth and Matthias Betz and Michael Kuchler and Helmut Mühle and Emiliano Carmona and Simon Baur and Martin Mücke and Sebastian Löw and Daniel Schulze and Steffen Zimmermann and Christoph Lenzen and Sebastian Wiesner and Saika Aida and Ralph Kahle and Peter Willburger and Sebastian Hartung and Daniele Dietrich and Nicolae Plesia and Mirco Tegler and Katharina Schork and Kevin Alonso and David Marshall and Birgit Gerasch and Peter Schwind and Miguel Pato and Mathias Schneider and Raquel {de los Reyes} and Maximilian Langheinrich and Julian Wenzel and Martin Bachmann and Stefanie Holzwarth and Nicole Pinnel and Luis Guanter and Karl Segl and Daniel Scheffler and Saskia Foerster and Niklas Bohn and Astrid Bracher and Mariana A. Soppa and Ferran Gascon and Rob Green and Raymond Kokaly and Jose Moreno and Cindy Ong and Manuela Sornig and Ricarda Wernitz and Klaus Bagschik and Detlef Reintsema and Laura {La Porta} and Anke Schickling and Sebastian Fischer}
}

@dataset{Green_2023,
  author       = {Green, R. and Thorpe, A. and Brodrick, P. and Chadwick, D. and Lopez, A. and Elder, C. and Villanueva-Weeks, C. and Fahlen, J. and Coleman, R. W. and Jensen, D. and Bender, H. and Vinckier, Q. and Xiang, C. and Olson-Duvall, W. and Lundeen, S. and Thompson, D.},
  title        = {{EMIT L2B Estimated Methane Plume Complexes 60 m V001}},
  year         = {2023},
  publisher    = {NASA Land Processes Distributed Active Archive Center},
  doi          = {10.5067/EMIT/EMITL2BCH4PLM.001},
  url          = {https://doi.org/10.5067/EMIT/EMITL2BCH4PLM.001},
  note         = {Date Accessed: 2026-01-29}
}

@article{kasuya_2009,
  title={Greenhouse gases observing satellite (GOSAT) program overview and its development status},
  author={Kasuya, Masahiro and Nakajima, Masakatsu and Hamazaki, Takashi},
  journal={Transactions of the Japan Society for Aeronautical and Space Sciences, Space Technology Japan},
  volume={7},
  number={ists26},
  pages={To\_4\_5--To\_4\_10},
  year={2009},
  publisher={THE JAPAN SOCIETY FOR AERONAUTICAL AND SPACE SCIENCES}
}

@Article{Vaughan_2024,
AUTHOR = {Vaughan, A. and Mateo-Garc\'{\i}a, G. and G\'omez-Chova, L. and R\r{u}\v{z}i\v{c}ka, V. and Guanter, L. and Irakulis-Loitxate, I.},
TITLE = {CH4Net: a deep learning model for monitoring methane super-emitters with Sentinel-2 imagery},
JOURNAL = {Atmospheric Measurement Techniques},
VOLUME = {17},
YEAR = {2024},
NUMBER = {9},
PAGES = {2583--2593},
URL = {https://amt.copernicus.org/articles/17/2583/2024/},
DOI = {10.5194/amt-17-2583-2024}
}

@article{Irakulis_2022,
author = {Irakulis-Loitxate, Itziar and Guanter, Luis and Maasakkers, Joannes D. and Zavala-Araiza, Daniel and Aben, Ilse},
title = {Satellites Detect Abatable Super-Emissions in One of the World’s Largest Methane Hotspot Regions},
journal = {Environmental Science \& Technology},
volume = {56},
number = {4},
pages = {2143-2152},
year = {2022},
doi = {10.1021/acs.est.1c04873},
    note ={PMID: 35102741},

URL = { 
    
        https://doi.org/10.1021/acs.est.1c04873
    
    

},
eprint = { 
    
        https://doi.org/10.1021/acs.est.1c04873
    
    

}

}

@unknown{wang_2020,
author = {Wang, Jiayang and Nadarajah, Selvaprabu and Wang, Jingfan and Ravikumar, Arvind},
year = {2020},
month = {11},
pages = {},
title = {A Machine Learning Approach to Methane Emissions Mitigation in the Oil and Gas Industry},
doi = {10.31223/X57W29}
}

@article{rouet_2024,
  title={Automatic detection of methane emissions in multispectral satellite imagery using a vision transformer},
  author={Rouet-Leduc, Bertrand and Hulbert, Claudia},
  journal={Nature Communications},
  volume={15},
  number={1},
  pages={3801},
  year={2024},
  publisher={Nature Publishing Group UK London}
}

@INPROCEEDINGS{Kumar_2020,
  author={Kumar, Satish and Torres, Carlos and Ulutan, Oytun and Ayasse, Alana and Roberts, Dar and Manjunath, B.S.},
  booktitle={2020 IEEE Winter Conference on Applications of Computer Vision (WACV)}, 
  title={Deep Remote Sensing Methods for Methane Detection in Overhead Hyperspectral Imagery}, 
  year={2020},
  volume={},
  number={},
  pages={1765-1774},
  doi={10.1109/WACV45572.2020.9093600}}

@Article{Bruno_2024,
AUTHOR = {Bruno, J. H. and Jervis, D. and Varon, D. J. and Jacob, D. J.},
TITLE = {U-Plume: automated algorithm for plume detection and source quantification by satellite point-source imagers},
JOURNAL = {Atmospheric Measurement Techniques},
VOLUME = {17},
YEAR = {2024},
NUMBER = {9},
PAGES = {2625--2636},
URL = {https://amt.copernicus.org/articles/17/2625/2024/},
DOI = {10.5194/amt-17-2625-2024}
}

@article{JONGARAMRUNGRUANG_2022,
title = {MethaNet – An AI-driven approach to quantifying methane point-source emission from high-resolution 2-D plume imagery},
journal = {Remote Sensing of Environment},
volume = {269},
pages = {112809},
year = {2022},
issn = {0034-4257},
doi = {https://doi.org/10.1016/j.rse.2021.112809},
url = {https://www.sciencedirect.com/science/article/pii/S0034425721005290},
author = {Siraput Jongaramrungruang and Andrew K. Thorpe and Georgios Matheou and Christian Frankenberg}
}

@article{vaughan_2024b,
  title={AI for operational methane emitter monitoring from space},
  author={Vaughan, Anna and Mateo-Garcia, Gonzalo and Irakulis-Loitxate, Itziar and Watine, Marc and Fernandez-Poblaciones, Pablo and Turner, Richard E and Requeima, James and Gorro{\~n}o, Javier and Randles, Cynthia and Caltagirone, Manfredi and others},
  journal={arXiv preprint arXiv:2408.04745},
  year={2024}
}

@Article{Frankenberg_2005,
AUTHOR = {Frankenberg, C. and Platt, U. and Wagner, T.},
TITLE = {Iterative maximum a posteriori (IMAP)-DOAS for retrieval of strongly absorbing trace gases: Model studies for CH$_{4}$ and CO$_{2}$ retrieval from near infrared spectra of SCIAMACHY onboard ENVISAT},
JOURNAL = {Atmospheric Chemistry and Physics},
VOLUME = {5},
YEAR = {2005},
NUMBER = {1},
PAGES = {9--22},
URL = {https://acp.copernicus.org/articles/5/9/2005/},
DOI = {10.5194/acp-5-9-2005}
}

@Article{Krings_2011,
AUTHOR = {Krings, T. and Gerilowski, K. and Buchwitz, M. and Reuter, M. and Tretner, A. and Erzinger, J. and Heinze, D. and Pfl\"uger, U. and Burrows, J. P. and Bovensmann, H.},
TITLE = {MAMAP – a new spectrometer system for column-averaged methane and carbon dioxide observations from aircraft: retrieval algorithm and first inversions for point source emission rates},
JOURNAL = {Atmospheric Measurement Techniques},
VOLUME = {4},
YEAR = {2011},
NUMBER = {9},
PAGES = {1735--1758},
URL = {https://amt.copernicus.org/articles/4/1735/2011/},
DOI = {10.5194/amt-4-1735-2011}
}

@Article{Zhang_2022,
AUTHOR = {Zhang, Z. and Sherwin, E. D. and Varon, D. J. and Brandt, A. R.},
TITLE = {Detecting and quantifying methane emissions from oil and gas production: algorithm development with ground-truth calibration based on Sentinel-2 satellite imagery},
JOURNAL = {Atmospheric Measurement Techniques},
VOLUME = {15},
YEAR = {2022},
NUMBER = {23},
PAGES = {7155--7169},
URL = {https://amt.copernicus.org/articles/15/7155/2022/},
DOI = {10.5194/amt-15-7155-2022}
}

@ARTICLE{perezcarrasco_2025,
  author={Pérez-Carrasco, Manuel and Nasr, Maya and Roche, Sebastien and Miller, Chris Chan and Zhang, Zhan and Park, Core Francisco and Walker, Eleanor and Garraffo, Cecilia and Finkbeiner, Douglas and Ayvazov, Sasha and Franklin, Jonathan and Luo, Bingkun and Liu, Xiong and Gautam, Ritesh and Wofsy, Steven},
  journal={IEEE Transactions on Geoscience and Remote Sensing}, 
  title={Deep Learning for Clouds and Cloud Shadow Segmentation in Methane Satellite and Airborne Imaging Spectroscopy}, 
  year={2026},
  volume={},
  number={},
  pages={1-1},
  doi={10.1109/TGRS.2026.3672371}}

@INPROCEEDINGS{Wofsy_2019,
       author = {{Wofsy}, S.~C. and {Hamburg}, S.},
        title = "{MethaneSAT - A New Observing Platform For High Resolution Measurements Of Methane and Carbon Dioxide}",
    booktitle = {AGU Fall Meeting Abstracts},
         year = 2019,
       series = {AGU Fall Meeting Abstracts},
       volume = {2019},
        month = dec,
          eid = {A53F-02},
        pages = {A53F-02},
       adsurl = {https://ui.adsabs.harvard.edu/abs/2019AGUFM.A53F..02W}
}

@article{Ren_2015,
  title={Faster r-cnn: Towards real-time object detection with region proposal networks},
  author={Ren, Shaoqing and He, Kaiming and Girshick, Ross and Sun, Jian},
  journal={Advances in neural information processing systems},
  volume={28},
  year={2015}
}

@Article{joyce_2023,
AUTHOR = {Joyce, P. and Ruiz Villena, C. and Huang, Y. and Webb, A. and Gloor, M. and Wagner, F. H. and Chipperfield, M. P. and Barrio Guill\'o, R. and Wilson, C. and Boesch, H.},
TITLE = {Using a deep neural network to detect methane point sources and quantify emissions from PRISMA hyperspectral satellite images},
JOURNAL = {Atmospheric Measurement Techniques},
VOLUME = {16},
YEAR = {2023},
NUMBER = {10},
PAGES = {2627--2640},
URL = {https://amt.copernicus.org/articles/16/2627/2023/},
DOI = {10.5194/amt-16-2627-2023}
}

@article{radman_2023,
title = {S2MetNet: A novel dataset and deep learning benchmark for methane point source quantification using Sentinel-2 satellite imagery},
journal = {Remote Sensing of Environment},
volume = {295},
pages = {113708},
year = {2023},
issn = {0034-4257},
doi = {https://doi.org/10.1016/j.rse.2023.113708},
url = {https://www.sciencedirect.com/science/article/pii/S0034425723002596},
author = {Ali Radman and Masoud Mahdianpari and Daniel J. Varon and Fariba Mohammadimanesh}
}

@Article{Schuit_2023,
AUTHOR = {Schuit, B. J. and Maasakkers, J. D. and Bijl, P. and Mahapatra, G. and van den Berg, A.-W. and Pandey, S. and Lorente, A. and Borsdorff, T. and Houweling, S. and Varon, D. J. and McKeever, J. and Jervis, D. and Girard, M. and Irakulis-Loitxate, I. and Gorro\~no, J. and Guanter, L. and Cusworth, D. H. and Aben, I.},
TITLE = {Automated detection and monitoring of methane super-emitters using satellite data},
JOURNAL = {Atmospheric Chemistry and Physics},
VOLUME = {23},
YEAR = {2023},
NUMBER = {16},
PAGES = {9071--9098},
URL = {https://acp.copernicus.org/articles/23/9071/2023/},
DOI = {10.5194/acp-23-9071-2023}
}

@Article{Varon_2018,
AUTHOR = {Varon, D. J. and Jacob, D. J. and McKeever, J. and Jervis, D. and Durak, B. O. A. and Xia, Y. and Huang, Y.},
TITLE = {Quantifying methane point sources from fine-scale satellite observations of
atmospheric methane plumes},
JOURNAL = {Atmospheric Measurement Techniques},
VOLUME = {11},
YEAR = {2018},
NUMBER = {10},
PAGES = {5673--5686},
URL = {https://amt.copernicus.org/articles/11/5673/2018/},
DOI = {10.5194/amt-11-5673-2018}
}

@Article{lauvaux_constraining_2012,
	title = {Constraining the CO2 budget of the corn belt: exploring uncertainties from the assumptions in a mesoscale inverse system},
	volume = {12},
	issn = {1680-7324},
	shorttitle = {Constraining the {CO}\&lt;sub\&gt;2\&lt;/sub\&gt; budget of the corn belt},
	url = {https://acp.copernicus.org/articles/12/337/2012/},
	doi = {10.5194/acp-12-337-2012},
	language = {en},
	number = {1},
	urldate = {2021-10-08},
	journal = {Atmospheric Chemistry and Physics},
	author = {Lauvaux, T. and Schuh, A. E. and Uliasz, M. and Richardson, S. and Miles, N. and Andrews, A. E. and Sweeney, C. and Diaz, L. I. and Martins, D. and Shepson, P. B. and Davis, K. J.},
	month = jan,
	year = {2012},
	pages = {337--354},
}

@misc{EDF_2025,
  author       = {{Environmental Defense Fund}},
  title        = {{MethaneSAT} Loses Contact with Satellite},
  year         = {2025},
  month        = jul,
  day          = {1},
  howpublished = {Press release},
  url          = {https://www.edf.org/media/methanesat-loses-contact-satellite},
  note         = {Accessed: 2026-03-03},
  organization = {Environmental Defense Fund}
}

@Article{Duren_2025,
AUTHOR = {Duren, R. and Cusworth, D. and Ayasse, A. and Howell, K. and Diamond, A. and Scarpelli, T. and Kim, J. and O'neill, K. and Lai-Norling, J. and Thorpe, A. and Zandbergen, S. R. and Shaw, L. and Keremedjiev, M. and Guido, J. and Giuliano, P. and Goldstein, M. and Nallapu, R. and Barentsen, G. and Thompson, D. R. and Roth, K. and Jensen, D. and Eastwood, M. and Reuland, F. and Adams, T. and Brandt, A. and Kort, E. A. and Mason, J. and Green, R. O.},
TITLE = {The Carbon Mapper emissions monitoring system},
JOURNAL = {EGUsphere},
VOLUME = {2025},
YEAR = {2025},
PAGES = {1--41},
URL = {https://egusphere.copernicus.org/preprints/2025/egusphere-2025-2275/},
DOI = {10.5194/egusphere-2025-2275}
}

@article{vaswani_2017,
  title={Attention is all you need},
  author={Vaswani, Ashish and Shazeer, Noam and Parmar, Niki and Uszkoreit, Jakob and Jones, Llion and Gomez, Aidan N and Kaiser, {\L}ukasz and Polosukhin, Illia},
  journal={Advances in neural information processing systems},
  volume={30},
  year={2017}
}

@Article{Cusworth_2018,
AUTHOR = {Cusworth, D. H. and Jacob, D. J. and Sheng, J.-X. and Benmergui, J. and Turner, A. J. and Brandman, J. and White, L. and Randles, C. A.},
TITLE = {Detecting high-emitting methane sources in oil/gas fields using satellite
observations},
JOURNAL = {Atmospheric Chemistry and Physics},
VOLUME = {18},
YEAR = {2018},
NUMBER = {23},
PAGES = {16885--16896},
URL = {https://acp.copernicus.org/articles/18/16885/2018/},
DOI = {10.5194/acp-18-16885-2018}
}

@article{foote_2020,
  title={Fast and accurate retrieval of methane concentration from imaging spectrometer data using sparsity prior},
  author={Foote, Markus D and Dennison, Philip E and Thorpe, Andrew K and Thompson, David R and Jongaramrungruang, Siraput and Frankenberg, Christian and Joshi, Sarang C},
  journal={IEEE Transactions on Geoscience and Remote Sensing},
  volume={58},
  number={9},
  pages={6480--6492},
  year={2020},
  publisher={IEEE}
}

@article{Roger_2024,
author = {Roger, Javier and Irakulis-Loitxate, Itziar and Valverde, Adriana and Gorroño, Javier and Chabrillat, Sabine and Brell, Max and Guanter, Luis},
year = {2024},
month = {01},
pages = {1-1},
title = {High-Resolution Methane Mapping With the EnMAP Satellite Imaging Spectroscopy Mission},
volume = {PP},
journal = {IEEE Transactions on Geoscience and Remote Sensing},
doi = {10.1109/TGRS.2024.3352403}
}

@Article{sanchezgarcia_2022,
AUTHOR = {S\'anchez-Garc\'{\i}a, E. and Gorro\~no, J. and Irakulis-Loitxate, I. and Varon, D. J. and Guanter, L.},
TITLE = {Mapping methane plumes at very high spatial resolution with the WorldView-3 satellite},
JOURNAL = {Atmospheric Measurement Techniques},
VOLUME = {15},
YEAR = {2022},
NUMBER = {6},
PAGES = {1657--1674},
URL = {https://amt.copernicus.org/articles/15/1657/2022/},
DOI = {10.5194/amt-15-1657-2022}
}

@techreport{iea_2023,
  author      = {{IEA}},
  title       = {Global Methane Tracker 2023},
  institution = {International Energy Agency},
  year        = {2023},
  address     = {Paris},
  url         = {https://www.iea.org/reports/global-methane-tracker-2023},
  note        = {Licence: CC BY 4.0 (report); CC BY NC SA 4.0 (Annex A)}
}

@article{ahsan_2025,
  title={AttMetNet: Attention-Enhanced Deep Neural Network for Methane Plume Detection in Sentinel-2 Satellite Imagery},
  author={Ahsan, Rakib and Shanto, MD and Arifin, Md Sultanul and Hashem, Tanzima},
  journal={arXiv preprint arXiv:2512.02751},
  year={2025}
}

@article{ruvzivcka_2025,
  title={HyperspectralViTs: General hyperspectral models for on-board remote sensing},
   author={Růžička, V. and Markham, A.},
  journal={IEEE Journal of Selected Topics in Applied Earth Observations and Remote Sensing},
  year={2025},
  publisher={IEEE}
}

@article{si_2024,
  title={Unlocking the potential: Multi-task deep learning for spaceborne quantitative monitoring of fugitive methane plumes},
  author={Si, Guoxin and Fu, Shiliang and Yao, Wei},
  journal={arXiv preprint arXiv:2401.12870},
  year={2024}
}

@inproceedings{zhu_2017,
  title={Unpaired image-to-image translation using cycle-consistent adversarial networks},
  author={Zhu, Jun-Yan and Park, Taesung and Isola, Phillip and Efros, Alexei A},
  booktitle={Proceedings of the IEEE international conference on computer vision},
  pages={2223--2232},
  year={2017}
}

@article{xie_2021,
  title={SegFormer: Simple and efficient design for semantic segmentation with transformers},
  author={Xie, Enze and Wang, Wenhai and Yu, Zhiding and Anandkumar, Anima and Alvarez, Jose M and Luo, Ping},
  journal={Advances in neural information processing systems},
  volume={34},
  pages={12077--12090},
  year={2021}
}

@inproceedings{liu_2023,
  title={Efficientvit: Memory efficient vision transformer with cascaded group attention},
  author={Liu, Xinyu and Peng, Houwen and Zheng, Ningxin and Yang, Yuqing and Hu, Han and Yuan, Yixuan},
  booktitle={Proceedings of the IEEE/CVF conference on computer vision and pattern recognition},
  pages={14420--14430},
  year={2023}
}

@article{loshchilov_2017,
  title={Decoupled weight decay regularization},
  author={Loshchilov, Ilya and Hutter, Frank},
  journal={arXiv preprint arXiv:1711.05101},
  year={2017}
}

@article{ganin_2016,
  title={Domain-adversarial training of neural networks},
  author={Ganin, Yaroslav and Ustinova, Evgeniya and Ajakan, Hana and Germain, Pascal and Larochelle, Hugo and Laviolette, Fran{\c{c}}ois and March, Mario and Lempitsky, Victor},
  journal={Journal of machine learning research},
  volume={17},
  number={59},
  pages={1--35},
  year={2016}
}

@inproceedings{zhao_2019,
  title={On learning invariant representations for domain adaptation},
  author={Zhao, Han and Des Combes, Remi Tachet and Zhang, Kun and Gordon, Geoffrey},
  booktitle={International conference on machine learning},
  pages={7523--7532},
  year={2019},
  organization={PMLR}
}

@inproceedings{kamnitsas_2017,
  title={Unsupervised domain adaptation in brain lesion segmentation with adversarial networks},
  author={Kamnitsas, Konstantinos and Baumgartner, Christian and Ledig, Christian and Newcombe, Virginia and Simpson, Joanna and Kane, Andrew and Menon, David and Nori, Aditya and Criminisi, Antonio and Rueckert, Daniel and others},
  booktitle={International conference on information processing in medical imaging},
  pages={597--609},
  year={2017},
  organization={Springer}
}

@inproceedings{Ottenheimer2026,
  author       = {Ottenheimer, Raia and Roper, Laine and Zhang, Zhan and Sargent, Maryann and Warren, Jack and Roche, Sébastien and Chan Miller, Christopher and Kyzivat, Ethan and Knapp, Marvin and Benmergui, Joshua and Pittman, Jasna V. and Samra, Jenna and Hawthrone, Jacob and Walker, Eleanor and Chulakadabba, Apisada and Manninen, Ethan and Bushey, Jacob and Luo, Bingkun and Miller, David J. and Nasr, Maya and Sun, Kang and Franklin, Jonathan and Liu, Xiong and Wofsy, Steven C.},
  title        = {From Shadows to Signals: A Novel Statistical Method for Discriminating Albedo Artifact-Induced False Methane Plumes},
  booktitle    = {106th American Meteorological Society Annual Meeting},
  year         = {2026},
  organization = {American Meteorological Society},
  address      = {Houston, TX, USA},
  note         = {Poster presented at the 106th AMS Annual Meeting, January 25--29, 2026},
  url          = {https://ams.confex.com/ams/106ANNUAL/meetingapp.cgi/Paper/470925}}

@article{yosinski_2014,
  title={How transferable are features in deep neural networks?},
  author={Yosinski, Jason and Clune, Jeff and Bengio, Yoshua and Lipson, Hod},
  journal={Advances in neural information processing systems},
  volume={27},
  year={2014}
}

@inproceedings{bengio_2009,
author = {Bengio, Yoshua and Louradour, J\'{e}r\^{o}me and Collobert, Ronan and Weston, Jason},
title = {Curriculum learning},
year = {2009},
isbn = {9781605585161},
publisher = {Association for Computing Machinery},
address = {New York, NY, USA},
url = {https://doi.org/10.1145/1553374.1553380},
doi = {10.1145/1553374.1553380},
booktitle = {Proceedings of the 26th Annual International Conference on Machine Learning},
pages = {41–48},
numpages = {8},
location = {Montreal, Quebec, Canada},
series = {ICML '09}
}

@inproceedings{ester1996density,
  title={A density-based algorithm for discovering clusters in large spatial databases with noise},
  author={Ester, Martin and Kriegel, Hans-Peter and Sander, J{\"o}rg and Xu, Xiaowei and others},
  booktitle={kdd},
  volume={96},
  number={34},
  pages={226--231},
  year={1996}
}

@Article{zhang_2026,
AUTHOR = {Zhang, Z. and Sargent, M. and Manninen, E. and Warren, J. D. and Chulakadabba, A. and Russi, M. and Ayvazov, S. and Benmergui, J. and Knapp, M. and Kyzivat, E. and Miller, C. C. and Roche, S. and Luo, B. and Miller, D. J. and Nasr, M. and Perez-Carrasco, M. and Sun, K. and Williams, J. P. and MacKay, K. and Omara, M. and Chen, J. and Guanter, L. and Gautam, R. and Franklin, J. and Liu, X. and Wofsy, S. C.},
TITLE = {Automatic methane plume masking based on wavelet transform image processing: application to MethaneAIR and MethaneSAT data},
JOURNAL = {Atmospheric Measurement Techniques},
VOLUME = {19},
YEAR = {2026},
NUMBER = {14},
PAGES = {4637--4651},
URL = {https://amt.copernicus.org/articles/19/4637/2026/},
DOI = {10.5194/amt-19-4637-2026}
}

@Article{Guanter_2026,
AUTHOR = {Guanter, L. and Roger, J. and Warren, J. and Sargent, M. and Zhang, Z. and Roche, S. and Miller, C. C. and Steiner, M. and Hadfield, H. and Omara, M. and Williams, J. and MacKay, K. and Franklin, J. E. and Luo, B. and Wofsy, S. C. and Hamburg, S. P. and Gautam, R.},
TITLE = {Surveying methane point-source super-emissions across oil and gas basins with MethaneSAT},
JOURNAL = {Atmospheric Chemistry and Physics},
VOLUME = {26},
YEAR = {2026},
NUMBER = {4},
PAGES = {2941--2963},
URL = {https://acp.copernicus.org/articles/26/2941/2026/},
DOI = {10.5194/acp-26-2941-2026}
}

@Article{Sun_2018,
AUTHOR = {Sun, K. and Zhu, L. and Cady-Pereira, K. and Chan Miller, C. and Chance, K. and Clarisse, L. and Coheur, P.-F. and Gonz\'alez Abad, G. and Huang, G. and Liu, X. and Van Damme, M. and Yang, K. and Zondlo, M.},
TITLE = {A physics-based approach to oversample multi-satellite, multispecies observations to a common grid},
JOURNAL = {Atmospheric Measurement Techniques},
VOLUME = {11},
YEAR = {2018},
NUMBER = {12},
PAGES = {6679--6701},
URL = {https://amt.copernicus.org/articles/11/6679/2018/},
DOI = {10.5194/amt-11-6679-2018}
}

@Article{Thorpe_2014,
AUTHOR = {Thorpe, A. K. and Frankenberg, C. and Roberts, D. A.},
TITLE = {Retrieval techniques for airborne imaging of methane concentrations using high spatial and moderate spectral resolution: application to AVIRIS},
JOURNAL = {Atmospheric Measurement Techniques},
VOLUME = {7},
YEAR = {2014},
NUMBER = {2},
PAGES = {491--506},
URL = {https://amt.copernicus.org/articles/7/491/2014/},
DOI = {10.5194/amt-7-491-2014}
}

@Article{Guanter_2025,
AUTHOR = {Guanter, L. and Warren, J. and Omara, M. and Chulakadabba, A. and Roger, J. and Sargent, M. and Franklin, J. E. and Wofsy, S. C. and Gautam, R.},
TITLE = {Detection and quantification of methane plumes with the MethaneAIR airborne spectrometer},
JOURNAL = {Atmospheric Measurement Techniques},
VOLUME = {18},
YEAR = {2025},
NUMBER = {15},
PAGES = {3857--3872},
URL = {https://amt.copernicus.org/articles/18/3857/2025/},
DOI = {10.5194/amt-18-3857-2025}
}

@Article{Ouerghi_2025,
AUTHOR = {Ouerghi, E. and Ehret, T. and Facciolo, G. and Meinhardt, E. and Marion, R. and Morel, J.-M.},
TITLE = {Tightening up methane plume source rate estimation in EnMAP and PRISMA images},
JOURNAL = {Atmospheric Measurement Techniques},
VOLUME = {18},
YEAR = {2025},
NUMBER = {18},
PAGES = {4611--4629},
URL = {https://amt.copernicus.org/articles/18/4611/2025/},
DOI = {10.5194/amt-18-4611-2025}
}

@Article{Warren_2025,
AUTHOR = {Warren, J. D. and Sargent, M. and Williams, J. P. and Omara, M. and Miller, C. C. and Roche, S. and MacKay, K. and Manninen, E. and Chulakadabba, A. and Himmelberger, A. and Benmergui, J. and Zhang, Z. and Guanter, L. and Wofsy, S. and Gautam, R.},
TITLE = {Sectoral contributions of high-emitting methane point sources from major US onshore oil and gas producing basins using airborne measurements from MethaneAIR},
JOURNAL = {Atmospheric Chemistry and Physics},
VOLUME = {25},
YEAR = {2025},
NUMBER = {18},
PAGES = {10661--10675},
URL = {https://acp.copernicus.org/articles/25/10661/2025/},
DOI = {10.5194/acp-25-10661-2025}
}

@Article{Williams_2025,
AUTHOR = {Williams, J. P. and Omara, M. and Himmelberger, A. and Zavala-Araiza, D. and MacKay, K. and Benmergui, J. and Sargent, M. and Wofsy, S. C. and Hamburg, S. P. and Gautam, R.},
TITLE = {Small emission sources in aggregate disproportionately account for a large majority of total methane emissions from the US oil and gas sector},
JOURNAL = {Atmospheric Chemistry and Physics},
VOLUME = {25},
YEAR = {2025},
NUMBER = {3},
PAGES = {1513--1532},
URL = {https://acp.copernicus.org/articles/25/1513/2025/},
DOI = {10.5194/acp-25-1513-2025}
}

@article{Omara_2022,
  author  = {Omara, Mark and Zavala-Araiza, Daniel and Lyon, David R. and Hmiel, Benjamin and Roberts, Katherine A. and Hamburg, Steven P.},
  title   = {Methane emissions from {US} low production oil and natural gas well sites},
  journal = {Nature Communications},
  year    = {2022},
  volume  = {13},
  number  = {1},
  pages   = {2085},
  doi     = {10.1038/s41467-022-29709-3},
  url     = {https://doi.org/10.1038/s41467-022-29709-3},
  issn    = {2041-1723}
}

@article{Thompson_2016,
author = {Thompson, D. R. and Thorpe, A. K. and Frankenberg, C. and Green, R. O. and Duren, R. and Guanter, L. and Hollstein, A. and Middleton, E. and Ong, L. and Ungar, S.},
title = {Space-based remote imaging spectroscopy of the Aliso Canyon CH4 superemitter},
journal = {Geophysical Research Letters},
volume = {43},
number = {12},
pages = {6571-6578},
doi = {https://doi.org/10.1002/2016GL069079},
url = {https://agupubs.onlinelibrary.wiley.com/doi/abs/10.1002/2016GL069079},
eprint = {https://agupubs.onlinelibrary.wiley.com/doi/pdf/10.1002/2016GL069079},
year = {2016}
}

@article{batchu_2026,
  title={Global monitoring of methane point sources using deep learning on hyperspectral radiance measurements from EMIT},
  author={Batchu, Vishal V and Conserva, Michelangelo and Wilson, Alex and Michalak, Anna M and Gulshan, Varun and Brodrick, Philip G and Thorpe, Andrew K and Arsdale, Christopher V},
  journal={arXiv preprint arXiv:2604.10094},
  year={2026}
}

@inproceedings{chen2018,
  title={Encoder-decoder with atrous separable convolution for semantic image segmentation},
  author={Chen, Liang-Chieh and Zhu, Yukun and Papandreou, George and Schroff, Florian and Adam, Hartwig},
  booktitle={European conference on computer vision},
  pages={833--851},
  year={2018},
  organization={Springer}
}

@article{oksuz_2021,
  title={Imbalance problems in object detection: A review},
  author={Oksuz, Kemal and Cam, Baris Can and Kalkan, Sinan and Akbas, Emre},
  journal={IEEE transactions on pattern analysis and machine intelligence},
  volume={43},
  number={10},
  pages={3388--3415},
  year={2020},
  publisher={IEEE},
  doi={10.1109/TPAMI.2020.2981890}
}

@inproceedings{shrivastava_2016,
  title={Training region-based object detectors with online hard example mining},
  author={Shrivastava, Abhinav and Gupta, Abhinav and Girshick, Ross},
  booktitle={Proceedings of the IEEE conference on computer vision and pattern recognition},
  pages={761--769},
  year={2016}
}

@inproceedings{girshick_2015,
  title={Fast r-cnn},
  author={Girshick, Ross},
  booktitle={Proceedings of the IEEE international conference on computer vision},
  pages={1440--1448},
  year={2015}
}

@inproceedings{lin_2020,
  title={Focal loss for dense object detection},
  author={Lin, Tsung-Yi and Goyal, Priya and Girshick, Ross and He, Kaiming and Doll{\'a}r, Piotr},
  booktitle={Proceedings of the IEEE international conference on computer vision},
  pages={2980--2988},
  year={2017},
}

@inproceedings{cui_2019,
  title={Class-balanced loss based on effective number of samples},
  author={Cui, Yin and Jia, Menglin and Lin, Tsung-Yi and Song, Yang and Belongie, Serge},
  booktitle={2019 IEEE/CVF conference on computer vision and pattern recognition (CVPR)},
  pages={9260--9269},
  year={2019},
  organization={IEEE}
}

@inproceedings{gupta_2019,
  title={LVIS: A dataset for large vocabulary instance segmentation},
  author={Gupta, Agrim and Dollar, Piotr and Girshick, Ross},
  booktitle={2019 IEEE/CVF Conference on Computer Vision and Pattern Recognition (CVPR)},
  pages={5351--5359},
  year={2019},
  organization={IEEE}
}

\vskip -2\baselineskip plus -1fil
\begin{IEEEbiography}[{\includegraphics[width=1in,height=1.25in,clip,keepaspectratio]{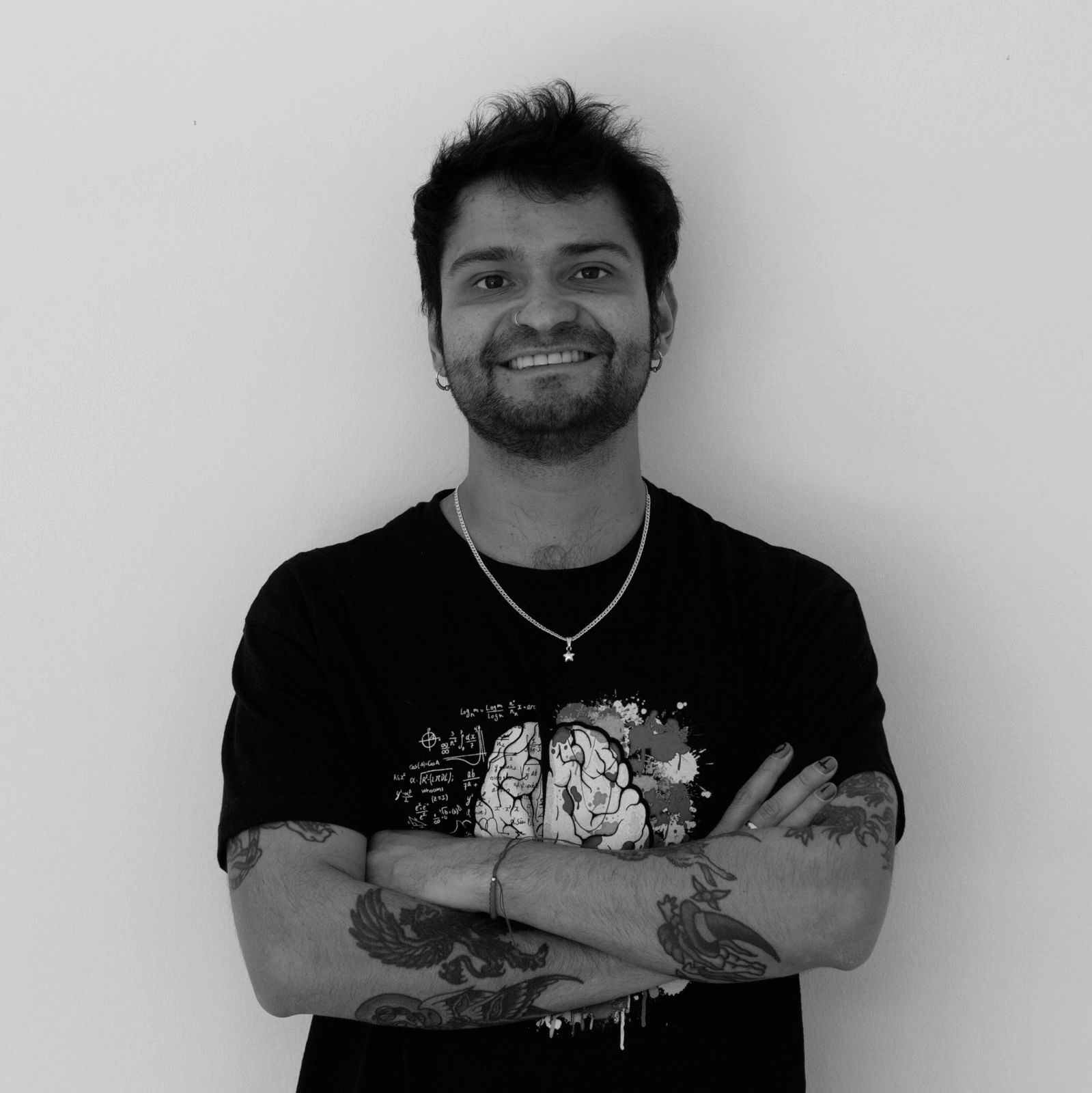}}]{Manuel Pérez-Carrasco}
is a PhD student at the Technical University of Munich and former SPARK Postbaccalaureate Fellow at the Center for Astrophysics $\vert$ Harvard \& Smithsonian. He received the Master's degree in Computer Science from the University of Concepción, Chile. He works on machine learning models for remote sensing.
\end{IEEEbiography}

\vskip -2\baselineskip plus -1fil
\begin{IEEEbiography}[{\includegraphics[width=1in,height=1.25in,clip,keepaspectratio]{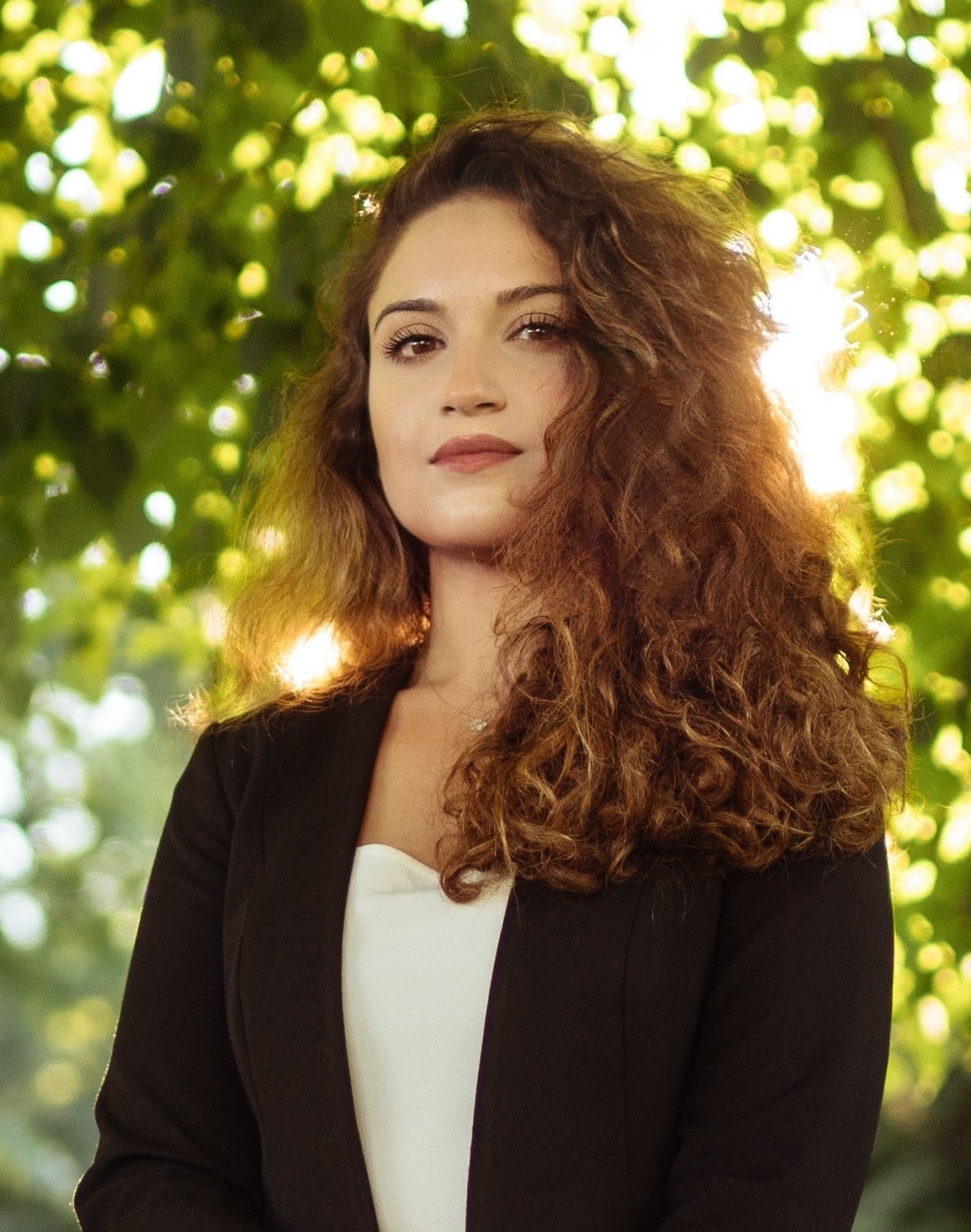}}]{Maya Nasr}
is a Science Engineer at Environmental Defense Fund and Harvard University, where she leads in-orbit lunar calibrations and the team developing machine learning models for the MethaneSAT mission. She brings expertise in space operations, mission planning, systems engineering, technology strategy, and space law and policy. She previously worked on space projects including NASA's Mars 2020 Perseverance rover mission, Cassini's mission activity on Titan, and the OneWeb satellites network. She holds Bachelor's, Master's, and Ph.D. degrees in Aerospace Engineering from MIT.
\end{IEEEbiography}

\vskip -2\baselineskip plus -1fil
\begin{IEEEbiography}[{\includegraphics[width=1in,height=1.25in,clip,keepaspectratio]{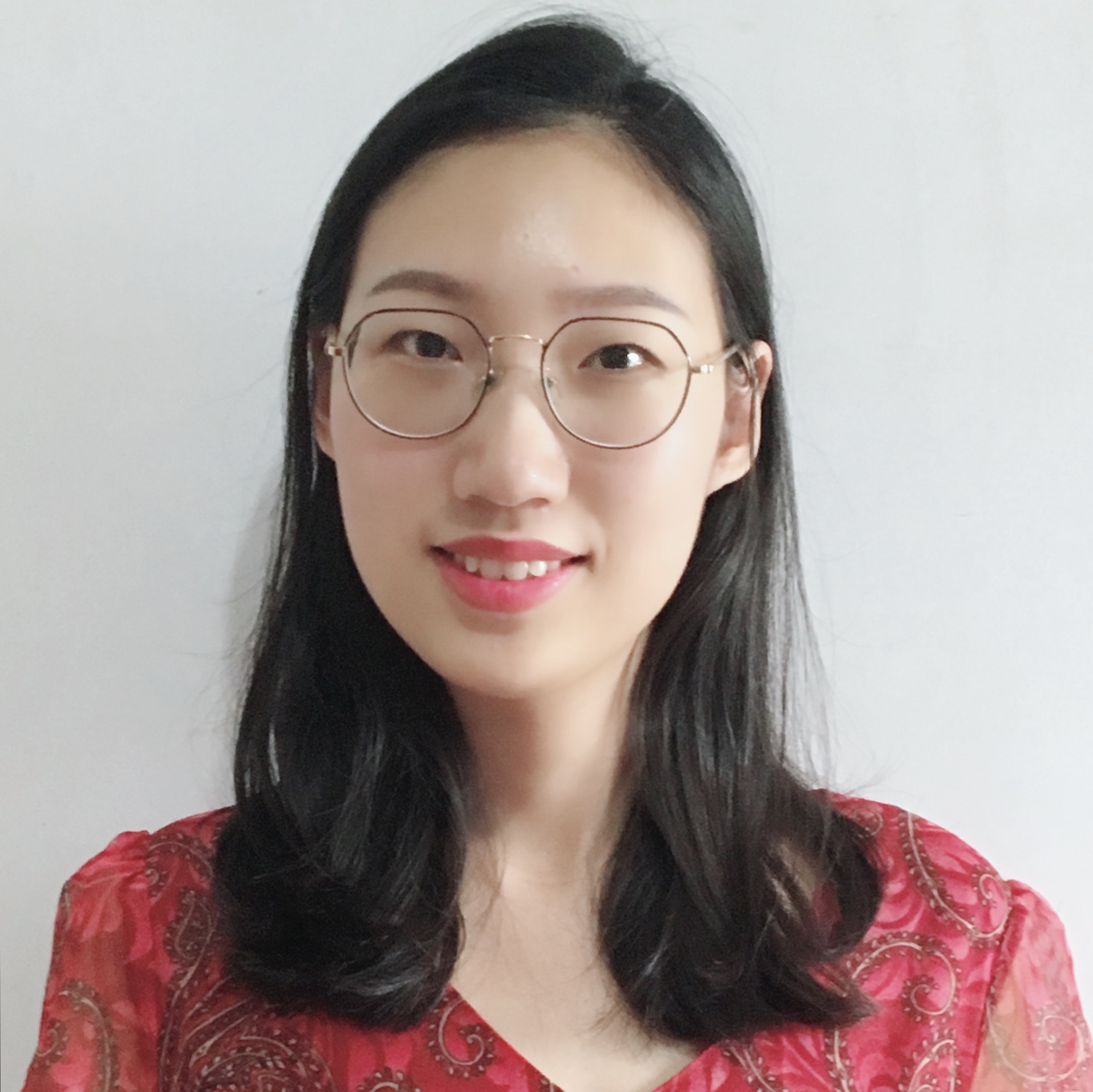}}]{Zhan Zhang}
received the B.S. degree in Environmental Science from Beijing Normal University, China, in 2016, and the M.S. degree in Ecology from Chinese Academy of Sciences, China, in 2019, and the Ph.D. degree in Energy Resources Engineering from Stanford University, USA, in 2024.
From 2024 to 2025, she served as the postdoctoral research fellow in Harvard University. Since 2025, she works as the Scientist, MethaneSAT Science in Environmental Defense Fund, where she mainly works on remote sensing methane emissions detection using MethaneSAT and MethaneAIR. Her research interest lies in remote sensing greenhouse gas emissions detection based on various satellite and airborne platforms.
Dr. Zhang has authored and coauthored more than 10 publications in high-impact international journals, including Geophysical Research Letters and Nature. She also serves as a reviewer for high-ranking journals including Remote Sensing of Environment and Environmental Science \& Technology.
\end{IEEEbiography}

\vskip -2\baselineskip plus -1fil
\begin{IEEEbiography}[{\includegraphics[width=1in,height=1.25in,clip,keepaspectratio]{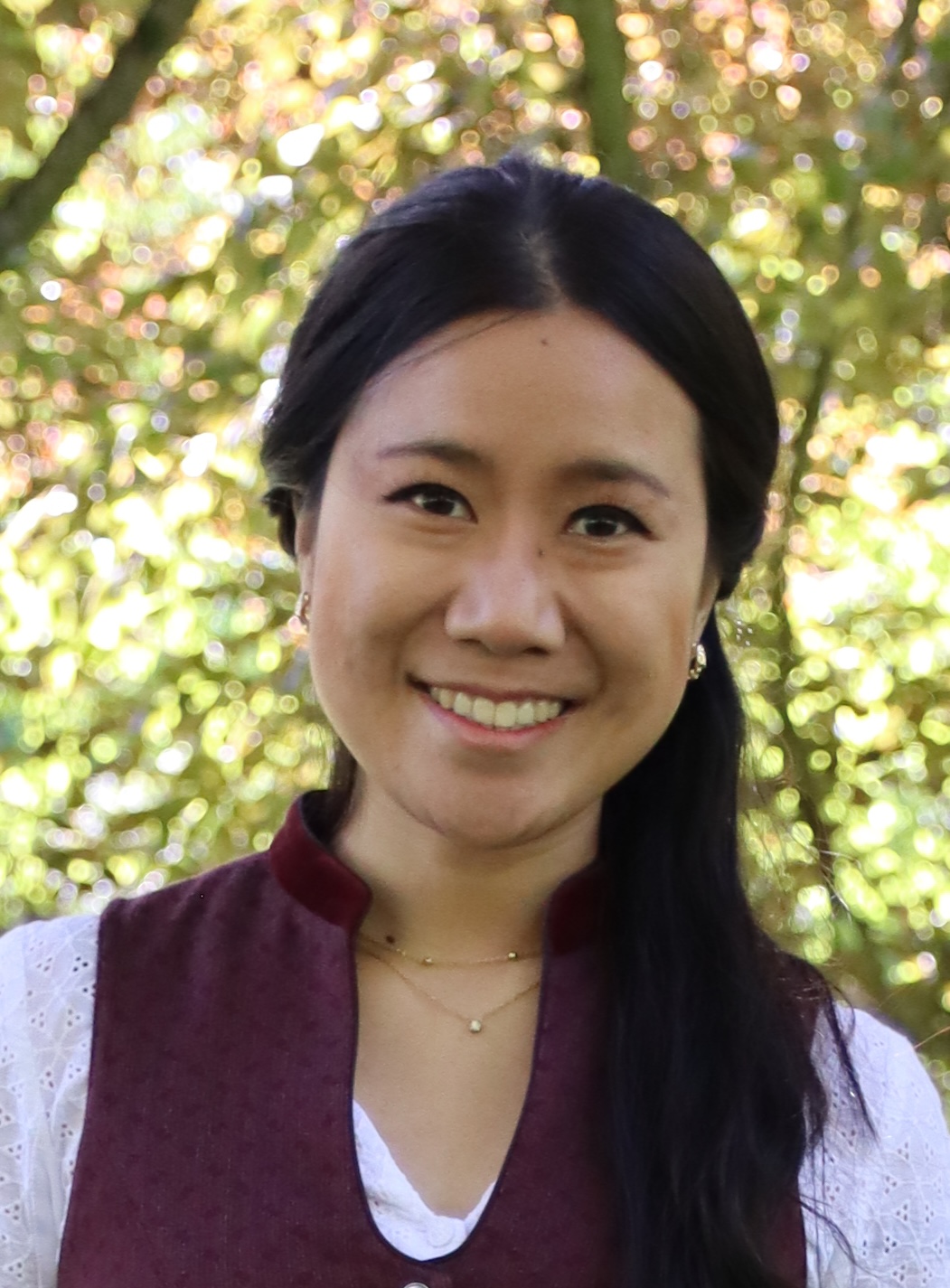}}]{Apisada Chulakadabba} is a postdoctoral researcher at the Technical University of Munich and a research associate at Harvard. She received the B.S. degree from MIT and the M.S. and Ph.D. degrees from Harvard University in Cambridge, MA. Her research focuses on methane emissions quantification using satellite, airborne, and ground-based observations, with an emphasis on inverse modeling and observing system simulation experiments.
\end{IEEEbiography}

\vskip -2\baselineskip plus -1fil
\begin{IEEEbiography}[{\includegraphics[width=1in,height=1.25in,clip,keepaspectratio]{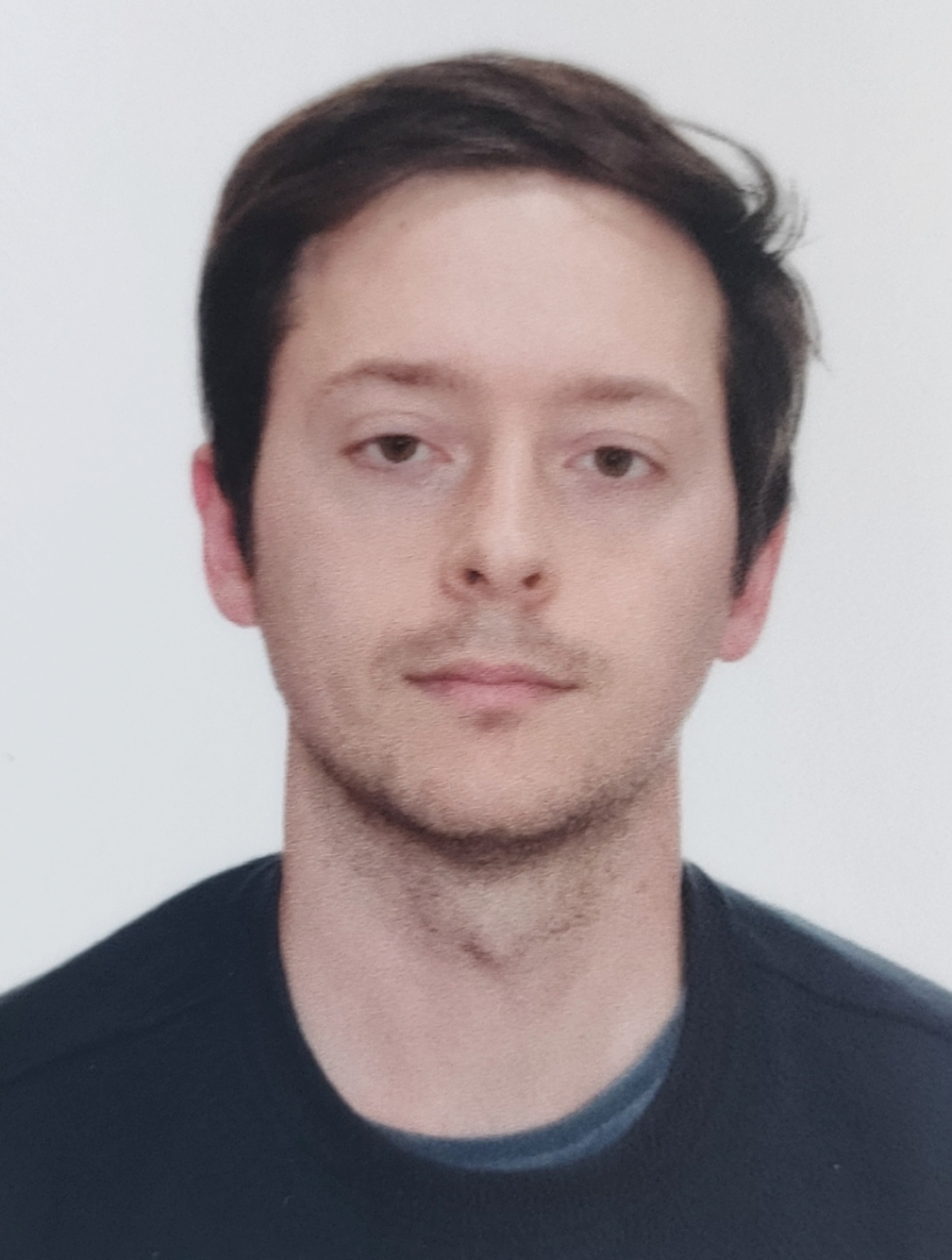}}]{Javier Roger} received the B.Sc. degree in Physics and the M.Sc. degree in Remote Sensing from the Universitat de València (UV), Valencia, Spain, in 2018 and 2021, respectively. He then completed the Ph.D. degree in Geomatics Engineering with the Land and Atmosphere Remote Sensing (LARS) Group at the Universitat Politècnica de València (UPV), Valencia, Spain. He currently holds a position as a Postdoctoral Researcher at the University of Bremen within the Greenhouse Gas Group of the Institute of Environmental Physics. His research focuses on the improvement of trace gas detection and quantification methods applied to hyperspectral data from missions such as the Environmental Mapping and Analysis Program (EnMAP) and the PRecursore IperSpettrale della Missione Applicativa (PRISMA). His objective is to develop robust tools for monitoring trace gases worldwide.
\end{IEEEbiography}

\vskip -2\baselineskip plus -1fil
\begin{IEEEbiography}[{\includegraphics[width=1in,height=1.25in,clip,keepaspectratio]{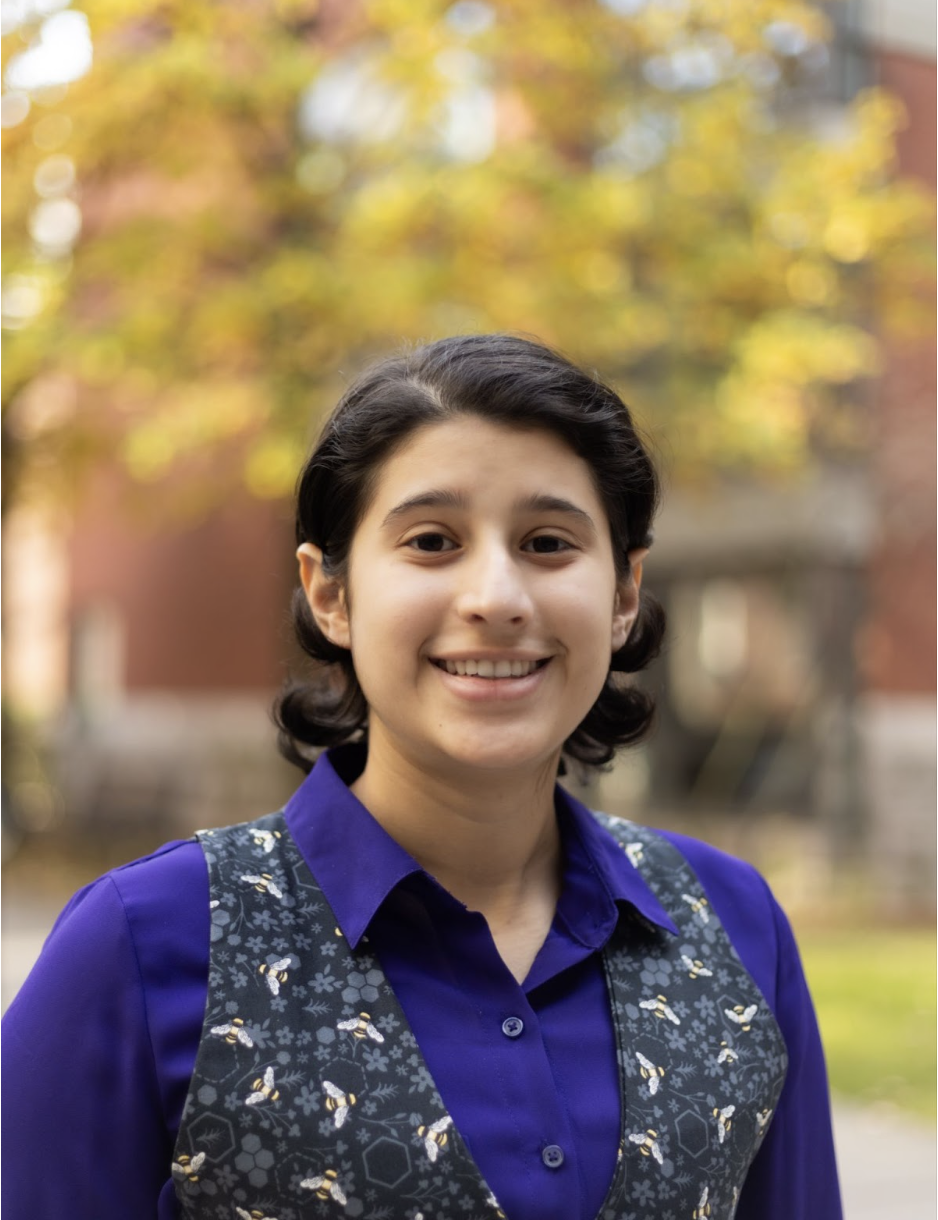}}]{Raia Ottenheimer} is pursuing a Ph.D. in Environmental Science and Engineering in the Wofsy Group at Harvard University. She is currently working on classifying spurious and diffuse plumes in MethaneAIR and MethaneSAT observations. Raia has received an H.B.Sc. in Biological Physics from the University of Toronto (2024), and an A.S. in Liberal Arts from Madison College (2021).
\end{IEEEbiography}

\vskip -2\baselineskip plus -1fil
\begin{IEEEbiography}[{\includegraphics[width=1in,height=1.25in,clip,keepaspectratio]{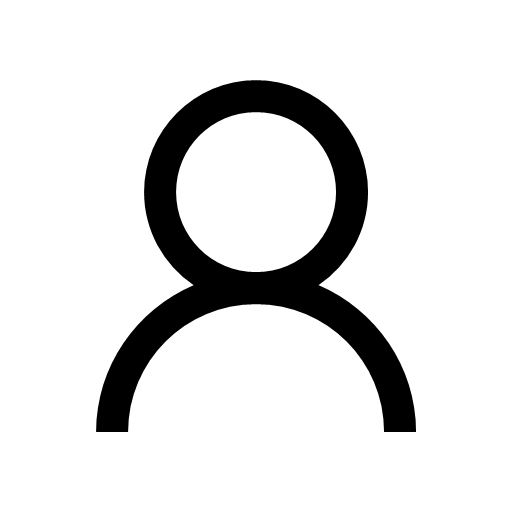}}]{Sébastien Roche}
is a Scientist at the Environmental Defense Fund and Harvard University where he contributes to level2 algorithm development and data analysis for MethaneSAT and MethaneAIR. His research focuses on improving greenhouse gas measurements from ground-, aircraft-, and spacecraft-based instruments with contributions to TCCON, MethaneAIR, MethaneSAT and the proposed Arctic Observing Mission.
\end{IEEEbiography}

\vskip -2\baselineskip plus -1fil
\begin{IEEEbiography}[{\includegraphics[width=1in,height=1.25in,clip,keepaspectratio]{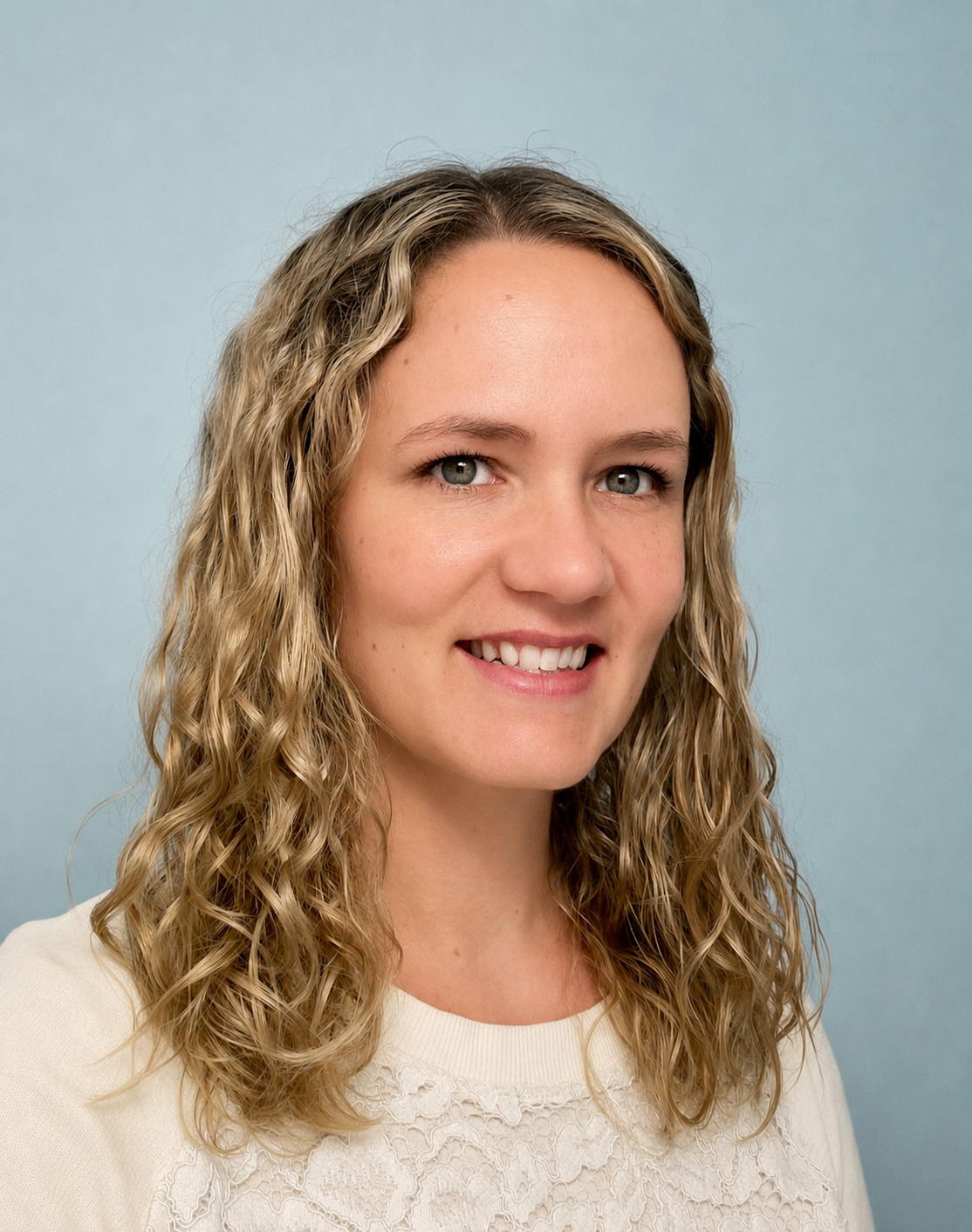}}]{Maryann Sargent} is a senior project scientist at Harvard University. Her research focuses on automated detection and quantification of methane plumes from satellite and airborne platforms.  She led the MethaneAIR aircraft campaigns from 2023-2025, consisting of over 100 science flights and sampling $>$ 80\% of US onshore oil and gas production.  She has an S.B. from MIT and an M.S. and Ph.D. from Harvard University.  
\end{IEEEbiography}

\vskip -2\baselineskip plus -1fil
\begin{IEEEbiography}[{\includegraphics[width=1in,height=1.25in,clip,keepaspectratio]{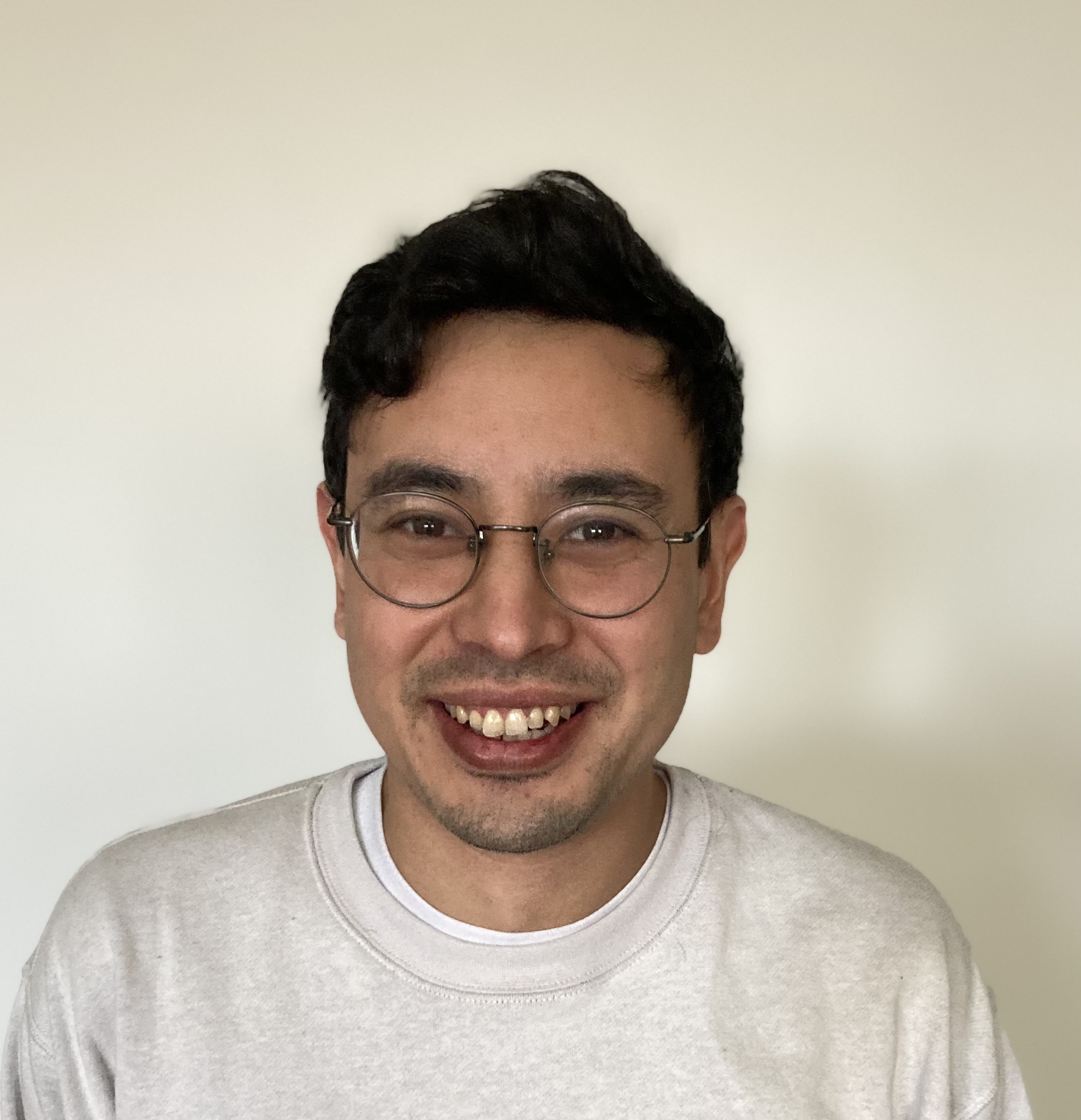}}]{Chris Chan Miller}
Christopher Chan Miller received the Ph.D. in Earth and Planetary Sciences from Harvard University in 2016. 
He is a Senior Scientist at the Environmental Defense Fund, where he leads level2 algorithm development for the MethaneSAT mission. 
His research focuses on atmospheric chemistry, satellite remote sensing, and radiative transfer, with contributions to MethaneSAT, TEMPO, and OMI satellite missions.
\end{IEEEbiography}

\vskip -2\baselineskip plus -1fil
\begin{IEEEbiography}[{\includegraphics[width=1in,height=1.25in,clip,keepaspectratio]{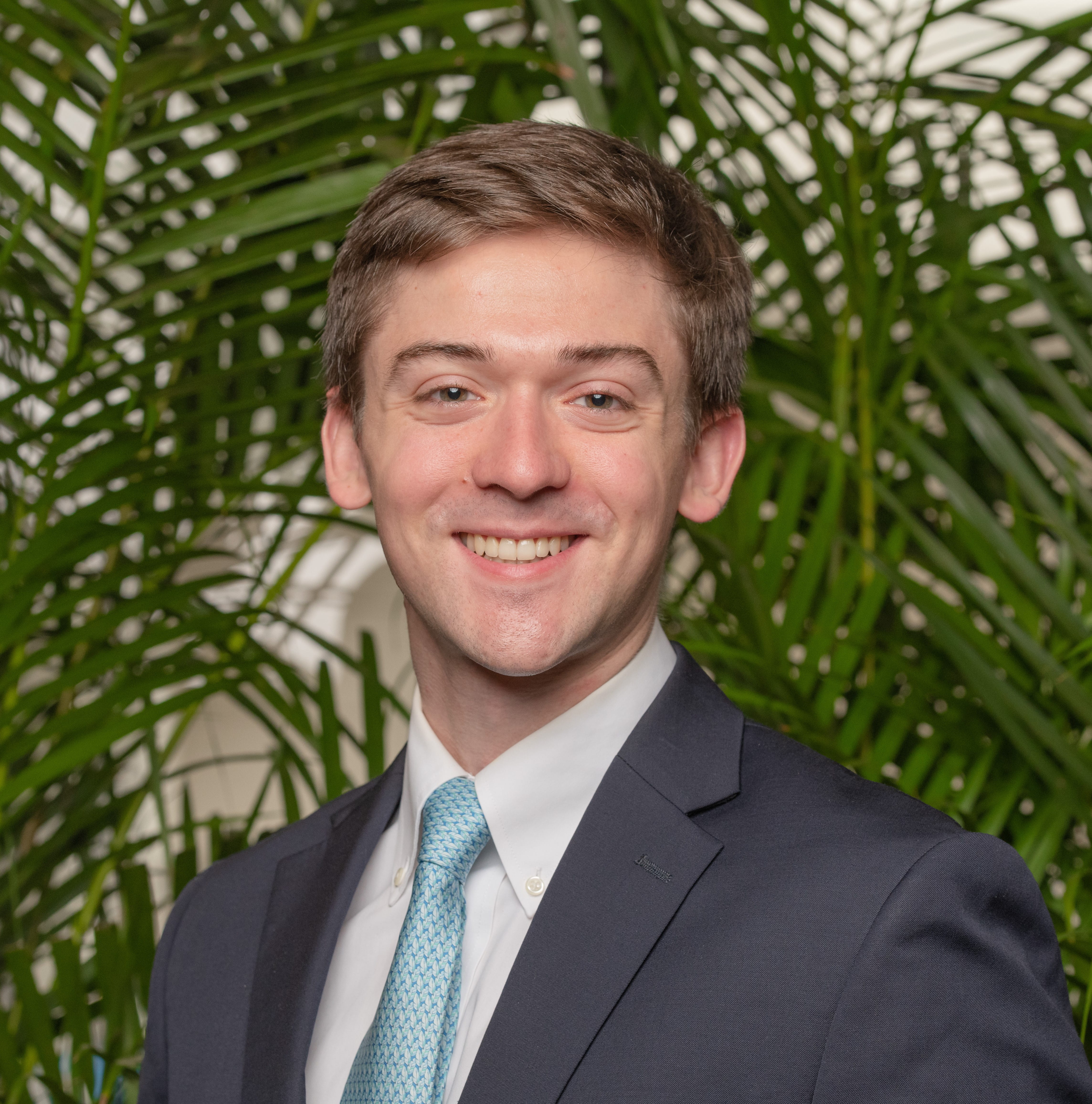}}]{Jack Warren} received a bachelors's degree from Columbia University. He is currently working at Environmental Defense Fund as part of the MethaneSAT science team. His research focuses on the validation, attribution, and analysis of methane emission sources detected by MethaneAIR and MethaneSAT observations globally.
\end{IEEEbiography}

\vskip -2\baselineskip plus -1fil
\begin{IEEEbiography}[{\includegraphics[width=1in,height=1.25in,clip,keepaspectratio]{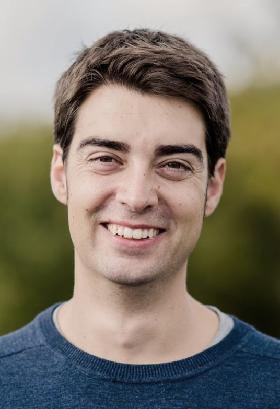}}]{Luis Guanter} received the B.Sc. degree in physics, and the Ph.D. degree in environmental physics from the Universitat de Valencia, Valencia, Spain, in 2002 and 2007, respectively. After several post-doctoral positions in Germany and U.K., he became the Head of the Remote Sensing Section, GFZ, Potsdam, Germany. Since March 2019, he has been a Full Professor of applied physics with the Universitat Politècnica de València (UPV), Valencia, where he is leading the Land and Atmosphere Remote Sensing (LARS) Group. Since March 2022, he has been sharing his position at UPV with a methane remote sensing scientist position at the Environmental Defense Fund, Amsterdam, The Netherlands. Dr. Guanter has been included in Clarivate’s Highly Cited Researchers List of the world’s most influential scientists since 2019
\end{IEEEbiography}

\vskip -2\baselineskip plus -1fil
\begin{IEEEbiography}[{\includegraphics[width=1in,height=1.25in,clip,keepaspectratio]{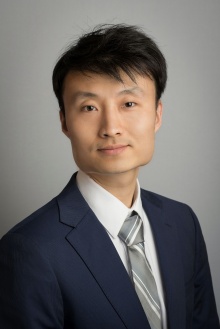}}]{Kang Sun} is an associate professor of Environmental Engineering at University at Buffalo (UB). He received his BS in Environmental Sciences from Peking University  in 2009, China and a PhD in Environmental Engineering from Princeton University in 2015. He worked as a postdoc and physicist at Smithsonian Astrophysical Observatory before joining UB in 2018. He received the NSF CAREER award in 2024 and is the inventor of physical oversampling and the directional derivative approach for satellite level-3 and level-4 product generation. 
\end{IEEEbiography}

\vskip -2\baselineskip plus -1fil
\begin{IEEEbiography}[{\includegraphics[width=1in,height=1.2in,clip,keepaspectratio]{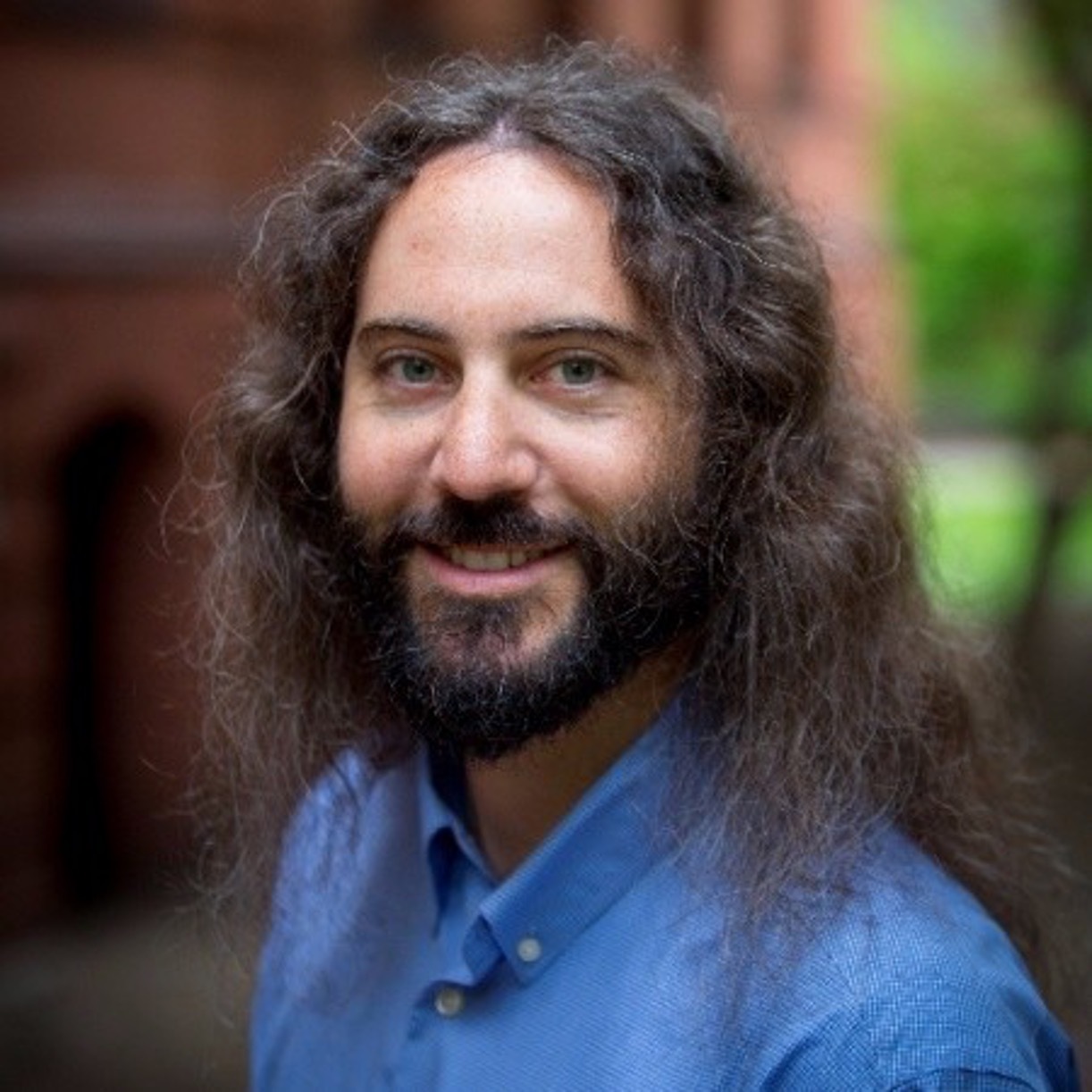}}]{Jonathan E. Franklin} (BS, Marlboro C., 2003; MSc, U. Massachusetts Amherst, 2005; PhD, Dalhousie U., 2015) is a Senior Project Scientist at Harvard University. His scientific work focuses on remote sensing measurements of greenhouse gases from ground, airborne, and satellite platforms. He is the calibration lead for the MethaneSAT and MethaneAIR missions.
\end{IEEEbiography}

\vskip -2\baselineskip plus -1fil
\begin{IEEEbiography}[{\includegraphics[width=1in,height=1.25in,clip,keepaspectratio]{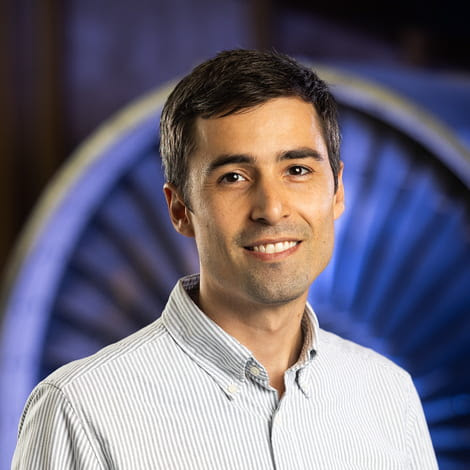}}]{Daniel Varon} is an Assistant Professor in the Department of Aeronautics and Astronautics and the Institute for Data, Systems, and Society at MIT. He received his PhD in atmospheric chemistry from Harvard University in 2020 and held postdoctoral fellowships at Harvard and the Princeton School of Public and International Affairs from 2020 to 2025. He is Model Scientist of the Integrated Methane Inversion (IMI), Co-Nested Model Scientist of the GEOS-Chem global chemical transport model, and Associate Editor for the Atmospheric Measurement Techniques journal of the European Geosciences Union.
\end{IEEEbiography}

\vskip -2\baselineskip plus -1fil
\begin{IEEEbiography}[{\includegraphics[width=1in,height=1.25in,clip,keepaspectratio]{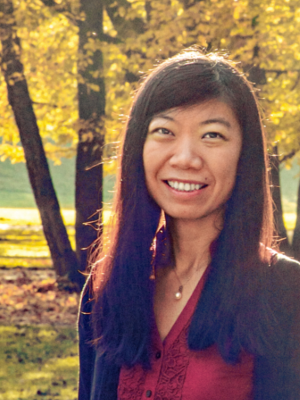}}]{Jia Chen} is Professor of Environmental Sensing and Modeling at the Technical University of Munich (TUM) in the department of Electrical Engineering. In addition, she is an Associate at Harvard University. She studied electrical engineering at KIT (Dipl.-Ing.) and was awarded the Dr.-Ing. degree (summa cum laude) from TUM in 2011. From 2011 to 2015, she was a postdoctoral fellow at Harvard University. Jia Chen’s research focuses on climate change and urban air pollution. She and her team develop novel sensors, sensor networks, mathematical methods, and atmospheric models to localize and quantify greenhouse gas emissions and understand the metabolism of air pollutants in urban environments. She is the Scientific Lead for the pilot city Munich in the EU project “ICOS Cities” and PI of the ERC Consolidator Grant “CoSense4Climate”.
\end{IEEEbiography}

\vskip -2\baselineskip plus -1fil
\begin{IEEEbiography}[{\includegraphics[width=1in,height=1.25in,clip,keepaspectratio]{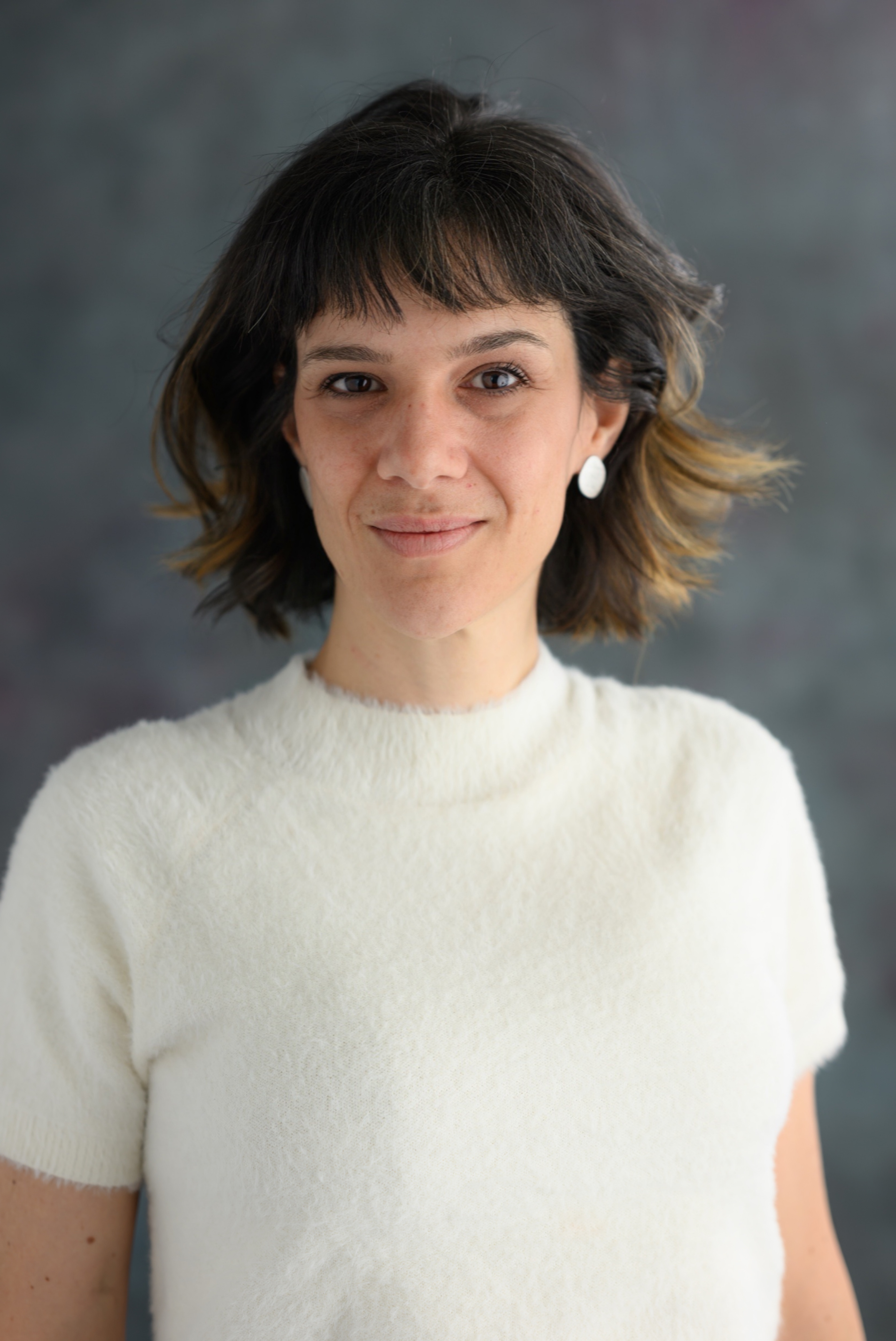}}]{Cecilia Garraffo}
Dr. Cecilia Garraffo is the founder and director of the AstroAI and the EarthAI Institutes at the Center for Astrophysics $\vert$ Harvard \& Smithsonian. She received the Presidential Early Career Award for Scientists and Engineers (PECASE) in 2024, the highest U.S. honor for early-career scientists. Originally from Argentina, she earned an MS in Astronomy, a PhD in Physics, and completed postdoctoral fellowships in AI and astrophysics. Her research spans AI-driven cosmology and the search for life in exoplanet atmospheres.
\end{IEEEbiography}

\vskip -2\baselineskip plus -1fil
\begin{IEEEbiography}[{\includegraphics[width=1in,height=1.2in,clip,keepaspectratio]{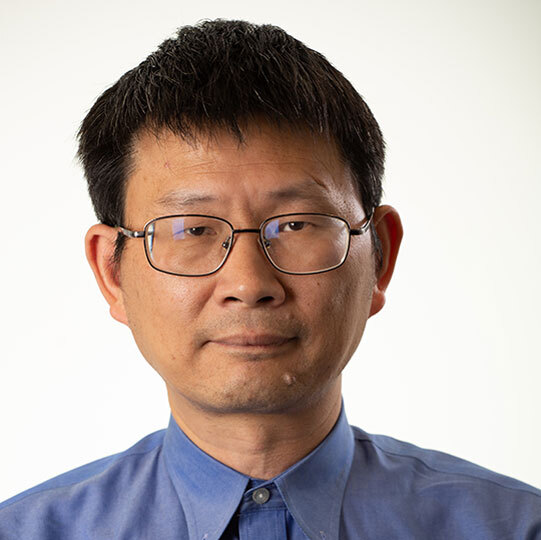}}]{Xiong Liu} received the B.S. degree in environmental chemistry from Nankai University in Tianjin, China, in 1995, and the M.S. degree in environmental chemistry from Research Center for Eco-Environmental Sciences, Chinese Academy of Sciences, Beijing, China, in 1998, and the M.S. degree in computer science and Ph.D. degree in atmospheric science from The University of Alabama in Huntsville, Huntsville, AL, USA, in 2002.
He is currently a Senior Physicist with the Center for Astrophysics $\vert$ Harvard \& Smithsonian, Cambridge, MA, USA. He is the Principal Investigator of the Tropospheric Emissions: Monitoring of Pollution (TEMPO) Project and leads the Atmospheric Measurements Group at CfA. His research interests include the remote sensing of atmospheric trace gases, aerosols, and clouds, satellite mission development, and instrument calibration.
Dr. Liu is a member of American Geophysical Union and American Meteorological Society. He received TEMPO Group Achievement Awards from NASA in 2013, 2020, and 2024, William T. Pecora Award to OMI International Science Team in 2018, AMS Special Award to OMI International Science Team in 2020, and Advances in Atmospheric Science Outstanding Editor Award in 2023 and 2024. 
\end{IEEEbiography}

\vskip -2\baselineskip plus -1fil
\begin{IEEEbiography}[{\includegraphics[width=1in,height=1.2in,clip,keepaspectratio]{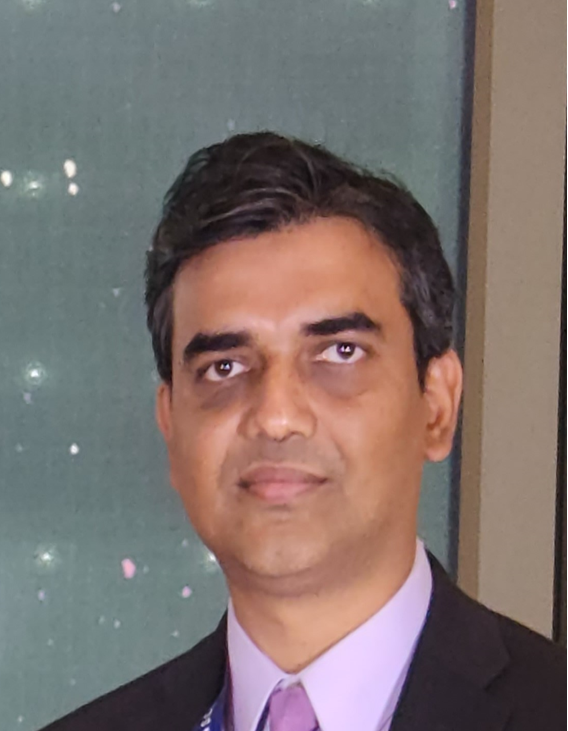}}]{Ritesh Gautam}
Dr. Ritesh Gautam is a Lead Senior Scientist on the MethaneSAT and MethaneAIR projects at Environmental Defense Fund (EDF). Since 2016, he has been overseeing remote sensing and science efforts at EDF in order to produce actionable data products and insights for quantifying and mitigating methane emissions, leveraging the broader multi-satellite ecosystem. Prior to EDF, he served as a tenured faculty member at the Indian Institute of Technology (IIT) Bombay, and previously was a Research Scientist at NASA Goddard Space Flight Center with USRA. His research interests over the last two decades have included fundamentals and applications of remote sensing of methane, aerosols, clouds and cryosphere, as well as developing improved understanding of pollution effects on air quality, climate and monsoon using Earth Observation approaches. He has published over 70 peer-reviewed scientific articles.
\end{IEEEbiography}

\vskip -2\baselineskip plus -1fil
\begin{IEEEbiography}[{\includegraphics[width=1in,height=1.25in,clip,keepaspectratio]{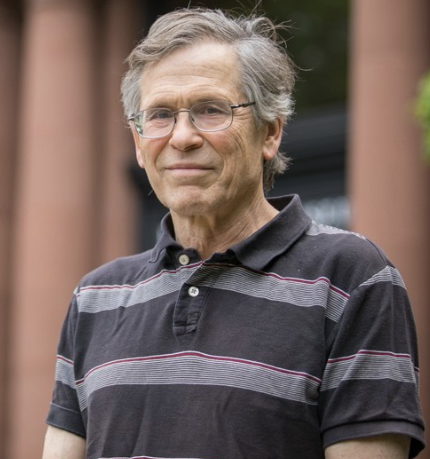}}]{Steven C. Wofsy}
(B.S., University of Chicago, 1966; Ph.D., Harvard, 1971) is the Abbott Lawrence Rotch Professor of Atmospheric and Environmental Chemistry in the John A. Paulson School of Engineering and Applied Science at Harvard University. His scientific work includes measurements and inverse modeling of fluxes of carbon dioxide and methane from ground based, aircraft, and remote sensing measurements. He is the science lead for the MethaneSAT and MethaneAIR imaging spectrometers to measure methane emissions worldwide.
\end{IEEEbiography}

\newpage 

\appendix

\subsection{MethaneAIR Target Basins}
\label{app:target-basins}

\begin{figure}[h]
    \centering
    \includegraphics[width=0.95\linewidth]{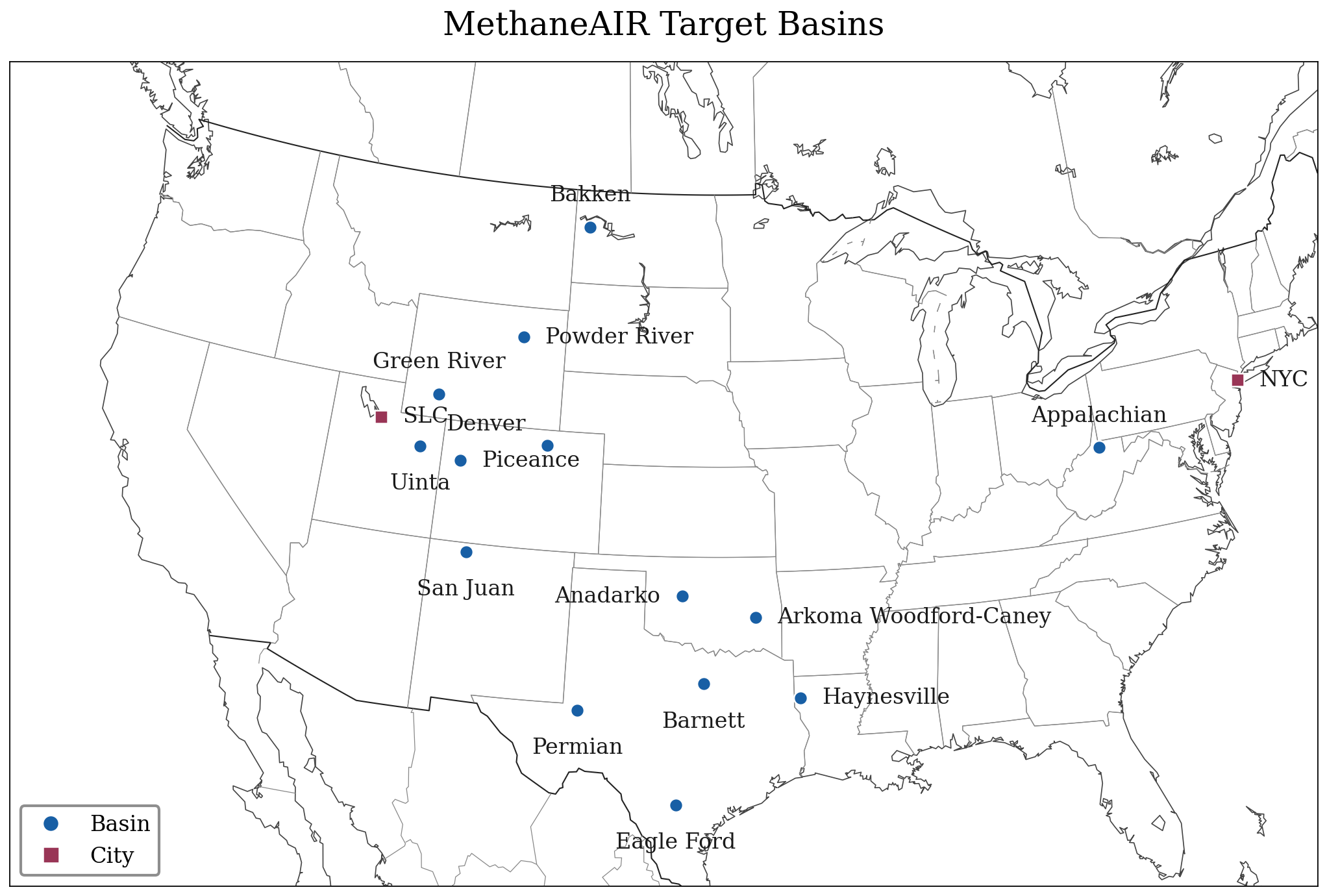}
    \caption{Geographic distribution of MethaneAIR target basins across the
    contiguous United States. Each marker indicates the approximate centroid of
    a basin sampled by the campaign.}
    \label{fig:target-basins}
\end{figure}

This appendix summarizes the geographic distribution of the U.S. oil and gas
basins targeted by MethaneAIR airborne campaigns. Figure~\ref{fig:target-basins}
shows the location of each basin used in this study. Basin centroids are
approximate and intended for visual reference only.

\subsection{MethaneSAT Scene Inventory, Splits, and Per-Scene Results}
\label{app:msat_scenes}

\mpc{Table~\ref{tab:scene_splits} lists the 27 MethaneSAT scenes, their producing basin, the number of reference plume mask labels per scene, and the role of each scene in the three cross-validation folds. The five test scenes (22 reference plumes, four basins) are held out from every fold. The remaining 22 scenes are split scene-exclusively into training and validation; two scenes with few reference plumes (c014703A0, c020C03B0) fall in the training partition of all three folds. Patches from one scene never appear on both sides of a split.}

\mpc{Table~\ref{tab:per_scene} breaks the scene-level results of Table~\ref{tab:scene_results_postprocessing} down by test scene for the fine-tuning configuration. The false positives of the baseline mode concentrate in the South Caspian scene c01A30410 and the two Permian scenes c00040060 and c03320380, and c01A30410 alone accounts for nine of the fifteen; the high-sensitivity mode removes false positives in these three scenes at a cost of at most one recovered plume each. The high-precision mode leaves at most one false positive in any scene, but its misses are not spread evenly: the QND classifier rejects two to three of the five reference plumes of c00040060 and two to three of the eight of c01A30410, and it recovers at most one of the two plumes of c01B90350, while it removes no further plumes in the Permian scene c03320380 or the Uinta-Piceance scene c067906E0. Aggregate scene-level metrics therefore depend strongly on which scenes are held out (Section~\ref{sec:limitations}).}

\begin{table*}[ht!]
\centering
\scriptsize
\begin{tabular}{llcccc}
\toprule
\textbf{Scene} & \textbf{Basin} & \textbf{Ref. plumes} & \textbf{Fold 0} & \textbf{Fold 1} & \textbf{Fold 2} \\
\midrule
c00040060 & Permian & 5 & Test & Test & Test \\
c01A30410 & South Caspian & 8 & Test & Test & Test \\
c01B90350 & Widyan-North & 2 & Test & Test & Test \\
c03320380 & Permian & 6 & Test & Test & Test \\
c067906E0 & Uinta-Piceance & 1 & Test & Test & Test \\
c004F0060 & Permian & 1 & Train & Val & Train \\
c00B90D50 & Uinta & 1 & Train & Val & Train \\
c010204F0 & Maturin & 3 & Train & Train & Val \\
c010C0020 & Appalachian-Marcellus & 15 & Train & Val & Train \\
c012805C0 & Zagros Fold Belt & 5 & Train & Val & Train \\
c01430050 & Permian & 3 & Val & Train & Train \\
c014703A0 & Oued Mya & 1 & Train & Train & Train \\
c01980410 & South Caspian & 6 & Train & Train & Val \\
c01A10330 & Widyan-North & 1 & Train & Train & Val \\
c01AC0330 & Widyan-North & 1 & Val & Train & Train \\
c01CC0380 & Permian & 10 & Val & Train & Train \\
c01D50380 & Permian & 4 & Train & Val & Train \\
c01E003E0 & Amu Dar'ya & 3 & Val & Train & Train \\
c020C03B0 & Sahara & 3 & Train & Train & Train \\
c02D60320 & Widyan-North & 2 & Train & Train & Val \\
c03580390 & Permian & 2 & Val & Train & Train \\
c040D03A0 & Oued Mya & 3 & Val & Train & Train \\
c04130590 & Illizi-Ghadames & 1 & Val & Train & Train \\
c04C605C0 & Zagros Fold Belt & 6 & Train & Train & Val \\
c052E0550 & Eagle Ford & 1 & Train & Val & Train \\
c064C04F0 & Maturin & 5 & Train & Train & Val \\
c06800590 & Illizi-Ghadames & 1 & Train & Val & Train \\
\bottomrule
\end{tabular}
\caption{MethaneSAT scenes used in this study, their producing basin, the number of reference plume mask labels retained after quality filtering, and their role in each cross-validation fold. Test scenes are held out from every fold; splits are scene-exclusive.}
\label{tab:scene_splits}
\end{table*}

\begin{table*}[ht!]
\centering
\small
\footnotesize
\setlength{\tabcolsep}{4pt}
\begin{tabular}{lcccccccccc}
\toprule
\textbf{Scene} & \textbf{Ref.} & \multicolumn{3}{c}{Baseline} & \multicolumn{3}{c}{High Sensitivity} & \multicolumn{3}{c}{High Precision (QND)} \\
\cmidrule(lr){3-5} \cmidrule(lr){6-8} \cmidrule(lr){9-11}
 & \textbf{plumes} & TP & FP & FN & TP & FP & FN & TP & FP & FN \\
\midrule
c00040060 & 5 & 5 & 1.7 (1--3) & 0 & 4.7 (4--5) & 1.0 (0--3) & 0.3 (0--1) & 2.3 (2--3) & 0 & 2.7 (2--3) \\
c01A30410 & 8 & 7.7 (7--8) & 9.0 (7--12) & 0.3 (0--1) & 7.3 (7--8) & 5.0 (4--6) & 0.7 (0--1) & 5.7 (5--6) & 1 & 2.3 (2--3) \\
c01B90350 & 2 & 2 & 0 & 0 & 2 & 0 & 0 & 0.7 (0--1) & 0 & 1.3 (1--2) \\
c03320380 & 6 & 6 & 4.0 (2--5) & 0 & 5.7 (5--6) & 2.3 (1--4) & 0.3 (0--1) & 5.7 (5--6) & 0.3 (0--1) & 0.3 (0--1) \\
c067906E0 & 1 & 1 & 0.3 (0--1) & 0 & 1 & 0 & 0 & 1 & 0 & 0 \\
\midrule
\textbf{Total} & 22 & 21.7 & 15.0 & 0.3 & 20.7 & 8.3 & 1.3 & 15.3 & 1.3 & 6.7 \\
\bottomrule
\end{tabular}
\caption{Per-scene instance detection results for the fine-tuning configuration on the five MethaneSAT test scenes at the operating point of Table~\ref{tab:scene_results_postprocessing} ($\tau = 0.8$, $\delta = 0.2$, $\theta = 0.1$). Counts are over the three cross-validation folds: a single integer when all folds agree, otherwise the fold mean with the min--max range in parentheses. Ref.\ plumes: number of reference plume mask labels in the scene.}
\label{tab:per_scene}
\end{table*}

\subsection{Hyperparameter Selection}
\label{app:hyperparams}

\subsubsection{Learning Rate}
\label{app:lr}

\begin{figure*}[ht!]
    \centering
    \includegraphics[width=\textwidth]{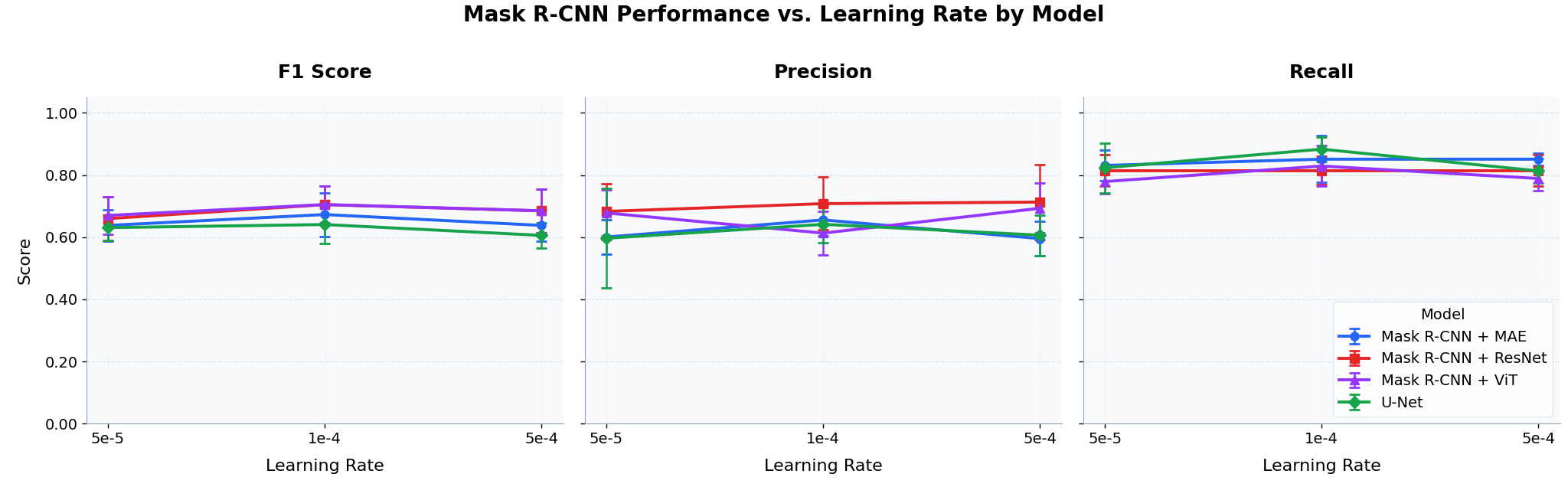}
    \caption{Mask R-CNN and U-Net performance on the MethaneSAT validation set as a function of learning rate for all four architectures evaluated. F1, precision, and recall are shown for learning rates $5\times10^{-5}$, $1\times10^{-4}$, and $5\times10^{-4}$. Error bars indicate standard deviation across three cross-validation folds. All models achieve their best or near-best F1 at $1\times10^{-4}$, which is selected as the operating learning rate for all experiments.}
    \label{fig:lr_comparison}
\end{figure*}

We swept three learning rates: $5\times10^{-5}$, $1\times10^{-4}$, and $5\times10^{-4}$, across all four model architectures (Mask R-CNN with ResNet-50, ViT, and MAE backbones, and U-Net) on the MethaneSAT validation set, holding all other hyperparameters fixed. Figure~\ref{fig:lr_comparison} shows F1, precision, and recall for each model across the three learning rates. Results are reported as mean $\pm$ standard deviation over three cross-validation folds. 

The overall pattern across all architectures is that performance is relatively stable across the range evaluated, with no model exhibiting large degradation at any learning rate. However, $1\times10^{-4}$ consistently achieves the best or near-best F1 for all four models and is selected as the operating learning rate for all experiments reported in the main paper.

The differences between $1\times10^{-4}$ and the other two learning rates are within the cross-fold variance. We therefore apply $1\times10^{-4}$ uniformly across all models for consistency and reproducibility.

\subsection{Trade-off Studies}
\label{app:tradeoff}

\subsubsection{Image Size}
\label{app:imagesize}

The spatial extent of training patches controls a fundamental trade-off in plume detection: larger patches provide more spatial context, capturing more of the plume's downwind tail and the surrounding background needed to characterize concentration gradients, while smaller patches allow more diverse crops per scene under a fixed memory budget. We swept three patch sizes for each instrument independently on the validation set, holding all other hyperparameters fixed, and report AP, precision, recall, and F1.

\begin{figure}[ht!]
    \centering
    \includegraphics[width=\columnwidth]{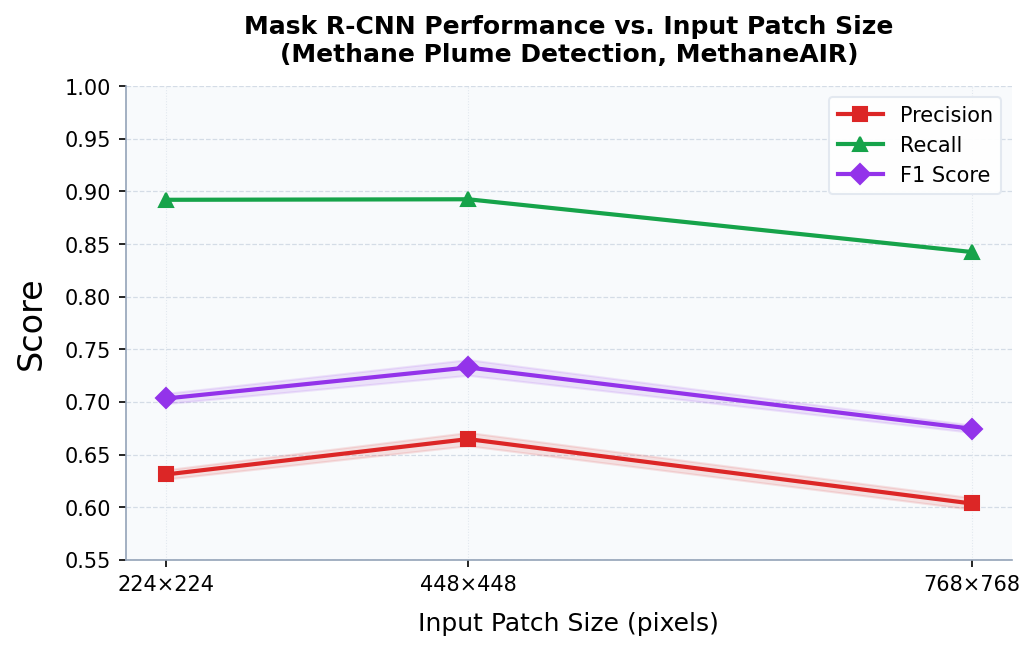}
    \caption{Mask R-CNN (ResNet-50) performance on the MethaneAIR validation set as a function of input patch size. Precision, recall, and F1 are shown for three patch sizes: $224\times224$, $448\times448$, and $768\times768$ pixels. Shaded bands indicate standard deviation across three cross-validation folds. F1 and precision both peak at $448\times448$, which is selected as the operating patch size for all MethaneAIR experiments. The recall decline at larger patch sizes reflects increased background heterogeneity rather than missed plume coverage, consistent with MethaneAIR's high spatial resolution relative to typical plume extents.}
    \label{fig:patch_size_mair}
\end{figure}

\textbf{MethaneAIR.}  Figure~\ref{fig:patch_size_mair} shows precision, recall, and F1 as a function of patch size across three configurations: $224\times224$, $448\times448$, and $768\times768$ pixels. The results reveal a non-monotonic pattern that contrasts with the MethaneSAT findings and reflects the higher spatial resolution of MethaneAIR data.

Recall is high and relatively stable across all three patch sizes, declining only modestly from $0.89$ at $224\times224$ and $448\times448$ to $0.84$ at $768\times768$. This stability indicates that even the smallest patch size provides sufficient spatial context to capture the plume body at MethaneAIR's 10~m resolution, where a $224\times224$ pixel crop covers approximately $2.24\times2.24$~km, a field of view that comfortably encompasses most observed plume extents in the training set.

Precision and F1 both peak at $448\times448$, with precision reaching $0.67$ and F1 reaching $0.74$, before declining at $768\times768$ to $0.61$ and $0.67$ respectively. The precision decline at the largest patch size suggests that at $768\times768$ pixels ($\sim7.7\times7.7$~km at 10~m resolution), each patch contains a larger and more heterogeneous background region, exposing the model to a greater diversity of dispersed enhancement patterns and retrieval artifacts that increase false positive activations without recovering additional true plumes. This behavior is the inverse of the MethaneSAT finding, where larger patches consistently improved recall by capturing more of the plume downwind tail: at MethaneAIR resolution, plumes are already well-contained within moderate-sized patches, so the cost of increased background heterogeneity at large patch sizes outweighs any spatial context benefit.

We select $448\times448$ as the operating patch size for all MethaneAIR experiments, as it achieves the best F1 and precision while maintaining high recall, with narrow confidence bands across cross-validation folds indicating stable generalization.

\begin{figure}[ht!]
    \centering
    \includegraphics[width=\columnwidth]{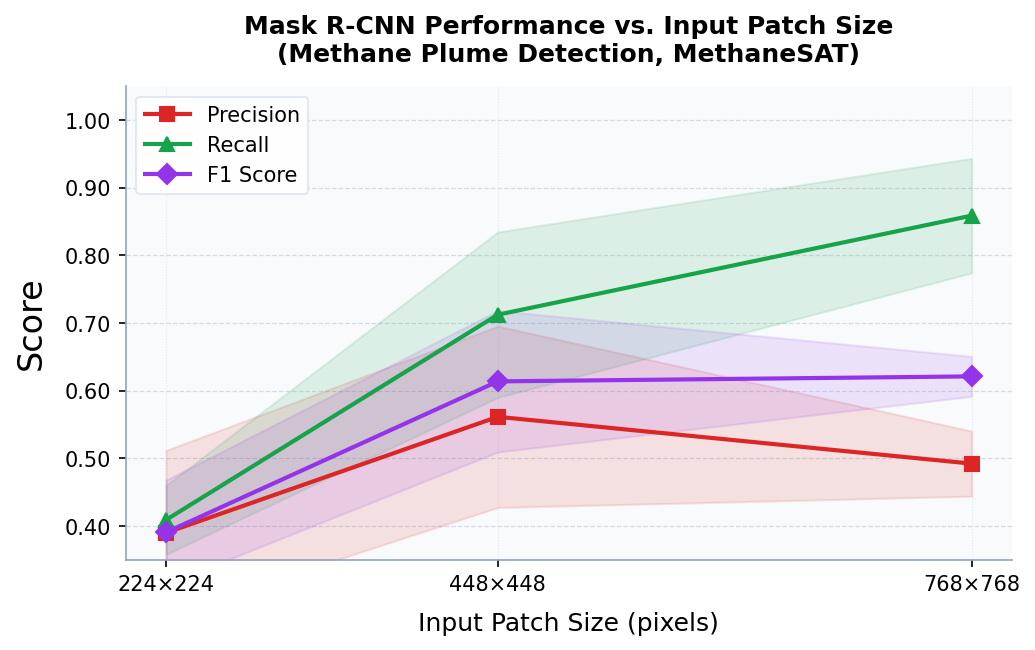}
    \caption{Mask R-CNN (ResNet-50) performance on the MethaneSAT validation set as a function of input patch size. Precision, recall, and F1 are shown for three patch sizes: $224\times224$, $448\times448$, and $768\times768$ pixels. Shaded bands indicate standard deviation across three cross-validation folds. We select 
    $768\times768$ for all MethaneSAT experiments based on its superior recall and competitive F1.}
    \label{fig:patch_size_msat}
\end{figure}

\textbf{MethaneSAT.} Figure~\ref{fig:patch_size_msat} shows precision, recall, and F1 as a function of patch size across three configurations: 
$224\times224$, $448\times448$, and $768\times768$ pixels. The results reveal a clear precision-recall trade-off driven by patch size. Recall improves monotonically and substantially with patch size, from $0.40$ at $224\times224$ to $0.86$ at $768\times768$, reflecting that larger patches provide sufficient spatial context to capture the full extent of wind-dispersed plumes, including their downwind tails, which are frequently truncated or missed when the receptive context is too narrow. 
Precision follows the opposite trend: it increases from $224\times224$ to $448\times448$ (from $0.40$ to $0.55$) but then declines at $768\times768$ (to $0.50$), suggesting that at the largest patch size the model is exposed to a greater diversity of background structures per training example, which broadens sensitivity but also introduces additional spurious activations from dispersed enhancements that 
fall within the larger field of view.

F1 captures this trade-off directly: it improves sharply from $224\times224$ to $448\times448$ (from $0.39$ to $0.60$) and then plateaus at $768\times768$ ($0.62$), with overlapping confidence intervals between the two larger configurations. Despite the modest F1 difference, we select $768\times768$ as the operating patch size for all MethaneSAT experiments. The substantially higher recall at 
this setting ($0.86$ vs. $0.71$ at $448\times448$) is operationally preferable: recovering a greater fraction of real plumes is the primary objective of the detection system, and the precision cost is subsequently addressed by the post-processing pipeline described in Section~\ref{sec:inference}. The wide confidence bands across all three configurations reflect the limited geographic diversity of the MethaneSAT validation set and should be interpreted accordingly.

\subsubsection{Confidence, IoU, and NMS Thresholds}
\label{app:thresholds}

Three thresholds govern the mapping from raw model outputs to instance-level detections at scene level: the confidence threshold $\tau$, which filters individual predictions before scene-level aggregation; the non-maximum suppression (NMS) IoU threshold $\delta$, which controls how aggressively overlapping predictions of the same plume are merged before evaluation; and the evaluation IoU threshold $\theta$, which determines whether a predicted mask is counted as a true positive against a \mpc{reference plume mask label}. $\tau$ and $\delta$ were selected by sweeping each parameter independently on the validation set while holding the others fixed.

\begin{figure}[ht!]
    \centering
    \includegraphics[width=\columnwidth]{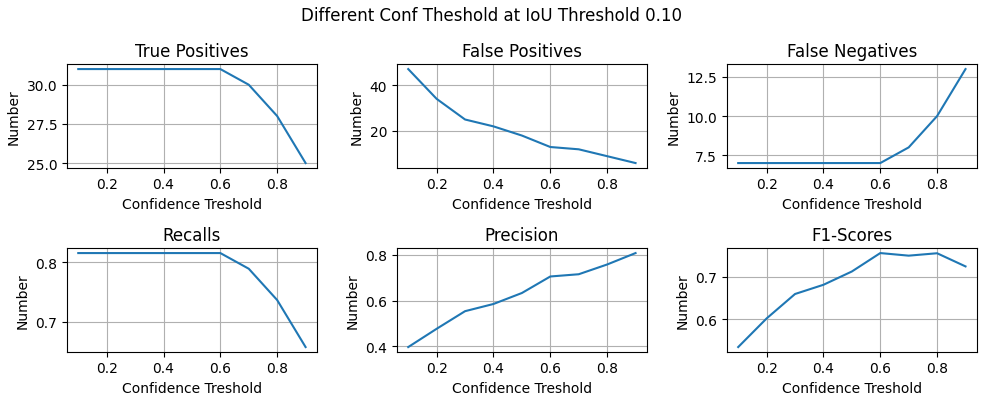}
    \caption{Instance-level detection metrics on the validation set as a function of confidence threshold $\tau$ at fixed IoU threshold $\theta = 0.1$. F1 peaks in the range $\tau \in [0.6, 0.8]$; beyond $\tau = 0.8$, false negatives increase sharply as real detections are discarded. We select $\tau = 0.8$ as the operating point.}
    \label{fig:conf_threshold}
\end{figure}

\textbf{Confidence threshold $\tau$.} Figure~\ref{fig:conf_threshold} shows instance-level TP, FP, FN, precision, recall, and F1 as a function of $\tau$ at fixed $\theta = 0.1$. Below $\tau = 0.4$ the model retains many low-quality predictions, with more than 40 false positives and precision below 0.5. False positives fall steadily as $\tau$ increases while true positives remain stable up to $\tau \approx 0.6$, and beyond $\tau = 0.8$ false negatives rise sharply as genuine detections are discarded. F1 peaks over $\tau \in [0.6, 0.8]$, and we select $\tau = 0.8$ on the validation set.

\begin{figure}[ht!]
    \centering
    \includegraphics[width=\columnwidth]{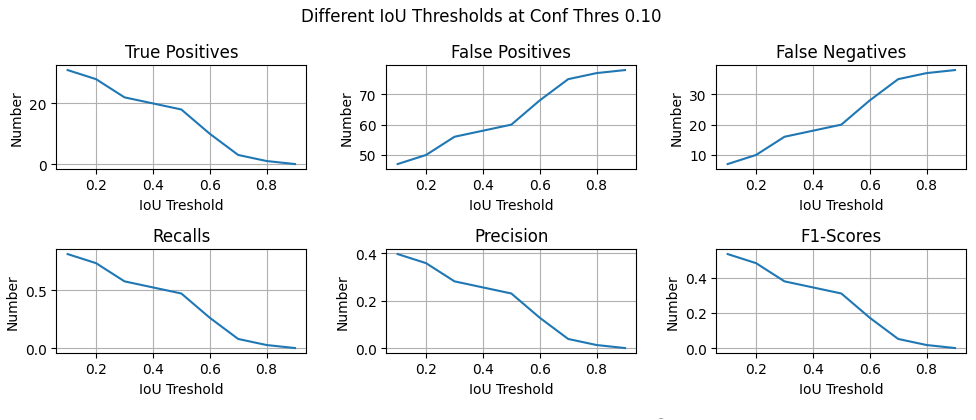}
    \caption{Instance-level detection metrics on the validation set as a function of IoU threshold $\theta$ at fixed confidence threshold $\tau = 0.1$. All metrics deteriorate monotonically as $\theta$ increases, reflecting the inherent difficulty of achieving precise mask overlap for diffuse wind-dispersed plumes. We fix $\theta = 0.1$ as the localization criterion for all scene-level results; stricter thresholds are reported in Appendix~\ref{app:theta_sensitivity}.}
    \label{fig:iou_threshold}
\end{figure}

\textbf{IoU threshold $\theta$.} Figure~\ref{fig:iou_threshold} shows the same six metrics as a function of $\theta$ at a fixed confidence threshold of $\tau = 0.1$. The sweep reveals a strong monotonic deterioration in all performance metrics as $\theta$ increases. At $\theta=0.1$, the model recovers approximately 31 true positives with around 47 false positives; at $\theta=0.5$, true positives fall to about 18 and F1 to about 0.3, and beyond $\theta=0.6$ true positives drop below 10 and F1 below 0.2. This rapid degradation reflects the inherent difficulty of achieving precise mask overlap for wind-dispersed methane plumes, whose lateral boundaries are diffuse by nature and whose annotated extents are determined by the conservative thresholding applied in the wavelet-based labeling method. Requiring strict spatial overlap between predicted and \mpc{reference plume mask labels} therefore penalizes detections that correctly localize a plume source and recover its primary axis but do not exactly reproduce the annotated boundary. \mpc{We do not select $\theta$ from this sweep, since any metric-maximizing choice would favour the lowest value; we fix $\theta = 0.1$ as a localization criterion and report the sensitivity of the test-set results to stricter thresholds in Appendix~\ref{app:theta_sensitivity}.}

\begin{figure}[ht!]
    \centering
    \includegraphics[width=\columnwidth]{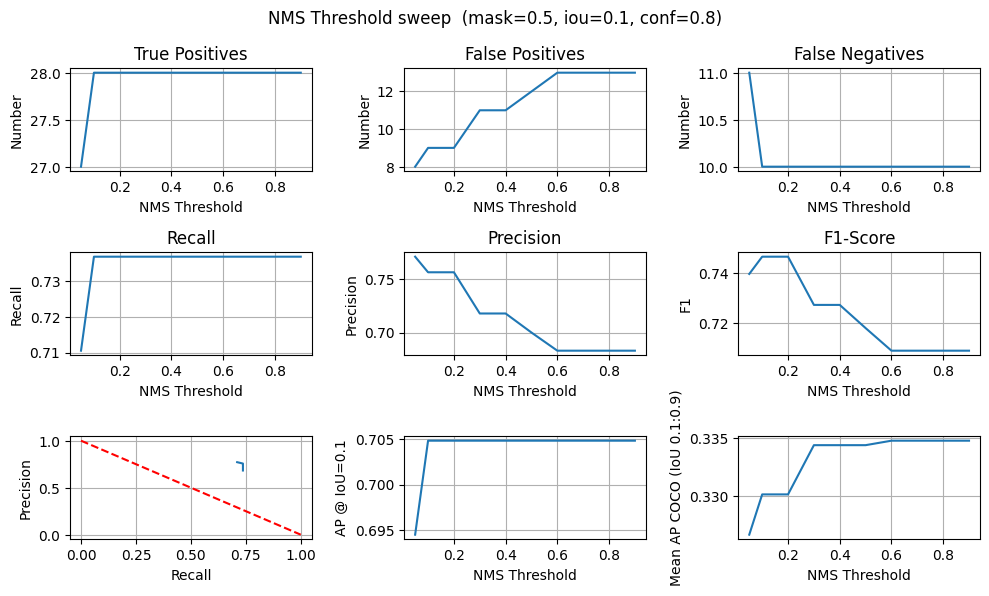}
    \caption{Instance-level detection metrics on the validation set as a function of the NMS IoU threshold $\delta$ at fixed evaluation IoU $\theta = 0.1$ and confidence threshold $\tau = 0.8$. Very aggressive suppression ($\delta = 0.05$) discards a real detection, while loose suppression ($\delta \geq 0.3$) admits duplicate predictions from overlapping sliding-window patches. F1 is maximised on the plateau $\delta \in [0.1, 0.2]$; we select $\delta = 0.2$ as the operating point, which retains all recoverable true positives and leaves headroom for the downstream proximity-based fragment merging stage.}
    \label{fig:nms_threshold}
\end{figure}

Figure~\ref{fig:nms_threshold} shows the same six metrics as a function of $\delta$ at fixed $\theta = 0.1$ and $\tau = 0.8$. At $\delta = 0.05$ suppression is too aggressive and removes one true positive; over $\delta \in [0.1, 0.2]$ the model recovers all 28 detectable true positives with 9 to 10 false positives and precision above 0.74; beyond $\delta \approx 0.3$ near-duplicate predictions from overlapping windows are no longer merged and false positives grow to 13 by $\delta = 0.6$, with F1 falling from 0.747 to 0.71. AP$_{\theta=0.1}$ and the COCO-style mean AP are flat above $\delta = 0.1$. We select $\delta = 0.2$: it retains all true positives and leaves margin for the proximity-based merging stage (Section~\ref{sec:inference}) to act on overlaps just below the NMS cutoff.

\subsubsection{Post-Processing Settings}
\label{app:postprocessing}

Having established the confidence threshold $\tau = 0.8$ and evaluation IoU threshold $\theta = 0.1$ in Appendix~\ref{app:thresholds}, we analyze the incremental contribution of each remaining post-processing stage to scene-level detection performance on the validation set. The two stages evaluated here are morphological filtering, which applies physics-informed constraints on fiber-to-major-axis ratio and fiber width to reject predictions inconsistent with real plume morphology, and proximity-based merging, which groups and unions predictions whose masks intersect across overlapping patch boundaries into single coherent instances. The QND-based high-precision filtering stage is evaluated in Section~\ref{sec:results_scene}.

\begin{figure*}[ht!]
    \centering
    \includegraphics[width=\textwidth]{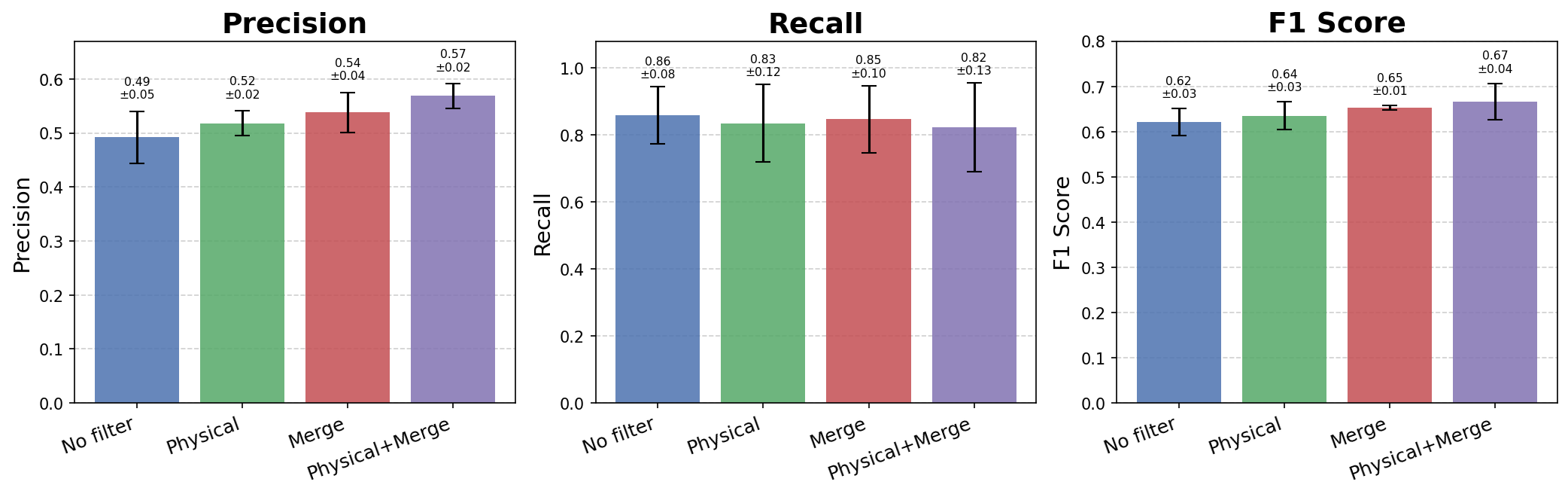}
    \caption{Impact of post-processing stages on instance-level precision, recall, and F1 on the validation set. Four configurations are shown: 
    no filter beyond confidence thresholding and NMS (No filter), morphological filtering alone (Physical), proximity-based merging alone (Merge), and both stages combined (Physical+Merge). Error bars indicate standard deviation across three cross-validation folds. 
    Combining both stages yields the best precision-recall balance, with a 7.6 percentage point precision gain over the baseline at a cost of 
    3.6 percentage points in recall.}
    \label{fig:postprocessing_comparison}
\end{figure*}

Figure~\ref{fig:postprocessing_comparison} shows precision, recall, and F1 on the validation set for four configurations: no filter beyond confidence thresholding and NMS (baseline), morphological filtering alone (Physical), proximity-based merging alone (Merge), and both stages combined (Physical+Merge). Each configuration uses the same 
$\tau = 0.8$ and $\theta = 0.1$ established in Appendix~\ref{app:thresholds}.

The baseline configuration achieves precision of $0.49 \pm 0.05$, recall of $0.86 \pm 0.08$, and F1 of $0.62 \pm 0.03$, confirming that confidence thresholding and NMS alone leave a substantial false positive rate that motivates further filtering. Morphological filtering improves precision to $0.52 \pm 0.02$ with a modest recall reduction to $0.83 \pm 0.12$, yielding an F1 of $0.64 \pm 0.03$. The precision gain reflects the removal of compact circular or irregular predictions that pass the confidence threshold but are inconsistent with the elongated, dispersing morphology of real wind-driven plumes. The recall reduction is small and falls within the variance across folds, indicating that the morphological constraints do not systematically discard real 
plume detections.

Proximity-based merging provides a larger and more consistent precision improvement: applied alone, it raises precision to $0.54 \pm 0.04$ and F1 to $0.65 \pm 0.01$ while maintaining recall at $0.85 \pm 0.10$, comparable to the baseline. This improvement reflects the correction of a systematic artifact of sliding window inference: without merging, a single plume that spans multiple overlapping patches generates several partially overlapping predictions that are each counted as separate false positive instances against the single \mpc{reference plume mask label}. Merging these into a unified instance removes this redundancy and improves the instance-level precision without discarding any real detections.

Combining both stages yields the strongest overall performance: precision of $0.56 \pm 0.02$, recall of $0.82 \pm 0.13$, and F1 of $0.66 \pm 0.04$. The combined improvement represents a 7.6 percentage point precision gain over the baseline with only a 3.6 percentage point recall reduction, a favorable trade-off that confirms the two stages address complementary sources of false positives. 

While the absolute performance gains are moderate, they are achieved entirely through physics-informed constraints grounded in established plume phenomenology, without introducing additional learned parameters or requiring further annotation. The remaining false positives after Physical+Merge are dominated by small enhancements, whose removal is handled by the QND classifier in the high-precision mode (Section~\ref{sec:results_scene}); size filtering is compared in Appendix~\ref{app:size_threshold_vs_QND}.

\subsubsection{MethaneSAT Patch Size with Simulated Data}
\label{app:size_simulated}

\begin{figure*}[ht!]
    \centering
    \includegraphics[width=\textwidth]{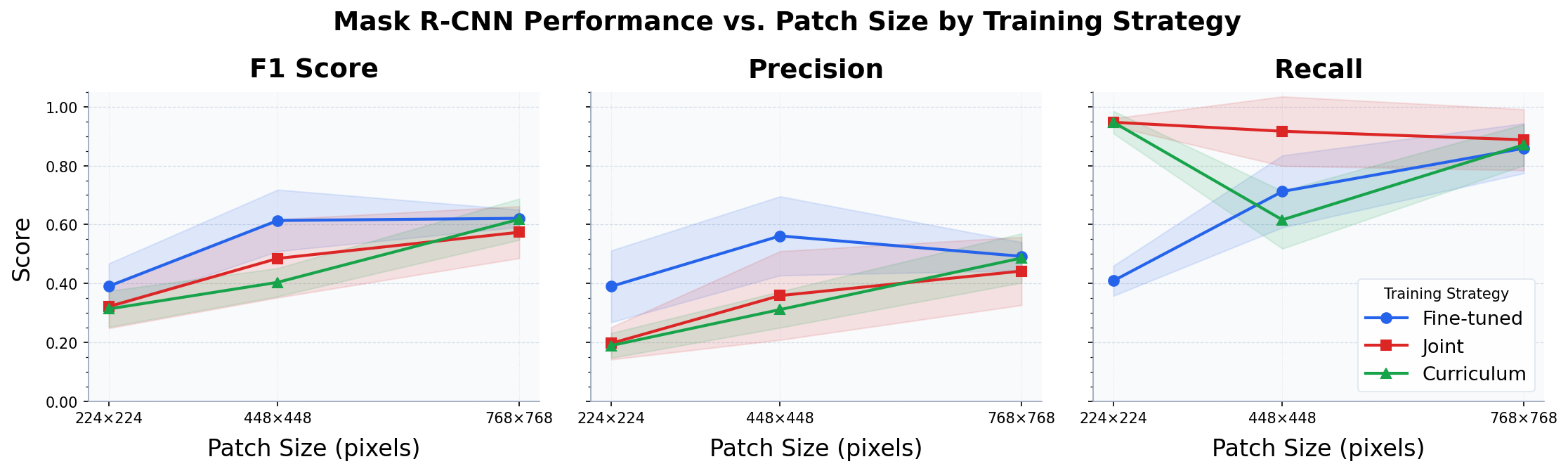}
    \caption{Mask R-CNN (ResNet-50) performance on the MethaneSAT validation set as a function of input patch size, shown separately for three cross-sensor transfer strategies: fine-tuning, joint training, and curriculum learning. F1, precision, and recall are shown for patch sizes of $224\times224$, $448\times448$, and 
    $768\times768$ pixels. Shaded bands indicate standard deviation across three cross-validation folds. All strategies converge toward comparable F1 at $768\times768$, confirming this as the operating patch size for all transfer learning experiments.}
    \label{fig:patch_size_strategies}
\end{figure*}

To assess whether the optimal patch size established for real MethaneSAT data in Appendix~\ref{app:imagesize} generalizes to training configurations that incorporate synthetic plumes, we repeated the patch size sweep across the three cross-sensor transfer strategies that use MethaneAIR pre-trained weights: fine-tuning, joint training, and curriculum learning. Figure~
\ref{fig:patch_size_strategies} shows F1, precision, and recall as a function of patch size for each strategy across the same three configurations evaluated previously: $224\times224$, $448\times448$, and $768\times768$ pixels.

The dominant pattern across all three strategies is a consistent improvement in F1 with patch size, converging toward comparable performance at $768\times768$. At the smallest patch size ($224\times224$), fine-tuning leads with F1 of $0.39$ while joint training and curriculum learning both produce lower F1 values around $0.31$--$0.33$, with correspondingly low precision ($0.19$--$0.40$). As patch size increases to $448\times448$, all three strategies improve substantially in F1, with fine-tuning reaching $0.61$, and joint and curriculum learning reaching $0.49$ and $0.40$ respectively. The precision-recall decomposition at this patch size reveals an interesting divergence: fine-tuning achieves a more balanced profile ($0.56$ precision, $0.71$ recall), while joint training achieves higher recall ($0.91$) at the cost of lower precision ($0.35$).

At $768\times768$, all three strategies converge toward similar F1 scores in the range $0.58$--$0.63$, with recall values clustering around $0.82$--$0.88$ and precision between $0.44$ and $0.49$. The gap between the strategies narrows as patch size grows. Notably, the recall advantage of joint training observed at $448\times448$ is largely preserved at $768\times768$ ($0.88$ vs. $0.82$ for fine-tuning), while its precision gap narrows substantially, yielding comparable F1 scores across all three strategies at the selected operating patch size.

The wide confidence bands across all strategies and patch sizes reflect the limited size of the MethaneSAT validation set and should be interpreted with caution. However, the consistent ordering of strategies within each patch size, and the monotonic F1 improvement with patch size across all three configurations, together confirm that $768\times768$ pixels is the appropriate operating point regardless of which transfer strategy is employed. All cross-sensor transfer experiments reported in the main paper therefore use this patch size.

\subsection{Random Forest Post-Classification for False Positive Suppression}
\label{app:rfc}

Following the approach of Ottenheimer et al. \cite{Ottenheimer2026}, we apply a random forest classifier (RFC) trained on MethaneAIR plume scenes to discriminate true from false plume instances predicted by the Mask R-CNN pipeline on MethaneSAT. The RFC was trained on features extracted from unique plumes observed during 52 MethaneAIR flights from 2021 to 2023.

Input features to the classifier consist of two components. The first is a set of plume metrics computed from each instance mask using Density-Based Spatial Clustering (DBSCAN) \citep{ester1996density} to isolate the high-concentration core: all pixels above the 98th percentile within the predicted mask are extracted and clustered with a neighborhood radius of 10 pixels and a minimum cluster size of 25 pixels. Three summary statistics are then derived:

\begin{align}
\text{Contrast} &= \mu_{\text{mask}} / \mu_{\text{bkrd}} \\
\text{Z-score}  &= (\mu_{\text{mask}} - \mu_{\text{bkrd}}) / 
                    \sigma_{\text{bkrd}} \\
\text{Intensity} &= \mu_{\text{mask}} - \mu_{\text{bkrd}}
\end{align}

\noindent where $\mu_{\text{mask}}$ is the mean XCH$_4$ within the DBSCAN-generated plume mask and $\mu_{\text{bkrd}}$, $\sigma_{\text{bkrd}}$ are the mean and standard deviation of XCH$_4$ outside the DBSCAN mask but within the predicted instance mask.

The second component is a set of Quantile Normality Deviation (QND) polynomial descriptors computed for both methane concentration and surface albedo. For each quantity, the Kolmogorov-Smirnov D-value is evaluated at each percentile $i = 1\ldots100$:

\begin{equation}
D_i = \left\lvert \frac{1}{N} \sum_{j=1}^{N} 
      \mathbf{1}(x_j \leq x_i) - 
      \Phi\!\left(\frac{x_i - \mu}{\sigma}\right) 
      \right\rvert
\end{equation}

\noindent where $x_i = \text{Percentile}_{q_i}(\mathbf{X}_{\text{plume}})$, $N$ is the number of pixels in the mask, and $\mu$, $\sigma$ are the mean and standard deviation of $\mathbf{X}_{\text{plume}}$. A sixth-degree polynomial is fit to the resulting curve, and five descriptors are extracted: the polynomial value at the 50th percentile ($\text{QND}_{\text{CH}_4}(50)$), the albedo values at the 50th and 90th percentiles ($\text{QND}_{\text{alb}}(50, 90)$), and the minimum, maximum, and mean of the polynomial's critical points between the 1st and 99th percentiles, generated for both $\text{QND}_{\text{CH}_4}$ and $\text{QND}_{\text{alb}}$.

For MethaneSAT application, the same feature extraction procedure is applied directly to the Mask R-CNN predicted instance masks at 45~m L3 resolution. No retraining or domain adaptation of the RFC is performed; the classifier is applied zero-shot to the MethaneSAT test set to assess cross-sensor generalizability.

\subsection{Size Filtering vs. QND Classification for High-Precision Mode}
\label{app:size_threshold_vs_QND}

The false positive analysis in Section~\ref{sec:results_scene} identified small compact enhancements as the dominant spurious detection category, accounting for approximately 60\% of validation false positives. A natural and computationally simple response is to apply a minimum mask area threshold, discarding detections below a fixed size. Here we compare this approach with the QND classifier used in the high-precision mode. The QND classifier was chosen before any evaluation on the test scenes, for the physical reasons given in Section~\ref{sec:inference}; the size threshold was selected on the validation folds, and the test-set comparison below is reported for reference, not used for selection.

\begin{figure*}[ht!]
    \centering
    \includegraphics[width=\textwidth]{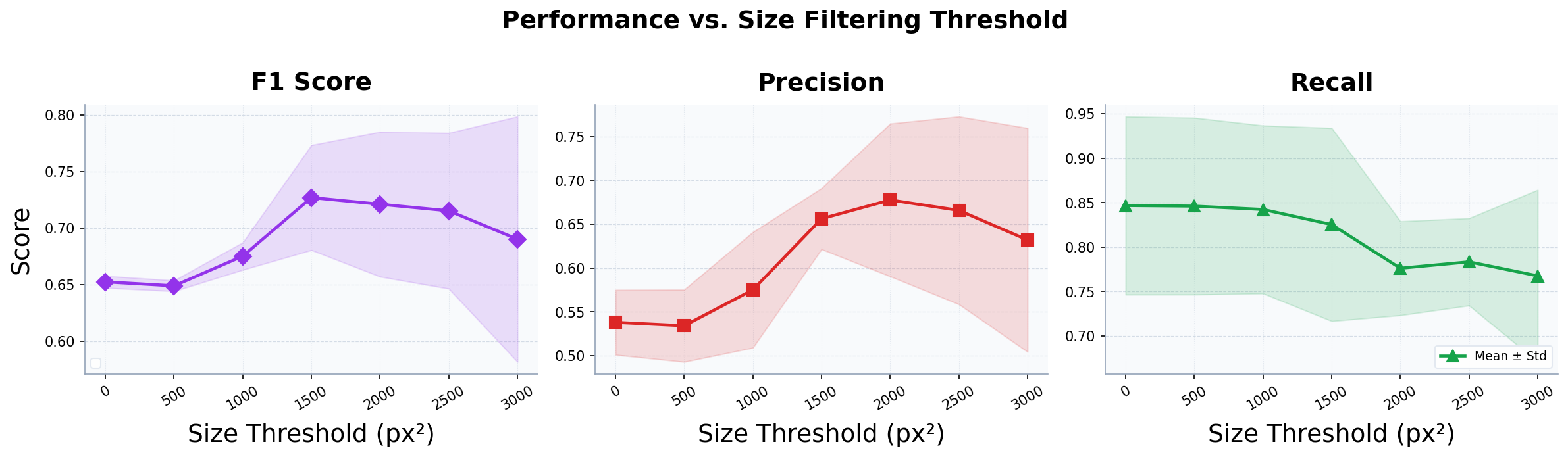}
    \caption{\textbf{Performance vs. size filtering threshold} applied on top of the high-sensitivity mode for the fine-tuning configuration on the MethaneSAT validation set. Precision, recall, and F1 are shown as a function of minimum mask area threshold (px$^2$). Shaded bands indicate standard deviation across three cross-validation folds. F1 peaks at $1500\ \text{px}^2$ before declining as the threshold encroaches on real small plume detections.}
    \label{fig:size_threshold}
\end{figure*}

\textbf{Size filtering.} Figure~\ref{fig:size_threshold} shows precision, recall, and F1 on the validation folds as a function of the minimum mask area, applied on top of the high-sensitivity output of the fine-tuning configuration. This sweep is used only to select the threshold. Validation precision rises from $0.54$ without filtering to $0.66$ at $1500\ \text{px}^2$, while recall decreases slowly (from $0.85$ to $0.83$), and F1 peaks at $1500\ \text{px}^2$ ($0.73$). Beyond $2000\ \text{px}^2$, precision and F1 decline as the filter starts to remove real small plumes, and the cross-fold variance widens. We therefore fix the threshold at $1500\ \text{px}^2$ and apply it once to the test scenes. On the test set (Table~\ref{tab:size_vs_qnd}), it raises precision from $0.71 \pm 0.07$ to $0.78 \pm 0.06$ and lowers recall from $0.94 \pm 0.05$ to $0.83 \pm 0.08$, leaving F1 essentially unchanged ($0.81 \pm 0.07$ to $0.80 \pm 0.05$).

The fundamental limitation of size filtering is that it operates on a single geometric property with no access to the physical content of each detection. A small mask that corresponds to a real weak source and a small mask that corresponds to a surface artifact are treated identically: both are discarded if they fall below the threshold. This means that improving precision via size filtering necessarily sacrifices recall on real low-flux plumes, with no mechanism to distinguish between the two cases.

\textbf{QND classification.} The QND classifier addresses this limitation directly. Rather than filtering by size, it characterizes the within-mask distributions of XCH$_4$ and surface albedo using percentile-based polynomial descriptors, the Quantile Normality Deviation (QND) features, and combines them with geometric plume metrics in a random forest classifier. Real methane enhancements produce a characteristic distributional signature in the methane-albedo feature space. This classifier can in principle retain small real plumes that size filtering would discard, while still rejecting small artifacts whose distributional signatures reveal a non-physical origin.

\textbf{Comparison.} Table~\ref{tab:size_vs_qnd} presents a direct comparison of the two approaches applied on top of the high-sensitivity output for the fine-tuning configuration. At their respective optimal operating points, the QND classifier achieves substantially higher precision ($0.92$ vs. $0.78$) with a lower false positive count (FP = 1 vs. FP = 4), at the cost of a larger recall reduction (recall $0.70$ vs. $0.83$). The F1 scores are comparable ($0.79$ vs. $0.80$), but the operating characteristics differ in a way that is consequential for different use cases. Size filtering at $1500\ \text{px}^2$ retains more true positives (TP = 18 vs. 15) while accepting more false positives, making it more appropriate when the cost of missed detections is high. The QND classifier achieves near-perfect precision (FP = 1) at the cost of three additional false negatives relative to size filtering, making it more appropriate for regulatory reporting and source attribution workflows where false positive rate must be minimized.

\begin{table*}[ht!]
\centering
\begin{tabular}{lcccccc}
\toprule
\textbf{Method} & \textbf{TP} & \textbf{FP} & \textbf{FN} & \textbf{Precision} & \textbf{Recall} & \textbf{F1} \\
\midrule
High Sensitivity (no filtering) & 21 & 8 & 1 & 0.71\scriptsize{$\pm$0.07} & 0.94\scriptsize{$\pm$0.05} & 0.81\scriptsize{$\pm$0.07} \\
Size filtering ($1500\ \text{px}^2$) & 18 & 4 & 4 & 0.78\scriptsize{$\pm$0.06} & 0.83\scriptsize{$\pm$0.08} & 0.80\scriptsize{$\pm$0.05} \\
QND classifier & 15 & 1 & 7 & \textbf{0.92\scriptsize{$\pm$0.04}} & 0.70\scriptsize{$\pm$0.05} & 0.79\scriptsize{$\pm$0.05} \\
\bottomrule
\end{tabular}
\caption{Comparison of size filtering and QND classification as high-precision post-processing strategies, applied on top of the high-sensitivity mode for the fine-tuning configuration. Results are on the MethaneSAT test set, mean $\pm$ standard deviation over three cross-validation folds. IoU threshold $\theta = 0.1$.}
\label{tab:size_vs_qnd}
\end{table*}

A key practical advantage of the QND classifier is its interpretability and cross-instrument generalizability. As demonstrated by Ottenheimer et al., the classifier trained on MethaneAIR data generalizes to Tanager without retraining, suggesting that the distributional signatures it captures reflect physical properties of methane enhancements rather than instrument-specific noise. Size filtering, by contrast, is implicitly instrument- and resolution-dependent: a threshold calibrated at 45 m/pixel MethaneSAT resolution will not transfer directly to instruments with different pixel scales without recalibration. For these reasons, the QND classifier is adopted as the primary high-precision mechanism in this work, with size filtering retained here as a transparent and reproducible baseline for future comparisons.

\subsection{Domain Adaptation}
\label{app:domain_adaptation}

Domain-adversarial neural networks (DANNs) \cite{ganin_2016} offer a complementary feature-level adaptation strategy to fine-tuning, joint training, and curriculum learning. By incorporating a domain classifier with a gradient reversal layer between the feature extractor, DANNs train the backbone to produce representations that are simultaneously discriminative for the primary detection task and invariant to sensor origin. This adversarial alignment occurs during training without requiring paired data or a separate generative model. The feature-level alignment mechanism is a natural fit for settings where source and target instruments share the same retrieval physics, as is the case for MethaneAIR and MethaneSAT.

Starting from MethaneAIR pre-trained weights, the shared ResNet-50 FPN backbone is trained jointly with two domain classifiers operating at different levels of the feature hierarchy. The first operates on globally pooled FPN feature maps and enforces image-level domain invariance across instruments. The second operates on RoI-pooled features processed by the box head, enforcing instance-level invariance across individual candidate plume regions. Each classifier receives features routed through a Gradient Reversal Layer (GRL), which passes activations unchanged during the forward pass but negates and scales gradients during backpropagation, incentivizing the backbone to produce sensor-invariant representations. The total training objective is:

\begin{equation}
\mathcal{L} = \mathcal{L}^{\text{src}}_{\text{det}} + \mathcal{L}^{\text{tgt}}_{\text{det}} + \lambda_{\text{da}} \left( \mathcal{L}^{\text{img}}_{\text{dom}} + \mathcal{L}^{\text{inst}}_{\text{dom}} \right)
\end{equation}

\noindent where $\mathcal{L}^{\text{src}}_{\text{det}}$ and $\mathcal{L}^{\text{tgt}}_{\text{det}}$ are the standard Mask R-CNN multi-task losses on MethaneAIR and MethaneSAT mini-batches respectively, and $\mathcal{L}^{\text{img}}_{\text{dom}}$ and $\mathcal{L}^{\text{inst}}_{\text{dom}}$ are binary cross-entropy domain classification losses. Mini-batches are composed with equal sampling from both instruments to ensure balanced domain signal throughout training. The GRL reversal strength $\lambda_{\text{da}}$ is annealed from 0 to 1 following the schedule $\lambda(p) = \frac{2}{1 + e^{-\gamma p}} - 1$, where $p \in [0,1]$ is the training progress fraction and $\gamma = 10$, matching the original schedule of \cite{ganin_2016}. This annealing ensures that domain alignment pressure is introduced gradually, allowing the detection objective to stabilize before the adversarial signal becomes dominant.

\textbf{Patch-level results.} Domain adaptation achieves a patch-level F1 of $69.96 \pm 6.43\%$, recall of $79.49 \pm 4.83\%$, and precision of $67.44 \pm 9.76\%$ on the MethaneSAT test set. This places it below the other MethaneAIR-initialized strategies trained on real MethaneSAT labels and close to the MethaneSAT-only baseline, despite starting from MethaneAIR pre-trained weights. The elevated cross-fold variance relative to fine-tuning, joint training, and curriculum learning is a consistent indicator of unstable adversarial training dynamics.

\textbf{Scene-level results.} Table~\ref{tab:scene_results_dann} presents instance-level scene detection results across the three operating modes. At baseline, domain adaptation produces 17 true positives, 32 false positives, and 5 false negatives, with precision of 0.35 and F1 of 0.48. This is substantially worse than fine-tuning at baseline (FP = 15, F1 = 0.74) and only modestly better than the MethaneSAT-only configuration in terms of false positive count, despite the MethaneAIR initialization. The high-sensitivity mode reduces false positives to 21 while recall declines slightly (TP = 16, FN = 6), and the high-precision mode further suppresses false positives to 6 at the cost of 9 false negatives, yielding precision of 0.68 and F1 of 0.62. High cross-fold variance persists across all modes (e.g. $\pm$0.17 in F1 under high-precision mode), indicating that the instability observed at patch level propagates through the full post-processing pipeline.

\begin{table*}[ht!]
\centering
\begin{tabular}{lccccccc}
\toprule
\textbf{Mode} & \textbf{TP} & \textbf{FP} & \textbf{FN} & \textbf{mAP} & \textbf{F1} & \textbf{Precision} & \textbf{Recall} \\
\midrule
Baseline & 17 & 32 & 5 & 0.76\scriptsize{$\pm$0.11} & 0.48\scriptsize{$\pm$0.11} & 0.35\scriptsize{$\pm$0.09} & 0.77\scriptsize{$\pm$0.16} \\
High Sensitivity & 16 & 21 & 6 & 0.69\scriptsize{$\pm$0.13} & 0.53\scriptsize{$\pm$0.09} & 0.43\scriptsize{$\pm$0.04} & 0.71\scriptsize{$\pm$0.20} \\
High Precision & 13 & 6 & 9 & 0.59\scriptsize{$\pm$0.17} & 0.62\scriptsize{$\pm$0.17} & 0.68\scriptsize{$\pm$0.12} & 0.58\scriptsize{$\pm$0.20} \\
\bottomrule
\end{tabular}
\caption{Instance-level scene detection results for domain-adversarial adaptation (DANN) across operating modes. Evaluation protocol follows Section~\ref{sec:results_scene}: confidence threshold $\tau = 0.8$, NMS IoU $\delta = 0.2$, evaluation IoU $\theta = 0.1$, patch size $768\times768$. Results are mean $\pm$ standard deviation over three cross-validation folds.}
\label{tab:scene_results_dann}
\end{table*}

These results are consistent with known limitations of adversarial training under limited target data. When the target domain is underrepresented relative to the source, or the domain gap is small compared to intra-class variability, the gradient reversal objective can interfere with task-discriminative feature learning rather than isolating domain-invariant representations \cite{zhao_2019, kamnitsas_2017}. In the MethaneAIR-to-MethaneSAT setting, where both instruments share the same retrieval physics and plume morphology, the domain gap may be sufficiently small that adversarial alignment provides no benefit over standard fine-tuning while introducing optimization instability. We therefore do not recommend domain-adversarial adaptation as a primary transfer strategy for this setting, and report these results here for completeness.

\subsection{Quantitative Evaluation of Confidence-Weighted Detection Maps}
\label{app:prob_maps}

To characterize the correspondence between the confidence-weighted detection maps and XCH$_4$ maps across the full test set, we computed pixel-level Pearson and Spearman correlations between $\hat{P}$ and XCH$_4$ for all pixels with $\hat{P} > 0$, pooled across all MethaneSAT test scenes. Restricting the analysis to pixels with nonzero $\hat{P}$ isolates the population of pixels that the model has claimed as part of at least one detected instance, which is the relevant population for assessing whether the spatial structure of $\hat{P}$ reflects genuine concentration gradients within plume regions rather than uninformative background scatter.

\begin{figure}[ht!]
    \centering
    \includegraphics[width=\columnwidth]{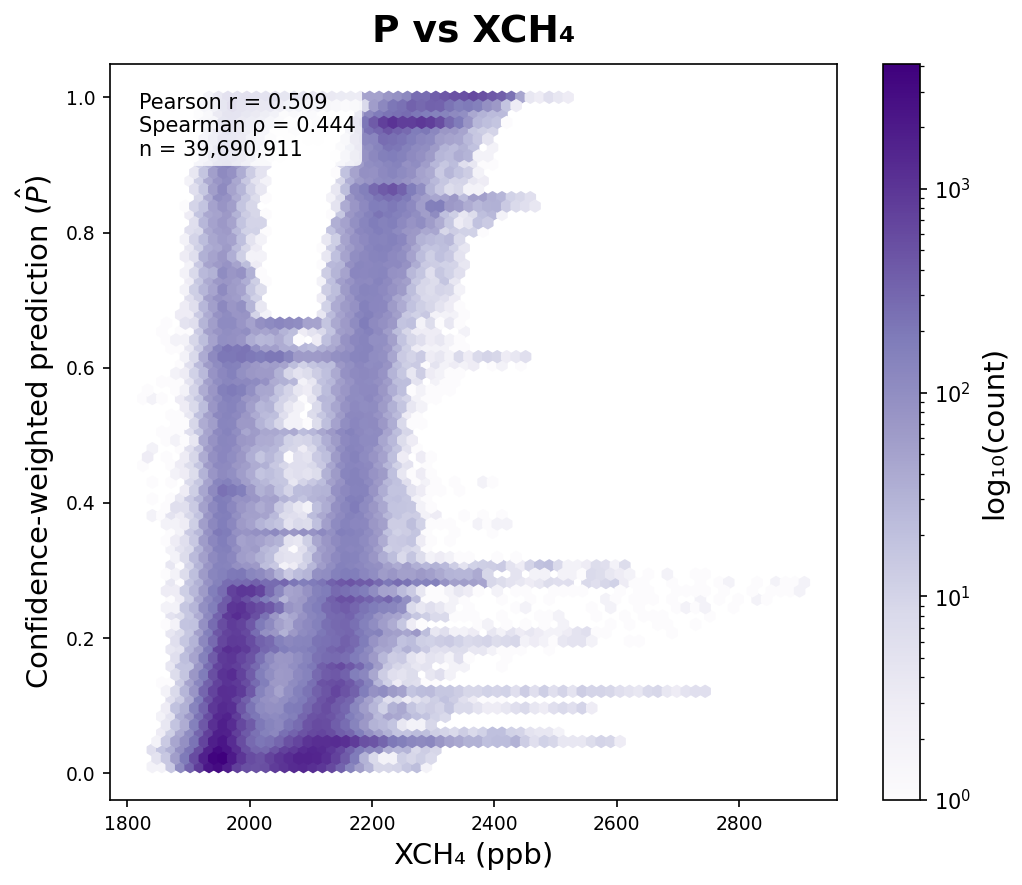}
    \caption{\textbf{Pixel-level correspondence between the confidence-weighted detection map $\hat{P}$ and XCH$_4$, pooled across all MethaneSAT test scenes}. Bins are colored by $\log_{10}$ pixel count. Pearson $r = 0.509$ and Spearman $\rho = 0.444$ indicate a consistent positive tendency for higher XCH$_4$ values to receive higher $\hat{P}$ within detected regions, while the concentration range of the dense low-$\hat{P}$ population reflects plume tail and boundary pixels where detection confidence is moderate despite modest absolute 
    enhancement.}
    \label{fig:prob_scatter_pooled}
\end{figure}

Figure~\ref{fig:prob_scatter_pooled} shows the resulting scatter plot. The pooled correlations are $r = 0.509$ (Pearson) and $\rho = 0.444$ (Spearman). The moderate positive Pearson correlation indicates a consistent linear tendency for higher XCH$_4$ values to receive higher $\hat{P}$ within detected regions, which is physically expected: along a wind-dispersed plume, concentration decreases from source to downwind tail while detection confidence tracks the same gradient. The Spearman correlation is comparable in magnitude to Pearson, reflecting that this positive tendency holds approximately in rank order across the population. The gap between them is modest, suggesting the relationship is reasonably monotonic within the $\hat{P} > 0$ subpopulation, in contrast to the full-pixel analysis of individual scenes where the dominance of near-zero background pixels produces a larger divergence between the two statistics.

The scatter structure in Figure~\ref{fig:prob_scatter_pooled} reveals two partially distinct pixel populations. The dense vertical band concentrated around $\text{XCH}_4 \approx 1950$--$2050$~ppb spans nearly the full range of $\hat{P}$ values, corresponding to pixels within detected instances that lie in low-enhancement regions adjacent to source hotspots or at the dispersed downwind tails of plumes, where the model assigns positive but moderate $\hat{P}$ despite relatively modest absolute concentration. The sparser population extending toward higher XCH$_4$ values above 2200~ppb is dominated by high-confidence pixels ($\hat{P} > 0.8$), corresponding to the high-concentration cores of confirmed plume instances. This two-population structure explains why both correlations are moderate rather than strong: the detection map does not simply reproduce the XCH$_4$, but instead reflects the model's spatial confidence in plume localization, which integrates concentration gradient, morphological plausibility, and cross-patch consistency.

These results should be interpreted as a preliminary characterization rather than a validated uncertainty product. The confidence-weighted aggregation scheme is not a calibrated probability estimator: the $\hat{P}$ values reflect the internal confidence rankings of the Mask R-CNN detector, which are not guaranteed to correspond to well-calibrated posterior probabilities of plume presence. Nonetheless, the consistent positive correlation between $\hat{P}$ and XCH$_4$ within detected regions across $\sim$40 million pixels from the full test set confirms that the detection maps carry meaningful spatial information beyond the binary instance masks. This motivates their use as spatial priors for downstream emission quantification, as discussed in Section~\ref{sec:discussion}, and supports further investigation of calibration methods that could convert the raw aggregated scores into reliable uncertainty estimates.

\subsection{Computational Performance}
\label{app:computational-performance}

\begin{table*}[ht!]
\centering
\label{tab:compute}
\small
\begin{tabular}{lcccccc}
\toprule
 & \multicolumn{1}{c}{Throughput} & \multicolumn{1}{c}{Memory} &
   \multicolumn{1}{c}{Training} &
   \multicolumn{3}{c}{Inference (s / $1{,}000\,\text{km}^2$)} \\
\cmidrule(lr){5-7}
Sensor & (MP/s) & (GB) & (hrs/epoch) &
  Baseline & High-sens. & High-prec. \\
\midrule
MethaneAIR & 5.40  & 1.23 & 0.98 & 3.44 & 4.73 & 4.78 \\
MethaneSAT & 11.86 & 1.23 &  0.25 & 0.17 & 0.23 & 0.25 \\
\bottomrule
\end{tabular}
\caption{Computational cost of the Mask R-CNN segmentor for MethaneAIR
($10\,\text{m/pixel}$) and MethaneSAT ($45\,\text{m/pixel}$). Throughput,
memory, and per-epoch training time are reported once per sensor.
Inference cost per $1{,}000\,\text{km}^2$ is reported for each of the three
modes.}
\end{table*}

We report inference and training cost for the Mask R-CNN segmentor on both applications. All numbers are averaged across the three cross-validation folds (best checkpoint per fold) and measured on a single \texttt{NVIDIA RTX A6000} GPU with mixed precision disabled. Memory reflects peak allocator usage during a
forward pass. Throughput is computed over the validation set in batches of one and therefore represents a worst-case latency estimate; batched throughput is higher. Training time per epoch is the mean wall-clock time for one full pass over the training fold. Inference time per $1{,}000\,\text{km}^2$ is computed by running the complete sliding-window pipeline on $N = 6$ real L3 scenes per sensor and dividing wall-clock scene time by mean scene area.

Three inference configurations are reported. \textbf{Baseline} is the per-patch non-maximum suppression ($\delta=0.2$) and confidence threshold  ($\tau=0.8$) across the assembled scene; \textbf{High-sensitivity} adds skeletonisation-based physical-shape constraints and a proximity-graph spatial-fragment merging step to suppress over-segmented and physically implausible predictions. \textbf{High-precision} further applies the QND based classifier, reducing false positives and maximizing precision.

\subsection{Sensitivity of Scene-Level Results to the Matching Threshold}
\label{app:theta_sensitivity}

\mpc{Tables~\ref{tab:theta_sweep} and~\ref{tab:theta_sweep_counts} recompute the scene-level test-set results of Table~\ref{tab:scene_results_postprocessing} for the fine-tuning configuration at the stricter matching thresholds $\theta = 0.3$ and $\theta = 0.5$; predictions, post-processing, and reference labels are identical, only the matching criterion changes. All metrics decrease with $\theta$: baseline F1 falls from $0.74$ to $0.55$ and $0.41$, and recall from $0.98$ to $0.73$ and $0.55$, so that at $\theta = 0.5$ only 12 of the 22 reference plumes are matched. Two features of the sweep matter for interpretation. First, the ranking of the modes changes: the F1 advantage of the high-sensitivity and high-precision modes over the baseline disappears at $\theta \geq 0.3$, because fragment merging produces spatially larger masks whose IoU with the conservative reference masks is lower even when the source is correctly identified. Second, the loss is concentrated in a single scene: in the South Caspian scene c01A30410 the baseline matches 7--8 of 8 reference plumes at $\theta = 0.1$ but 4 of 8 at $\theta = 0.3$, whereas the other four scenes lose at most one plume each. Stricter thresholds therefore measure how closely predicted extents follow the conservative reference boundaries, and this is where the present system is weakest; they do not change the source-detection picture of Section~\ref{sec:results_scene}.}

\begin{table*}[ht!]
\centering
\small
\footnotesize
\setlength{\tabcolsep}{4pt}
\begin{tabular}{llccccccccc}
\toprule
\textbf{Strategy} & \textbf{Mode} & \multicolumn{3}{c}{$\theta = 0.1$} & \multicolumn{3}{c}{$\theta = 0.3$} & \multicolumn{3}{c}{$\theta = 0.5$} \\
\cmidrule(lr){3-5} \cmidrule(lr){6-8} \cmidrule(lr){9-11}
 &  & \textbf{F1} & \textbf{Precision} & \textbf{Recall} & \textbf{F1} & \textbf{Precision} & \textbf{Recall} & \textbf{F1} & \textbf{Precision} & \textbf{Recall} \\
\midrule
\multirow{3}{*}{Fine-tuning} & Baseline & 0.74\scriptsize{$\pm$0.05} & 0.60\scriptsize{$\pm$0.07} & 0.98\scriptsize{$\pm$0.03} & 0.55\scriptsize{$\pm$0.05} & 0.44\scriptsize{$\pm$0.06} & 0.73\scriptsize{$\pm$0.00} & 0.41\scriptsize{$\pm$0.03} & 0.33\scriptsize{$\pm$0.04} & 0.55\scriptsize{$\pm$0.05} \\
 & High Sensitivity & 0.81\scriptsize{$\pm$0.07} & 0.71\scriptsize{$\pm$0.07} & 0.94\scriptsize{$\pm$0.05} & 0.55\scriptsize{$\pm$0.10} & 0.49\scriptsize{$\pm$0.10} & 0.64\scriptsize{$\pm$0.09} & 0.42\scriptsize{$\pm$0.06} & 0.37\scriptsize{$\pm$0.06} & 0.48\scriptsize{$\pm$0.05} \\
 & High Precision (QND) & 0.79\scriptsize{$\pm$0.05} & 0.92\scriptsize{$\pm$0.04} & 0.70\scriptsize{$\pm$0.05} & 0.52\scriptsize{$\pm$0.05} & 0.60\scriptsize{$\pm$0.04} & 0.45\scriptsize{$\pm$0.05} & 0.43\scriptsize{$\pm$0.08} & 0.50\scriptsize{$\pm$0.09} & 0.38\scriptsize{$\pm$0.07} \\
\bottomrule
\end{tabular}
\caption{Scene-level instance detection on the MethaneSAT test set as a function of the matching IoU threshold $\theta$ (Mask R-CNN ResNet-50, fine-tuning). Confidence threshold $\tau = 0.8$, NMS IoU $\delta = 0.2$, patch size $768\times768$; mean $\pm$ standard deviation over three cross-validation folds. The first column group is the operating point of Table~\ref{tab:scene_results_postprocessing}.}
\label{tab:theta_sweep}
\end{table*}

\begin{table*}[ht!]
\centering
\small
\begin{tabular}{llcccccccccccc}
\toprule
\textbf{Strategy} & \textbf{Mode} & \multicolumn{4}{c}{$\theta = 0.1$} & \multicolumn{4}{c}{$\theta = 0.3$} & \multicolumn{4}{c}{$\theta = 0.5$} \\
\cmidrule(lr){3-6} \cmidrule(lr){7-10} \cmidrule(lr){11-14}
 &  & \textbf{mAP} & \textbf{TP} & \textbf{FP} & \textbf{FN} & \textbf{mAP} & \textbf{TP} & \textbf{FP} & \textbf{FN} & \textbf{mAP} & \textbf{TP} & \textbf{FP} & \textbf{FN} \\
\midrule
\multirow{3}{*}{Fine-tuning} & Baseline & 0.92\scriptsize{$\pm$0.04} & 22 & 15 & 0 & 0.70\scriptsize{$\pm$0.02} & 16 & 21 & 6 & 0.52\scriptsize{$\pm$0.03} & 12 & 25 & 10 \\
 & High Sensitivity & 0.91\scriptsize{$\pm$0.05} & 21 & 8 & 1 & 0.69\scriptsize{$\pm$0.06} & 14 & 15 & 8 & 0.52\scriptsize{$\pm$0.04} & 11 & 18 & 11 \\
 & High Precision (QND) & 0.68\scriptsize{$\pm$0.04} & 15 & 1 & 7 & 0.48\scriptsize{$\pm$0.04} & 10 & 7 & 12 & 0.43\scriptsize{$\pm$0.07} & 8 & 8 & 14 \\
\bottomrule
\end{tabular}
\caption{Mean average precision and instance counts on the MethaneSAT test set as a function of the matching IoU threshold $\theta$ (same protocol as Table~\ref{tab:theta_sweep}; TP, FP, FN are fold means over 22 reference plumes).}
\label{tab:theta_sweep_counts}
\end{table*}

\vfill
\end{document}